\documentclass{article}

\usepackage{microtype}
\usepackage[graphicx]{realboxes}
\usepackage{subcaption}
\usepackage{booktabs} 

\usepackage{hyperref}

\usepackage[accepted]{icml2024}

\usepackage{amsmath}
\usepackage{amssymb}
\usepackage{mathtools}
\usepackage{amsthm}
\usepackage{multirow}
\usepackage{duckuments}
\usepackage{longtable}
\usepackage{wrapfig}
\usepackage{placeins}
\usepackage{caption}
\usepackage{colortbl}
\usepackage{xcolor}
\usepackage{microtype}

\usepackage[capitalize,noabbrev]{cleveref}

\theoremstyle{plain}

\theoremstyle{definition}

\theoremstyle{remark}

\usepackage[textsize=tiny]{todonotes}
\usepackage{colortbl}
\usepackage{xcolor}

\newcommand{\textoverline}[1]{$\overline{\mbox{#1}}$}

\icmltitlerunning{MOMENT: A Family of Open Time-series Foundation Models}

\begin{document}

\twocolumn[
\icmltitle{MOMENT: A Family of Open Time-series Foundation Models}

\icmlsetsymbol{equal}{*}

\begin{icmlauthorlist}
\icmlauthor{Mononito Goswami}{cmu}
\icmlauthor{Konrad Szafer}{equal,cmu}
\icmlauthor{Arjun Choudhry}{equal,cmu}
\icmlauthor{Yifu Cai}{cmu}
\icmlauthor{Shuo Li}{upenn}
\icmlauthor{Artur Dubrawski}{cmu}
\end{icmlauthorlist}

\icmlaffiliation{cmu}{Auton Lab, Robotics Institute, Carnegie Mellon University, Pittsburgh, USA}
\icmlaffiliation{upenn}{University of Pennsylvania, Philadelphia, USA}

\icmlcorrespondingauthor{Mononito Goswami}{mgoswami@andrew.cmu.edu}

\icmlkeywords{Time series analysis, foundation models, pre-training, evaluation, forecasting, classification, anomaly detection, imputation}

\vskip 0.3in
]



\printAffiliationsAndNotice{\icmlEqualContribution} 

\begin{abstract}
We introduce \textbf{\texttt{MOMENT}}, a family of open-source foundation models for general-purpose time series analysis. Pre-training large models on time series data is challenging due to (1) the absence of a large and cohesive public time series repository, and (2) diverse time series characteristics which make multi-dataset training onerous. Additionally, (3) experimental benchmarks to evaluate these models, especially in scenarios with limited resources, time, and supervision, are still in their nascent stages. To address these challenges, we compile a large and diverse collection of public time series, called the Time series Pile, and systematically tackle time series-specific challenges to unlock large-scale multi-dataset pre-training. Finally, we build on recent work to design a benchmark to evaluate time series foundation models on diverse tasks and datasets in limited supervision settings. Experiments on this benchmark demonstrate the effectiveness of our pre-trained models with minimal data and task-specific fine-tuning. Finally, we present several interesting empirical observations about large pre-trained time series models. Pre-trained models (\texttt{AutonLab/MOMENT-1-large}) and Time Series Pile (\texttt{AutonLab/Timeseries-PILE}) are available on \href{Huggingface}{https://huggingface.co/AutonLab}.

\end{abstract}
\section{Introduction}
\label{sec:intro}

\begin{figure}[!tb]
\centering
\includegraphics[width=0.8\linewidth]{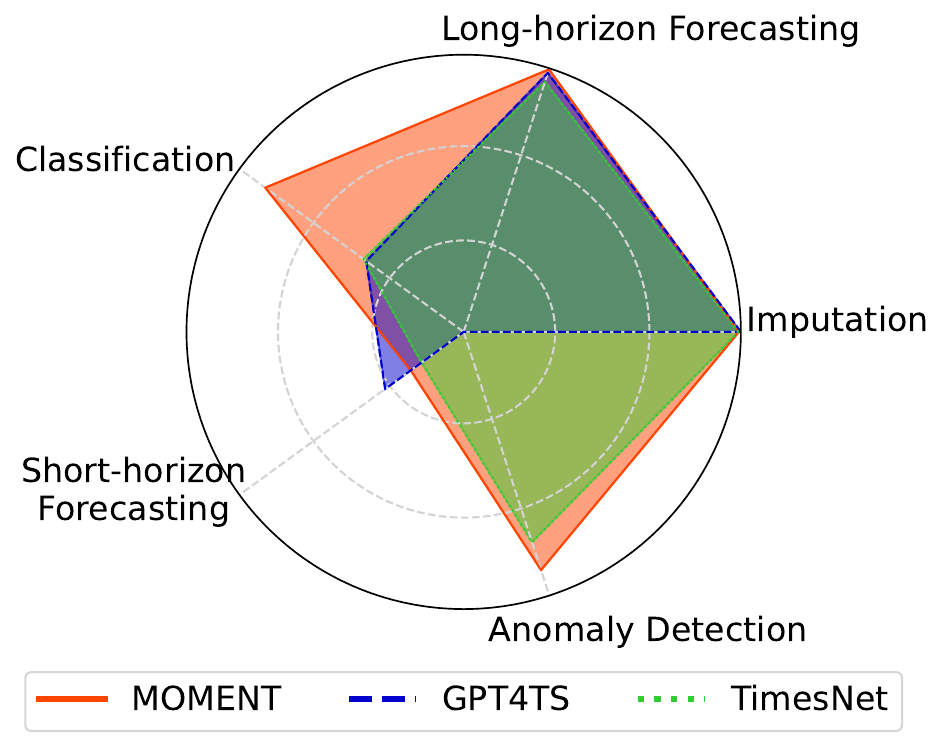}
\caption{\texttt{MOMENT} can solve multiple time series analysis tasks well (App.~\ref{app:experimental-setup-and-results}).}
\label{fig:model_comparison}
\end{figure}

Time series analysis is an important field encompassing a wide range of applications ranging from forecasting weather patterns~\citep{climate_modeling} or detecting irregular heartbeats using Electrocardiograms~\citep{ecg_timeseries}, to identifying anomalous software deployments~\citep{anomaly_detection_kpis}. Due to its significant practical value and the unique challenges that modeling time series data poses, time series analysis continues to receive substantial interest from academia and industry alike. However, modeling such data typically requires substantial domain expertise, time, and task-specific design. 

Large pre-trained language~\citep{llama2, bert, flant5}, vision~\citep{blip2}, and video~\citep{video-pretrained-transformer} models, typically perform well on a variety of tasks on data from diverse domains, with little or no supervision, and they can be specialized to perform well on specific tasks. We unlock these key capabilities for time series data and release the \textbf{first family of open-source large pre-trained time series models}, which we call \texttt{MOMENT}. The models in this family (1) serve as a building block for diverse \textbf{time series analysis tasks} (e.g., forecasting, classification, anomaly detection, and imputation, etc.), (2) are effective \textbf{out-of-the-box}, i.e., with no (or few) particular task-specific exemplars (enabling e.g., zero-shot forecasting, few-shot classification, etc.), and (3) are \textbf{tunable} using in-distribution and task-specific data to improve performance. 

\texttt{MOMENT} is a family of high-capacity transformer models, pre-trained using a masked time series prediction task on large amounts of time series data drawn from diverse domains. Below we summarize our key contributions. 

\textbf{C1: Pre-training data.} A key limiting factor for pre-training large time series models from scratch was the lack of a large cohesive public time series data repositories \citep{one-fits-all, llmtime, timellm, tinytimemixers, tempo}. Therefore, we compiled \textbf{The Time  series Pile}, a large collection of publicly available data from diverse domains, ranging from healthcare to engineering to finance. The Time Series Pile comprises of over 5 public time series databases, from several diverse domains for pre-training and evaluation (Tab. \ref{tab:time-series-pile}).

\textbf{C2: Multi-dataset pre-training.} Unlike text and images, which have largely consistent sampling rates and number of channels, time series frequently vary in their temporal resolution, number of channels\footnote{Temporal resolution reflects sampling frequency of time series (e.g., hourly, daily); Channel is a single univariate time series in multivariate data \citep{tinytimemixers}.}, lengths, and amplitudes, and sometimes have missing values. As a result, large-scale mixed dataset pre-training is largely unexplored. Instead, most methods are trained on a single dataset, and transferred across multiple datasets, but with modest success \citep{timesnet, meta-learning-zero-shot-ts-forecasting, Meta-learning-few-shot-ts-classification}. 

\textbf{C3: Evaluation.} Holistic benchmarks to evaluate time series foundation models on diverse datasets and tasks are in their nascent stages. Recent studies \citep{aqua} have highlighted the importance of well-defined benchmarks and large-scale experimentation in order to accurately assess the impact and effectiveness of novel methodologies. To evaluate \texttt{MOMENT}, we build on the multi-task time series modeling benchmark first proposed by \citet{timesnet} along multiple dimensions. For each of the \textbf{5} time series modeling tasks, namely, short- and long-horizon forecasting, classification, anomaly detection, and imputation we evaluate \texttt{MOMENT} against (1) both state-of-the-art deep learning as well as statistical baselines, on (2) more task-specific datasets, (3) using multiple evaluation metrics, (4) exclusively in limited supervision settings (e.g., zero-shot imputation, linear probing for forecasting, unsupervised representation learning for classification). 

Finally, we explore various properties of these pre-trained time series models. In particular, we study whether \texttt{MOMENT} is aware of intuitive time series characteristics such as frequency and trend, and the impact of initialization, model size scaling, and cross-modal transfer.

\section{Related Work}
\label{sec:lit-review}

\textbf{Transformers and patching for time series modeling.} There is a growing body of work utilizing transformers for various time series analysis tasks \citep{transformers-in-ts-survey}. One issue with applying transformers to time series data is the complexity of the self-attention mechanism, which grows quadratically with the size of input tokens (or length of time series) \citep{li2019enhancing}. \citet{patchtst} demonstrated that treating time series sub-sequences (or patches) as tokens instead of individual time points is a simple, efficient, and effective mechanism for learning useful representations for forecasting. Drawing inspiration from prior work, we build on top of the transformer architecture which takes disjoint time series sub-sequences (or patches) as input.

\textbf{Masked Representation Learning.} Masked pre-training is a widely-used self-supervised learning task where a model learns to accurately reconstruct masked portions of its input. Masked language \citep{bert,unified-text-to-text-transformer} and image modeling \citep{SimMIM,scaling-lang-image-pretraining-via-masking} have been successfully utilized to learn models from vast quantities of unlabeled data, which can generalize to a variety of downstream tasks. 

For time series data, prior work has primarily focused on contrastive representation learning \citep{TS2Vec,ts-tcc,unsupervised-scalable-representation-learning-for-multivariate-ts}. However, contrastive learning relies on data augmentation, which is both subjective and data-dependent. In contrast, some studies mask portions of time series using zeros and learn a model to reconstruct them \citep{patchtst,transformer-multivariate-ts-representation-learning,simmtm,ti-mae}. 

Representation learning via masking is well-suited to all the downstream tasks we care about, especially forecasting and imputation, as they are instances of the masked reconstruction problem. Due to its simplicity and success in vision and language domains, we use the masked prediction task to pre-train our model, using a special embedding (see \texttt{[MASK]} in Fig.~\ref{fig:MOMENT-overview}) to mask time series patches instead of zeros.

\textbf{Cross-modal transfer learning using language models.} \citet{pretrained-transformers-as-universal-computation-engines} had first shown that transformers pre-trained on text data (LLMs) can effectively solve sequence modeling tasks in other modalities. Subsequently, \citet{orca} introduced ORCA, a general cross-modal fine-tuning framework that extends the applicability of a single large-scale pretrained model to diverse modalities by adapting to a target task via an align-then-refine workflow. Given the target input, ORCA first learns an embedding network that aligns the embedded feature distribution with the pretraining modality, then the pretrained model is fine-tuned on the embedded data, exploiting the knowledge shared across modalities. Some recent studies have leveraged this inherent ability of language pre-trained transformers to ``reprogram" LLMs for time series analysis using parameter efficient fine-tuning and suitable tokenization strategies \citep{one-fits-all, llmtime, timellm, tempo, tinytimemixers}. However, some of these models \citep{timellm, llmtime} with billions of parameters demand significant memory and computational resources to perform well. We complement this line of research with three empirical observations (Sec~\ref{sec:moment_properties}): we show that (1) transformers trained on time series can also model sequences across modalities, (2) during pre-training, randomly initializing weights lead to lower pre-training loss, than initializing with language modeling weights, and (3) models pre-trained on time series outperform LLM-based models such as \citep{one-fits-all, timellm} on many tasks and datasets. 

\textbf{Unanswered Questions.} To the best of our knowledge, two questions remain largely unanswered in prior work on time series modeling. First, all existing time series models are (pre-)trained and fine-tuned on individual datasets \citep{patchtst,TS2Vec,timesnet,one-fits-all}, and the benefits (or drawbacks) of large-scale multi-dataset pre-training remains unexplored \citep{transformers-in-ts-survey}. Second, there is very limited work on time series modeling in limited supervision settings, such as zero-shot forecasting \citep{meta-learning-zero-shot-ts-forecasting}, or few-shot classification \citep{Meta-learning-few-shot-ts-classification}. In our work, we consider both these questions and \textit{show that pre-training a model of sufficient capacity on a large corpus of unlabeled time series data can in fact enable it to provide reasonably accurate predictions in limited-supervision settings.}

\section{Methodology}
\label{sec:methodology}

We first collect a large number of public time series data into the \textbf{Time Series Pile} and then use it to pre-train a \textbf{transformer model} on the \textbf{masked time series prediction task}. We discuss each of these steps in the following sections.

\subsection{The Time Series Pile}
\label{subsec:timeseries_pile}

\begin{figure}[tb!]
\centering
\includegraphics[width=\linewidth]{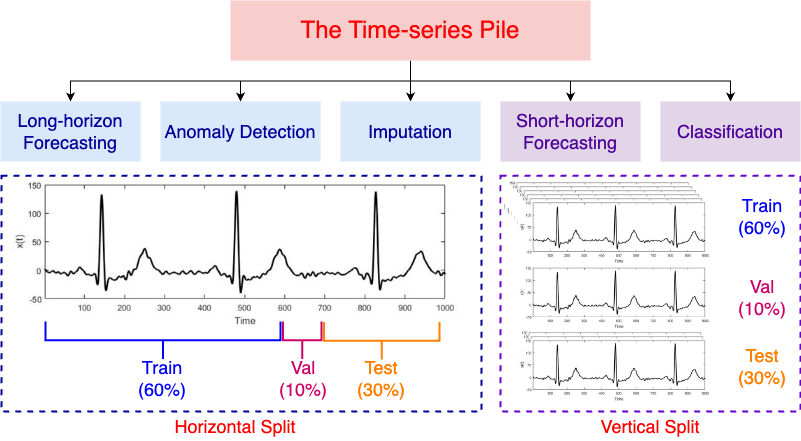}
\caption{\textbf{Time Series Pile data splits}. To avoid data contamination, we carefully partition all datasets into disjoint train, validation, and test splits. We adhere to the predefined splits provided by the creators of each dataset. In cases where such splits are unavailable, we randomly sample 60\% of the data for training, 10\% for validation, and 30\% for testing. We only use the training splits of all datasets for pre-training.}
\label{fig:data-split}
\end{figure}

Unlike natural language processing and computer vision, where large-scale datasets such as The Pile \citep{pile}, and ImageNet-1K \citep{ImageNet} are easily available for pre-training, public time series datasets are much smaller, scattered, and largely task-specific \citep{ts-pretrained-models-survey, one-fits-all, llmtime}. To bridge this gap, we collate multiple time series from 4 task-specific, widely-used \textbf{public} repositories resulting in a large number of time series spanning diverse domains, and time series characteristics such as lengths, amplitudes, and temporal resolutions. We call this collection the Time Series Pile. 

\textbf{Informer long-horizon forecasting datasets} \citep{Informer} is a collection of 9 datasets that are widely used to evaluate long-horizon forecasting performance \citep{timesnet, patchtst, NHITS}: 2 hourly and minutely subsets of the Electricity Transformer Temperature (\texttt{ETT})~\citep{Informer}, Electricity~\citep{electricity-load-diagrams}, Traffic~\citep{PeMS}, Weather~\citep{MPGWeatherData}, Influenza-like Illness (\texttt{ILI})~\citep{CDCFluDashboard}, and Exchange-rate \citep{long-short-temporal-patterns-with-dnns}.

\textbf{Monash time series forecasting archive} \citep{monash} is a collection of 58 publicly available short-horizon forecasting datasets with a total of over 100K time series, spanning a variety of domains and temporal resolutions. 

\textbf{UCR/UEA classification archive} \citep{UCR-Archive} comprises of 159 time series datasets which are frequently used to benchmark classification algorithms \citep{dl-for-ts-classification}. These datasets belonging to seven different categories (Image Outline, Sensor Readings, Motion Capture, Spectrographs, ECG, Electric Devices, and Simulated Data), vary substantially in terms of the number of classes and the size of the training set.

\textbf{TSB-UAD anomaly benchmark} \citep{tsb-uad} is a recent collection of 1980 univariate time series with labeled anomalies from 18 anomaly detection datasets proposed over the past decade. This collection includes both synthetic and real-world time series originating from a wide range of sources such as the human body, spaceships, environment, and web serves.

\textbf{Minimizing data contamination using careful train-test splitting.} We carefully split each dataset into disjoint training, validation, and test splits, based on splits specified by data creators. When these splits are not available, we randomly sample 60\% of the data for training, 10\% for validation, and 30\% for testing. Long-horizon forecasting and anomaly detection datasets are typically long time series, which are split horizontally as shown in Fig.~\ref{fig:data-split}. Conversely, short-horizon forecasting and classification datasets often contain multiple short time series. For these datasets, a complete time series is either training, validation, or testing. We use the same random seed, set to 13, throughout our experiments, from pre-training to downstream evaluation, thus ensuring that \texttt{MOMENT} only observes the training splits of datasets during pre-training.

\subsection{Model Architecture}
\label{sec:model_architecture}

\begin{figure}[!htb]
\centering
\includegraphics[width=\linewidth]{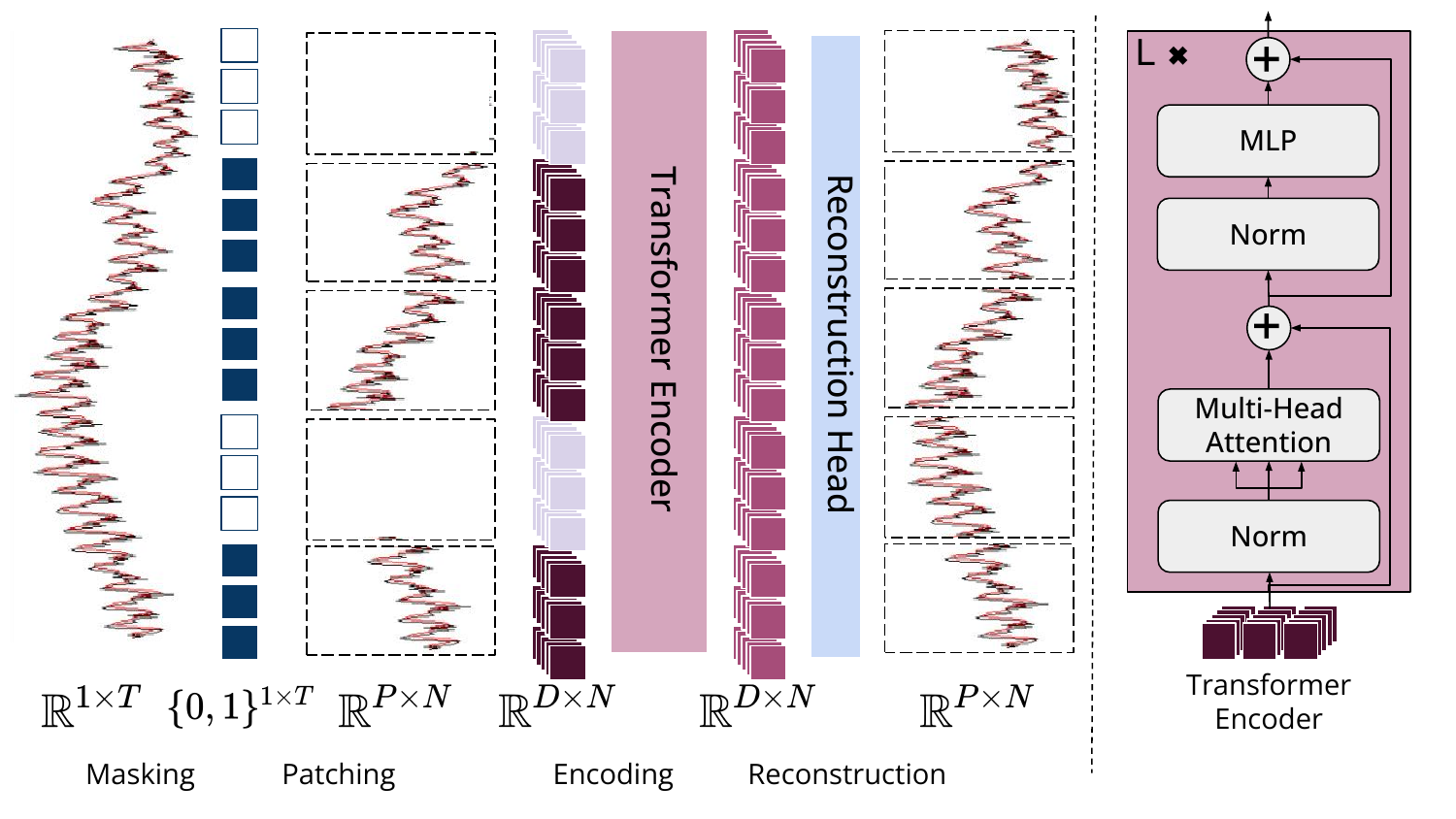}
\caption{\textbf{Overview of \texttt{MOMENT}}. A time series is broken into disjoint fixed-length sub-sequences called patches, and each patch is mapped into a $D$-dimensional patch embedding. During pre-training, we mask patches uniformly at random by replacing their patch embeddings using a special mask embedding \texttt{[MASK]}. The goal of pre-training is to learn patch embeddings which can be used to reconstruct the input time series using a light-weight reconstruction head.}
\label{fig:MOMENT-overview}
\end{figure}

\texttt{MOMENT} receives a univariate time series $\mathcal{T} \in \mathbb{R}^{1 \times T}$, and a mask $M = \{0, 1\}^{1 \times T}$ of length $T$. 0 and 1 denote unobserved and observed time-stamps, respectively. Reversible instance normalization \citep{reversible-instance-normalization} is applied to the observed time series before breaking it into $N$ disjoint patches of length $P$. Each patch is then mapped to a $D$-dimensional embedding, using a trainable linear projection if all time steps are observed, and a designated learnable mask embedding \texttt{[MASK]} $\in \mathbb{R}^{1 \times D}$, otherwise. These $N$ \textit{patch embeddings} serve as input to the transformer model which retains their shape (1 $\times D$) throughout its operations. Each transformed patch embedding is then used to reconstruct both masked and unmasked time series patches, using a \textit{lightweight} prediction head. The goal of the prediction head is to map the transformed patch embeddings to the desired output dimensions. Since this particular prediction head enables time series reconstruction, we call it the \textit{reconstruction head}. Fig. \ref{fig:MOMENT-overview} shows an overview of our model.

Our transformer encoder retains the modifications proposed by \citet{unified-text-to-text-transformer} to the original Transformer \citep{attention-is-all-you-need}. Specifically, we remove the additive bias from the Layer Norm \citep{layer-norm}, and place it before the residual skip connections \citep{deep-residual-learning}, and use the relative positional embedding scheme \citep{relative_position_embedding}. Below we summarize the intuition behind some of our key design decisions.

\textbf{Handling varying time series characteristics.} Time series vary in length, number of channels, amplitudes, and temporal resolutions. We address variable length by restricting \texttt{MOMENT}’s input to a univariate time series of a fixed length $T =$ 512. As is common practice, we sub-sample longer time series, and pad shorter ones with zeros on the left\footnote{We found a large majority of classification datasets to have time series shorter than 512. Besides, a look-back window of length 512 was found to be sufficient for accurate long-horizon forecasting \citep{patchtst}.}. Moreover, segmenting time series into patches quadratically reduces \texttt{MOMENT}’s memory footprint and computational complexity, and linearly increases the length of time series it can take as input. We handle multi-variate time series by independently operating on each channel along the batch dimension. Like recent studies \citep{one-fits-all, patchtst}, we found that modeling each channel independently is an effective strategy for modeling multivariate time series. Finally, re-scaling and centering time series using reversible instance normalization enables \texttt{MOMENT} to model time series with significantly different temporal distributions \citep{reversible-instance-normalization}. We did not explicitly model the temporal resolution of time series, since this information is often unavailable outside of time series forecasting datasets.

\textbf{Intentionally simple encoder.} Closely following the design of transformers in the language domain allows us to leverage their scalable and efficient implementations (e.g., gradient checkpointing, mixed precision training).

\textbf{Light-weight prediction head.} We use a lightweight prediction head instead of a decoder of the same size as the encoder, to enable the necessary architectural modifications for task-specific fine-tuning of a limited number of trainable parameters while keeping the bulk of parameters and the high-level features learned by the encoder intact.

\textbf{Additional absolute positional embeddings.} In addition to relative positional embeddings, we add absolute sinusoidal positional  embeddings~\cite{attention-is-all-you-need} to each patch\footnote{We initially included absolute positional embeddings by accident, but subsequent experiments showed that combining absolute and relative positional embeddings improve predictions.}. 

\begin{table*}[t!]
\centering
\resizebox{\linewidth}{!}{
\begin{tabular}{cccccc}
\toprule
\textbf{Tasks} & \textbf{Supervision} & \textbf{Datasets} & \textbf{Metrics} & \textbf{Baselines} & \textbf{Experimental Setting} \\ \midrule
\begin{tabular}[c]{@{}c@{}}Long-horizon \\ Forecasting\end{tabular} & Linear Probing & \begin{tabular}[c]{@{}c@{}}ETT-h1/h2/m1/m2,\\ Electricity, Traffic,\\ Weather, Exchange, ILI\end{tabular} & MSE, MAE & \begin{tabular}[c]{@{}c@{}}Time-LLM, GPT4TS,\\ TimesNet, PatchTST, FEDFormer,\\ DLinear, N-BEATS,\\ Stationary, LightTS \end{tabular} & \begin{tabular}[c]{@{}c@{}}Look-back window $L = 512$,\\ Forecast horizon $H = \{24, 60\}$ (ILI), $\{96, 720\}$ (rest)\end{tabular} \\ \midrule
\begin{tabular}[c]{@{}c@{}}Short-horizon \\ Forecasting\end{tabular} & Zero-shot & \begin{tabular}[c]{@{}c@{}}M3 and M4\\ competition\\datasets (subset)\end{tabular} & sMAPE\footnote{The definitions of sMAPE were different in the M3 and M4 competitions. In our experiments, we used the same definition as the M4 competition}. & \begin{tabular}[c]{@{}c@{}}GPT4TS, TimesNet, N-BEATS, \\ AutoARIMA, AutoTheta, AutoETS,\\ Seasonal Naive, Naive, Random Walk\end{tabular} & \begin{tabular}[c]{@{}c@{}}Statistical methods fit on individual time series.\\ Deep learning methods are trained on a source dataset\\\& evaluated on a target dataset of the same temporal resolution. \end{tabular} \\ \midrule
Classification & \begin{tabular}[c]{@{}c@{}}Unsupervised\\ representation\\ learning\end{tabular} & \begin{tabular}[c]{@{}c@{}}UCR Classification\\ Archive\end{tabular} & Accuracy & \begin{tabular}[c]{@{}c@{}}GPT4TS, TimesNet,\\ TS2Vec, T-Loss, TNC, TS-TCC, TST,\\ CNN, Encoder, FCN, MCNN,\\ MLP, ResNet, t-LeNet, TWIESN\\ DTW\end{tabular} & \begin{tabular}[c]{@{}c@{}}All models except \texttt{MOMENT} were trained on each\\ individual dataset. Quality of unsupervised representations\\measured using the accuracy of a SVM trained on them.\end{tabular} \\ \midrule
\begin{tabular}[c]{@{}c@{}}Anomaly\\ Detection\end{tabular} & \begin{tabular}[c]{@{}c@{}}Linear probing,\\ Zero-shot\end{tabular} & \begin{tabular}[c]{@{}c@{}}UCR Anomaly\\ Archive\end{tabular} & \begin{tabular}[c]{@{}c@{}}Adjusted Best F1\\ VUS-ROC\end{tabular} & \begin{tabular}[c]{@{}c@{}}GPT4TS, TimesNet, \\ Anomaly Transformer, DGHL,\\ Anomaly Nearest Neighbor\end{tabular} & \begin{tabular}[c]{@{}c@{}}Reconstruction-based anomaly detection with window size $= 512$\\MSE between observed and predicted time series\\is used as the anomaly criterion\end{tabular} \\ \midrule
Imputation & \begin{tabular}[c]{@{}c@{}}Linear probing,\\ Zero-shot\end{tabular} & \begin{tabular}[c]{@{}c@{}}ETT-h1/h2/m1/m2,\\ Electricity, Weather\end{tabular} & MSE, MAE & \begin{tabular}[c]{@{}c@{}}GPT4TS, TimesNet,\\ Linear, Naive, Cubic Spline, Nearest Neighbors\end{tabular} & \begin{tabular}[c]{@{}c@{}}Randomly mask contiguous sub-sequences of\\ length 8 \\ Masking ratios: $\{12.5\%, 25\%, 37.5\%, 50\%\}$\end{tabular} \\ \toprule
\end{tabular}}
\caption{\textbf{Experimental benchmark.} We evaluate \texttt{MOMENT} on 5 time series analysis tasks with an emphasis on limited memory, compute, and supervision settings.}
\label{tab:experimental-benchmark}
\end{table*}

\subsection{Pre-training using Masked Time series Modeling}

We pre-train \texttt{MOMENT} using the masked time series modeling task. Fig. \ref{fig:MOMENT-overview} presents an overview of our pre-training procedure. During training, we first mask a small number of patches uniformly at random by replacing their patch embeddings with a learnable mask embedding \texttt{[MASK]}. The corrupted time series patches are then fed into the transformer encoder to learn patch representations, which are used to reconstruct the original time series using a lightweight reconstruction head. The pre-training objective is to minimize the \textit{masked reconstruction error i.e.} the Mean Squared Error between the ground truth and the prediction, averaged over the masked patches. 

\textbf{Pre-training Setup.} We pre-train three different sizes of \texttt{MOMENT}, roughly corresponding to the sizes of encoders in \texttt{T5-Small}, \texttt{Base}, and \texttt{Large}. Specifically, the \texttt{Base} (\underline{\texttt{Small}}, \textoverline{\texttt{Large}}) model uses a 12 (\underline{6}, \textoverline{24}) layer Transform with hidden dimensions of size $D =$ 768 (\underline{512}, \textoverline{1024}), 12 (\underline{8}, \textoverline{16}) attention heads, and feed-forward networks of size 3072 (\underline{2048}, \textoverline{4096}), resulting in approximately 125 (\underline{40}, \textoverline{385}) million parameters. All weights are randomly initialized before pre-training. All models take an input time series of length $T =$ 512, breaking it into $N =$ 64 disjoint patches of length $P =$ 8. We mask 30\% of the patches uniformly at random during pre-training.

We use the Adam optimizer with weight decay \citep{decoupled-weight-decay-regularization} with $\lambda=$ 0.05, $\beta_1 =$ 0.9, $\beta_2 =$ 0.999. We clip the gradient at 5.0, train models using a batch size of 2048, and use cosine learning rate schedule with initial and final learning rates of $1e^{-4}$ and $1e^{-5}$, respectively. We use gradient checkpointing \citep{gradient-checkpointing} to improve training throughput and save memory, and train all models in a mixed precision setting, using \texttt{float-32} for numerically unstable operations, e.g. layer normalization, and \texttt{bfloat-16}\footnote{\url{https://cloud.google.com/tpu/docs/bfloat16}}, otherwise. We train all models for 2 epochs. 

\subsection{Fine-tuning on Downstream Tasks}
\label{subsec:finetuning-MOMENT-on-downstream-tasks}

\texttt{MOMENT} can be seamlessly used for multiple time series analysis tasks. In this work, we consider 5 practical time series analysis tasks as examples, namely: long- and short-horizon forecasting, classification, anomaly detection, and imputation. For forecasting tasks with horizon $H$, we replace the reconstruction head with a forecasting head, which first flattens all the $N$ 
$D$-dimensional patch embeddings into a $N \times D$ dimensional vector, and then projects it into a $H$-dimensional time series via a linear projection layer. For all other tasks, we retain the reconstruction head. We provide detailed descriptions of each task and \texttt{MOMENT}'s configuration in App.~\ref{app:experimental-setup-and-results}.

\textbf{Fine-tuning settings.} \texttt{MOMENT} can either be fine-tuned end-to-end, or linear probed ($\mathtt{{MOMENT_{LP}}}$) by freezing all parameters except for those in the reconstruction or forecasting head. Additionally, for some tasks such as anomaly detection, unsupervised representation learning and imputation, \texttt{MOMENT} can also be used in a zero-shot ($\mathtt{{MOMENT_{0}}}$) setting by retaining its reconstruction head. 

\section{Experimental Setup and Results}
\label{sec:results}

\begin{table*}[t!]
\centering
\large
\resizebox{\linewidth}{!}{
\begin{tabular}{cc|cc|cc|cc|cc|cc|cc|cc|cc|cc|cc}
\toprule
\multicolumn{2}{c}{Methods} & \multicolumn{2}{c}{$\mathtt{MOMENT_{LP}}$} & \multicolumn{2}{c}{\textbf{Time-LLM}} & \multicolumn{2}{c}{\textbf{GPT4TS}} & \multicolumn{2}{c}{\textbf{PatchTST}} & \multicolumn{2}{c}{\textbf{DLinear}} & \multicolumn{2}{c}{\textbf{TimesNet}} & \multicolumn{2}{c}{\textbf{FEDFormer}} & \multicolumn{2}{c}{\textbf{Stationary}} & \multicolumn{2}{c}{\textbf{LightTS}} & \multicolumn{2}{c}{\textbf{N-BEATS}} \\
\multicolumn{2}{c}{Metric} & MSE & MAE & MSE & MAE & MSE & MAE & MSE & MAE & MSE & MAE & MSE & MAE & MSE & MAE & MSE & MAE & MSE & MAE & MSE & MAE \\ \midrule
\multirow{4}{*}{Weather} & 96 & 0.154 & 0.209 & - & - & 0.162 & 0.212 & 0.149 & 0.198 & 0.176 & 0.237 & 0.172 & 0.220 & 0.217 & 0.296 & 0.173 & 0.223 & 0.182 & 0.242 & 0.152 & 0.210 \\
 & 192 & 0.197 & 0.248 & - & - & 0.204 & 0.248 & 0.194 & 0.241 & 0.220 & 0.282 & 0.219 & 0.261 & 0.276 & 0.336 & 0.245 & 0.285 & 0.227 & 0.287 & 0.199 & 0.260 \\
 & 336 & 0.246 & 0.285 & - & - & 0.254 & 0.286 & 0.245 & 0.282 & 0.265 & 0.319 & 0.280 & 0.306 & 0.339 & 0.380 & 0.321 & 0.338 & 0.282 & 0.334 & 0.258 & 0.311 \\
 & 720 & 0.315 & 0.336 & - & - & 0.326 & 0.337 & 0.314 & 0.334 & 0.333 & 0.362 & 0.365 & 0.359 & 0.403 & 0.428 & 0.414 & 0.410 & 0.352 & 0.386 & 0.331 & 0.359 \\ \midrule
\multirow{4}{*}{ETTh1} & 96 & 0.387 & 0.410 & 0.408 & 0.429 & 0.376 & 0.397 & 0.370 & 0.399 & 0.375 & 0.399 & 0.384 & 0.402 & 0.376 & 0.419 & 0.513 & 0.491 & 0.424 & 0.432 & 0.399 & 0.428 \\
 & 192 & 0.410 & 0.426 & - & - & 0.416 & 0.418 & 0.413 & 0.421 & 0.405 & 0.416 & 0.436 & 0.429 & 0.420 & 0.448 & 0.534 & 0.504 & 0.475 & 0.462 & 0.451 & 0.464 \\
 & 336 & 0.422 & 0.437 & - & - & 0.442 & 0.433 & 0.422 & 0.436 & 0.439 & 0.443 & 0.491 & 0.469 & 0.459 & 0.465 & 0.588 & 0.535 & 0.518 & 0.488 & 0.498 & 0.500 \\
 & 720 & 0.454 & 0.472 & 0.523 & 0.514 & 0.477 & 0.456 & 0.447 & 0.466 & 0.472 & 0.490 & 0.521 & 0.500 & 0.506 & 0.507 & 0.643 & 0.616 & 0.547 & 0.533 & 0.608 & 0.573 \\ \midrule
\multirow{4}{*}{ETTh2} & 96 & 0.288 & 0.345 & 0.285 & 0.348 & 0.285 & 0.342 & 0.274 & 0.336 & 0.289 & 0.353 & 0.340 & 0.374 & 0.358 & 0.397 & 0.476 & 0.458 & 0.397 & 0.437 & 0.327 & 0.387 \\
 & 192 & 0.349 & 0.386 & - & - & 0.354 & 0.389 & 0.339 & 0.379 & 0.383 & 0.418 & 0.402 & 0.414 & 0.429 & 0.439 & 0.512 & 0.493 & 0.520 & 0.504 & 0.400 & 0.435 \\
 & 336 & 0.369 & 0.408 & - & - & 0.373 & 0.407 & 0.329 & 0.380 & 0.448 & 0.465 & 0.452 & 0.452 & 0.496 & 0.487 & 0.552 & 0.551 & 0.626 & 0.559 & 0.747 & 0.599 \\
 & 720 & 0.403 & 0.439 & 0.399 & 0.435 & 0.406 & 0.441 & 0.379 & 0.422 & 0.605 & 0.551 & 0.462 & 0.468 & 0.463 & 0.474 & 0.562 & 0.560 & 0.863 & 0.672 & 1.454 & 0.847 \\ \midrule
\multirow{4}{*}{ETTm1} & 96 & 0.293 & 0.349 & 0.384 & 0.403 & 0.292 & 0.346 & 0.290 & 0.342 & 0.299 & 0.343 & 0.338 & 0.375 & 0.379 & 0.419 & 0.386 & 0.398 & 0.374 & 0.400 & 0.318 & 0.367 \\
 & 192 & 0.326 & 0.368 & - & - & 0.332 & 0.372 & 0.332 & 0.369 & 0.335 & 0.365 & 0.374 & 0.387 & 0.426 & 0.441 & 0.459 & 0.444 & 0.400 & 0.407 & 0.355 & 0.391 \\
 & 336 & 0.352 & 0.384 & - & - & 0.366 & 0.394 & 0.366 & 0.392 & 0.369 & 0.386 & 0.410 & 0.411 & 0.445 & 0.459 & 0.495 & 0.464 & 0.438 & 0.438 & 0.401 & 0.419 \\
 & 720 & 0.405 & 0.416 & 0.437 & 0.429 & 0.417 & 0.421 & 0.416 & 0.420 & 0.425 & 0.421 & 0.478 & 0.450 & 0.543 & 0.490 & 0.585 & 0.516 & 0.527 & 0.502 & 0.448 & 0.448 \\ \midrule
\multirow{4}{*}{ETTm2} & 96 & 0.170 & 0.260 & 0.181 & 0.269 & 0.173 & 0.262 & 0.165 & 0.255 & 0.167 & 0.269 & 0.187 & 0.267 & 0.203 & 0.287 & 0.192 & 0.274 & 0.209 & 0.308 & 0.197 & 0.271 \\
 & 192 & 0.227 & 0.297 & - & - & 0.229 & 0.301 & 0.220 & 0.292 & 0.224 & 0.303 & 0.249 & 0.309 & 0.269 & 0.328 & 0.280 & 0.339 & 0.311 & 0.382 & 0.285 & 0.328 \\
 & 336 & 0.275 & 0.328 & - & - & 0.286 & 0.341 & 0.274 & 0.329 & 0.281 & 0.342 & 0.321 & 0.351 & 0.325 & 0.366 & 0.334 & 0.361 & 0.442 & 0.466 & 0.338 & 0.366 \\
 & 720 & 0.363 & 0.387 & 0.366 & 0.388 & 0.378 & 0.401 & 0.362 & 0.385 & 0.397 & 0.421 & 0.408 & 0.403 & 0.421 & 0.415 & 0.417 & 0.413 & 0.675 & 0.587 & 0.395 & 0.419 \\ \midrule
\multirow{4}{*}{ILI} & 24 & 2.728 & 1.114 & 3.025 & 1.195 & 2.063 & 0.881 & 1.319 & 0.754 & 2.215 & 1.081 & 2.317 & 0.934 & 3.228 & 1.260 & 2.294 & 0.945 & 8.313 & 2.144 & 4.539 & 1.528 \\
 & 36 & 2.669 & 1.092 & - & - & 1.868 & 0.892 & 1.430 & 0.834 & 1.963 & 0.963 & 1.972 & 0.920 & 2.679 & 1.080 & 1.825 & 0.848 & 6.631 & 1.902 & 4.628 & 1.534 \\
 & 48 & 2.728 & 1.098 & - & - & 1.790 & 0.884 & 1.553 & 0.815 & 2.130 & 1.024 & 2.238 & 0.940 & 2.622 & 1.078 & 2.010 & 0.900 & 7.299 & 1.982 & 4.957 & 1.585 \\
 & 60 & 2.883 & 1.126 & 3.245 & 1.221 & 1.979 & 0.957 & 1.470 & 0.788 & 2.368 & 1.096 & 2.027 & 0.928 & 2.857 & 1.157 & 2.178 & 0.963 & 7.283 & 1.985 & 5.429 & 1.661 \\ \midrule
\multirow{4}{*}{ECL} & 96 & 0.136 & 0.233 & - & - & 0.139 & 0.238 & 0.129 & 0.222 & 0.140 & 0.237 & 0.168 & 0.272 & 0.193 & 0.308 & 0.169 & 0.273 & 0.207 & 0.307 & 0.131 & 0.228 \\
 & 192 & 0.152 & 0.247 & - & - & 0.153 & 0.251 & 0.157 & 0.240 & 0.153 & 0.249 & 0.184 & 0.289 & 0.201 & 0.315 & 0.182 & 0.286 & 0.213 & 0.316 & 0.153 & 0.248 \\
 & 336 & 0.167 & 0.264 & - & - & 0.169 & 0.266 & 0.163 & 0.259 & 0.169 & 0.267 & 0.198 & 0.300 & 0.214 & 0.329 & 0.200 & 0.304 & 0.230 & 0.333 & 0.170 & 0.267 \\
 & 720 & 0.205 & 0.295 & - & - & 0.206 & 0.297 & 0.197 & 0.290 & 0.203 & 0.301 & 0.220 & 0.320 & 0.246 & 0.355 & 0.222 & 0.321 & 0.265 & 0.360 & 0.208 & 0.298 \\ \midrule
\multirow{4}{*}{Traffic} & 96 & 0.391 & 0.282 & - & - & 0.388 & 0.282 & 0.360 & 0.249 & 0.410 & 0.282 & 0.593 & 0.321 & 0.587 & 0.366 & 0.612 & 0.338 & 0.615 & 0.391 & 0.375 & 0.259 \\
 & 192 & 0.404 & 0.287 & - & - & 0.407 & 0.290 & 0.379 & 0.256 & 0.423 & 0.287 & 0.617 & 0.336 & 0.604 & 0.373 & 0.613 & 0.340 & 0.601 & 0.382 & 0.403 & 0.274 \\
 & 336 & 0.414 & 0.292 & - & - & 0.412 & 0.294 & 0.392 & 0.264 & 0.436 & 0.296 & 0.629 & 0.336 & 0.621 & 0.383 & 0.618 & 0.328 & 0.613 & 0.386 & 0.426 & 0.285 \\
 & 720 & 0.450 & 0.310 & - & - & 0.450 & 0.312 & 0.432 & 0.286 & 0.466 & 0.315 & 0.640 & 0.350 & 0.626 & 0.382 & 0.653 & 0.355 & 0.658 & 0.407 & 0.508 & 0.335 \\ \bottomrule
\end{tabular}}
\caption{Long-term forecasting performance measured using Mean Squared Error (MSE) and Mean Absolute Error (MAE). PatchTST performs the best across most settings, closely followed by \texttt{MOMENT}. Complete results in Tab. \ref{tab:long_horizon_forecasting_appendix}.}
\label{tab:long_horizon_forecasting}
\end{table*}

\begin{table*}[t!]
\centering
\resizebox{0.95\linewidth}{!}{
\begin{tabular}{lc|cc|cccccc|ccc|ccc}
\toprule
& &  
\multicolumn{2}{c}{$\mathtt{MOMENT_{LP}}$} & 
\multicolumn{2}{c}{\textbf{GPT4TS}} &
\multicolumn{2}{c}{\textbf{TimesNet}} &
\multicolumn{2}{c}{\textbf{N-BEATS}} &
\multirow{2}{*}{\textbf{ARIMA}} & \multirow{2}{*}{\textbf{Theta}} & \multirow{2}{*}{\textbf{ETS}} & \multirow{2}{*}{\textbf{\begin{tabular}[c]{@{}c@{}}Seasonal\\Naive\end{tabular}}} 
& \multirow{2}{*}{\textbf{Naive}} 
& \multirow{2}{*}{\textbf{\begin{tabular}[c]{@{}c@{}}Random\\Walk\end{tabular}}}
\\ 
\multicolumn{2}{c}{\textbf{Datasets}} & M4 & FR & M4 & FR & M4 & FR & M4 & FR &  &  &  &  &  & \\
\midrule
\multirow{3}{*}{M3} 
 & Yearly & \textbf{16.74} & \textbf{16.97} & 18.39 & \textbf{17.40} & 27.48 & \textbf{16.21} & \textbf{16.82} & \textbf{15.92} & 17.90 & 16.70 & 16.47 & 17.54 & 17.54 & 16.77 \\
 & Quarterly & \textbf{10.09} & 10.62 & \textbf{10.18} & 10.29 & 14.41 & 12.68 & 11.26 & 11.30 & 10.18 & 9.19 & 8.99 & 11.02 & 11.45 & 11.72 \\
 & Monthly & 16.04 & 16.90 & \textbf{15.21} & 16.37 & \textbf{15.58} & 16.23 & \textbf{15.63} & 16.37 & 15.95 & 14.96 & 14.41 & 17.74 & 18.53 & 19.19 \\
 \midrule
\multirow{4}{*}{M4} 
 & Yearly & - & \textbf{14.84} & - & \textbf{14.80} & - & \textbf{14.40} & - & 14.18 & 16.19 & 14.04 & 14.06 & 16.33 & 16.33 & 14.22 \\
 & Quarterly & - & 12.02 & - & 11.77 & - & 13.21 & - & 12.25 & 10.86 & 10.21 & 10.24 & 12.55 & 11.65 & 11.46 \\
 & Monthly & - & 15.80 & - & 15.36 & - & 15.67 & - & 15.24 & 13.68 & 13.19 & 13.58 & 16.00 & 15.24 & 15.48 \\ \bottomrule
\end{tabular}}
\caption{Zero-shot short-horizon forecasting performance on a subset of the M3 and M4 datasets measured using sMAPE. Statistical methods outperformed their deeper counterparts. However, on some datasets (in \textbf{bold}), \texttt{MOMENT}, GPT4TS and N-BEATS achieved a lower sMAPE than ARIMA.} 
\label{table:short-horizon-forecasting}
\end{table*}

We extend the experimental benchmark introduced by \citet{timesnet} across various dimensions. Below, we outline the design choices of our benchmark and highlight its key distinctions from TimesNet\footnote{In this section, we use TimesNet to refer to the benchmark proposed by \citet{timesnet} instead of their model.}.

\textbf{Time series modeling with limited supervision.} Our benchmark comprises of 5 major time series modeling tasks of significant practical value, namely long- and short-horizon forecasting, imputation, classification, and anomaly detection, as outlined in Tab.~\ref{tab:experimental-benchmark}. In contrast to TimesNet, we exclusively consider scenarios characterized by limited compute and supervision resources. These scenarios mimic practical situations where training (or fine-tuning) a deep neural network is infeasible due to resource limitations or insufficiently characterized data. Accordingly, we assess \texttt{MOMENT} in zero-shot settings whenever feasible and through linear probing for a few epochs otherwise.

For classification, we consider the unsupervised representation learning problem, where the goal is to learn representations of time series that are useful for downstream classification, without access to labeled data. As is common in prior work \citep{TS2Vec, unsupervised-scalable-representation-learning-for-multivariate-ts}, the quality of representations is measured using the accuracy of a Support Vector Machine trained on them (App.~\ref{app:classification_app}). For short-horizon forecasting, we consider the zero-shot setting introduced by \citet{meta-learning-zero-shot-ts-forecasting}. In particular, we fine-tune \texttt{MOMENT} on a source dataset using a forecasting head, and evaluate its performance on a target dataset without any fine-tuning (App~\ref{app:zero_shot_short_horizon_forecasting}, Tab.~\ref{tab:short-horizon-forecasting-dataset-settings}). 

\textbf{Datasets.} We use the same datasets as TimesNet for forecasting and imputation. However, for classification and anomaly detection, we conduct experiments on larger and systematically chosen subset of datasets from the UCR classification archive \citep{UCR-Archive} and UCR anomaly archive \citep{ts-anomaly-detection-benchmarks-are-flawed}. Specifically, we run classification experiments on all 91 time series datasets with each time series shorter than 512 time steps (Tab.\ref{tab:classification_results_appendix}). For anomaly detection, while choosing the subset of time series, we prioritized coverage over different domains and data sources represented in the UCR anomaly archive (Tab.~\ref{tab:anomaly_detection_results_appendix}). We also note that the UCR anomaly archive was proposed as an improvement over pre-existing anomaly detection datasets such as the SMD~\citep{smd}, and SMAP~\citep{smap}, many of which are also used in TimesNet. Our proposed experimental setup is summarized in Tab.~\ref{tab:experimental-benchmark} and detailed in App.~\ref{app:experimental-setup-and-results}.

\textbf{Metrics.} We evaluate each experiment using \textit{multiple} metrics used in task-specific benchmarks, such as MSE and MAE for long-horizon forecasting, and sMAPE for short-horizon forecasting. We also note that TimesNet and GPT4TS~\citep{one-fits-all} evaluate anomaly detection performance using vanilla $F_1$ score which ignores the sequential nature of time series. Instead, we measure anomaly detection performance with the widely used adjusted best $F_1$ score \citep{goswami2023unsupervised, dghl}, and the recently proposed VUS-ROC~\citep{VUS-ROC}. 

\textbf{Baselines.} We compare \texttt{MOMENT} with state-of-the-art deep learning and statistical machine learning models across tasks (Tab.~\ref{tab:results-sources}). This is in contrast to TimesNet which primarily compared with transformer-based approaches. These comparisons are crucial for assessing the practical utility of the proposed methods. We found that statistical and non-transformer-based approaches like ARIMA for short-horizon forecasting, N-BEATS for long-horizon forecasting, and $k$-nearest neighbors for anomaly detection outperform many deep and transformer-based models.

\paragraph{Hyper-parameter tuning.} We do not perform hyper-parameter tuning. In all experiments that follow, unless mentioned otherwise, we fine-tune \texttt{MOMENT-Large} with a batch size of 64, and one cycle learning rate schedule with a peak learning rate between $5e-5$ and $1e-3$ \citep{one-cycle-lr}. For baseline methods, we capture recommended settings from their papers and public repositories. We report all hyper-parameters settings for \texttt{MOMENT} and baselines in App.~\ref{app:experimental-setup-and-results}.

\textbf{Research questions.} Through the following experiments we aim to answer 3 broad research questions. 

\textbf{RQ1: Effectiveness.} Is \texttt{MOMENT} effective for multiple time series analysis tasks in limited supervision settings?

\textbf{RQ2: Interpretability.} What is \texttt{MOMENT} learning? Does it capture intuitive time series characteristics such as varying frequencies, trends, and amplitudes?

\textbf{RQ3: Properties.} What is the impact of the size of scaling model size? Can \texttt{MOMENT}, akin to LLMs, be used for cross-modal transfer learning?

\begin{table*}[t!]
\centering
\resizebox{\linewidth}{!}{
\begin{tabular}{c|c|cc|ccccc|cccccccc|c}
\toprule
\textbf{} & $\mathtt{MOMENT_0}$ & \textbf{TimesNet} & \textbf{GPT4TS} & \textbf{TS2Vec} & \textbf{T-Loss} & \textbf{TNC} & \textbf{TS-TCC} & \textbf{TST} & \textbf{CNN} & \textbf{Encoder} & \textbf{FCN} & \textbf{MCNN} & \textbf{MLP} & \textbf{ResNet} & \textbf{t-LeNet} & \textbf{TWIESN} & \textbf{DTW} \\ \midrule
Mean & 0.794 & 0.572 & 0.566 & 0.851 & 0.833 & 0.786 & 0.793 & 0.658 & 0.751 & 0.743 & 0.809 & 0.702 & 0.750 & 0.825 & 0.348 & 0.726 & 0.764 \\
Median & 0.815 & 0.565 & 0.583 & 0.871 & 0.849 & 0.788 & 0.802 & 0.720 & 0.773 & 0.753 & 0.837 & 0.718 & 0.766 & 0.852 & 0.333 & 0.724 & 0.768 \\ \midrule
Std. & 0.147 & 0.238 & 0.234 & 0.134 & 0.136 & 0.168 & 0.176 & 0.220 & 0.180 & 0.159 & 0.188 & 0.194 & 0.169 & 0.177 & 0.221 & 0.164 & 0.152 \\ \midrule
Mean rank & 7.2 & 13.3 & 13.3 & 3.5 & 5.3 & 6.9 & 6.5 & 11.8 & 9.4 & 8.9 & 5.6 & 11.0 & 9.2 & 4.4 & 16.1 & 10.4 & 9.1 \\
Median rank & 7.0 & 14.0 & 14.0 & 3.0 & 5.0 & 6.5 & 6.0 & 13.0 & 9.0 & 9.0 & 4.0 & 12.0 & 10.0 & 3.0 & 17.0 & 11.0 & 9.0 \\ \midrule
Wins & 880 & 325 & 326 & 1220 & 1033 & 885 & 924 & 460 & 684 & 727 & 1031 & 533 & 696 & 1137 & 71 & 593 & 712 \\
Losses & 566 & 1121 & 1121 & 227 & 375 & 522 & 484 & 987 & 763 & 719 & 415 & 914 & 750 & 310 & 1375 & 854 & 734 \\ \bottomrule
\end{tabular}}
\caption{\textbf{Classification accuracy} of methods across 91 UCR datasets. Methods with mean and median accuracy higher than \texttt{MOMENT} are in \textbf{bold}. \texttt{MOMENT} without fine-tuning on individual datasets demonstrates promising accuracy. Complete results in Tab.~\ref{tab:classification_results_appendix}. Results on multi-variate UEA datasets are in Tab.~\ref{tab:multivariate_classification_results_appendix}.}
\label{tab:classification}
\end{table*}

\subsection{\texttt{MOMENT} can solve multiple time series modeling tasks in limited supervision settings}

\textbf{Long-horizon forecasting.} Linearly probing \texttt{MOMENT} achieves near state-of-the-art performance on most datasets and horizons, and is only second to PatchTST which generally achieves the lowest MSE (Tab.~\ref{tab:long_horizon_forecasting}). On many datasets and horizons, forecasting models based on LLMs-- TimeLLM and GPT4TS perform worse than \texttt{MOMENT}. Notably, N-BEATS outperforms several recent methods, emphasizing the importance of comparing forecasting performance beyond transformer-based approaches. 

\begin{figure*}[!bt]
\centering
\setlength{\tabcolsep}{0pt}
\begin{tabular}{ccccc}
\includegraphics[width=0.19\linewidth]{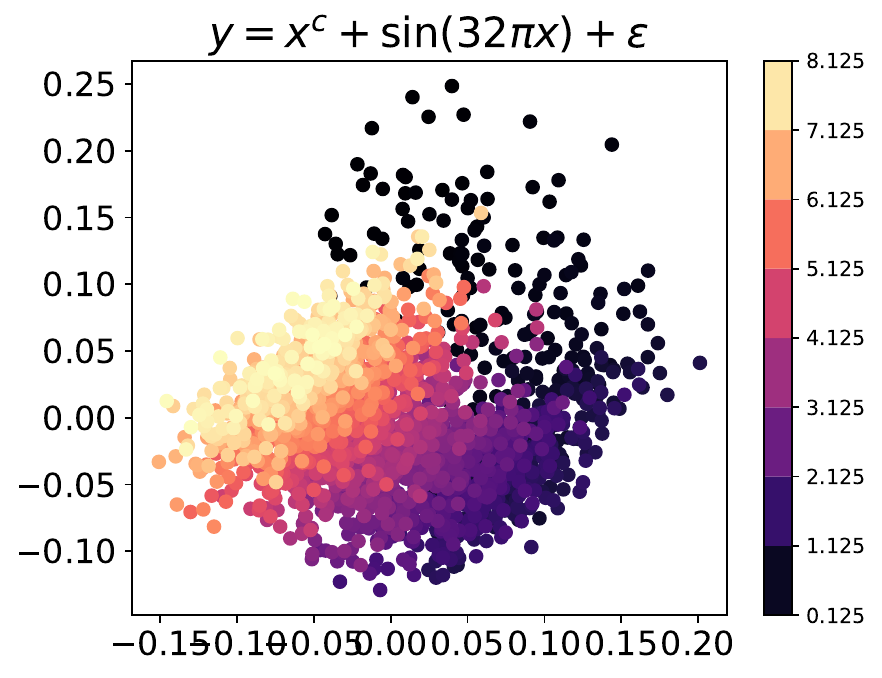} & 
\includegraphics[width=0.19\linewidth]{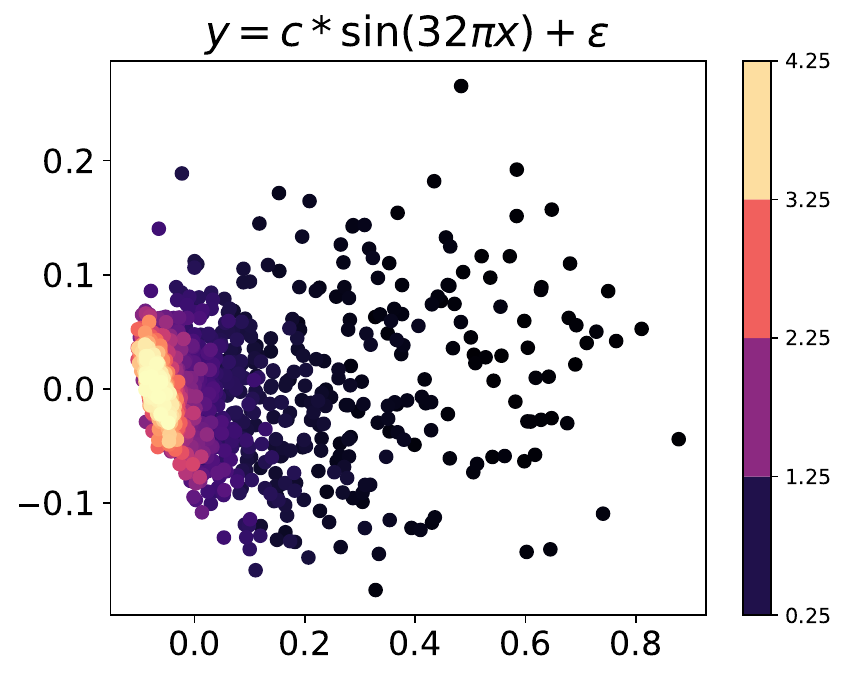} & 
\includegraphics[width=0.19\linewidth]{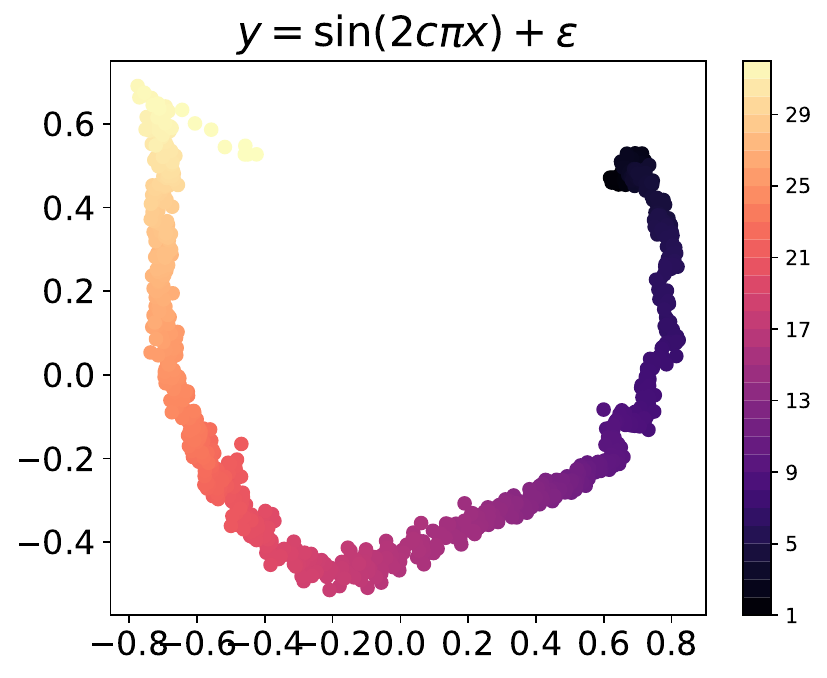} & 
\includegraphics[width=0.19\linewidth]{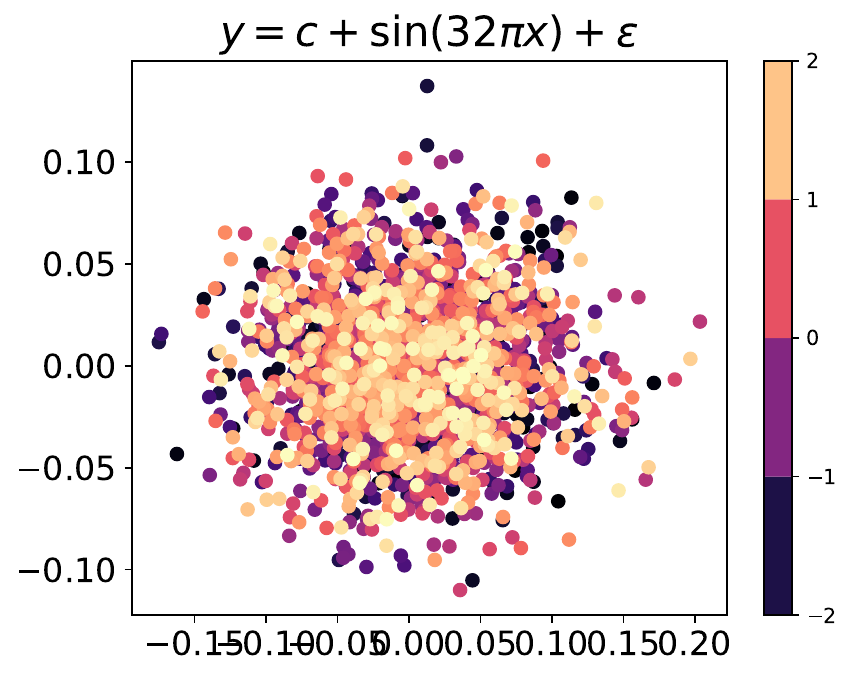} & 
\includegraphics[width=0.19\linewidth]{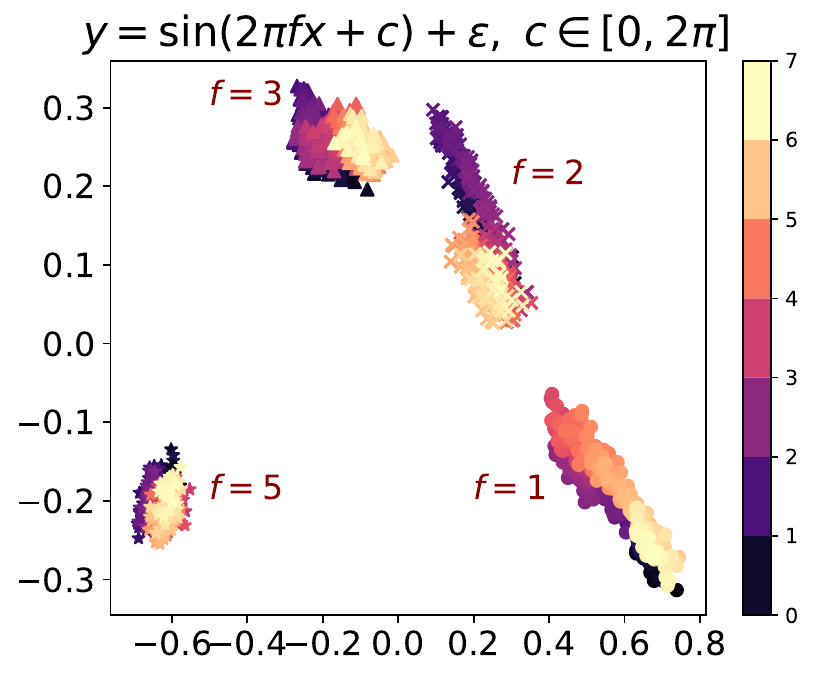} \\
\textit{(i)} Trend & 
\textit{(ii)} Amplitude & 
\textit{(iii)} Frequency & 
\textit{(iv)} Baseline Shift & 
\textit{(v)} Phase 
\end{tabular}
\caption{What is \texttt{MOMENT} learning? Principal components of the embeddings of synthetically generated sinusoids suggest that \texttt{MOMENT} can capture subtle trend, scale, frequency, and phase information. In each experiment, $c$ controls the factor of interest, for example the power of the trend polynomial $c \in (\frac{1}{8}, 8)$ \citep{N-BEATS} (Fig.~\ref{fig:interpretability_timeseries}), and frequency $c \in (1, 32)$ of the generated sine waves (Fig.~\ref{fig:interpretability_timeseries}). We generate multiple sine waves by varying $c$, derive their sequence-level representations using \texttt{MOMENT}, and visualize them in a 2- dimensional space using PCA and t-SNE \citep{tsne} in Fig.~\ref{fig:interpretability} and Fig.~\ref{fig:interpretability_appendix}.}
\label{fig:interpretability}
\end{figure*}

\textbf{Zero-shot short-horizon forecasting.} 
Among all tasks, we found zero-shot short-horizon forecasting to have the largest scope for improvement (Tab.~\ref{table:short-horizon-forecasting}). Statistical methods such as Theta and ETS outperformed their deeper counterparts. However, on some datasets, \texttt{MOMENT} achieved lower sMAPE than ARIMA. 

\textbf{Classification.} Without any data-specific fine-tuning, \texttt{MOMENT} can learn distinct representations for different classes of data (Fig.~\ref{fig:clustering_ucr}), and an SVM trained on its representations performs better than all but 4 methods specifically built for time series classification models and trained on each individual dataset. Recently proposed GPT4TS and TimesNet perform poorly despite being trained on each individual dataset with labels. 

\begin{table}
\centering
\resizebox{\linewidth}{!}{
\begin{tabular}{c|c|cc|c}
 \toprule
 \textbf{Model} & \textbf{Bit Memory} & \textbf{MNIST} & \textbf{CIFAR-10} & \textbf{IMDb} \\ \midrule
 GPT-2 & \textbf{1.000} & 0.975 & \textbf{0.711} & 0.867 \\
 Flan-T5 & \textbf{1.000} & \textbf{0.987} & 0.672 & 0.861 \\
 \texttt{MOMENT} & \textbf{1.000} & 0.982 & 0.620 & \textbf{0.872} \\\bottomrule
\end{tabular}}
\caption{\textbf{Cross-modal transfer experiments.} Accuracy measured on the test set, from the checkpoint with the lowest train loss. Even with frozen self-attention and feed-forward layers, \texttt{MOMENT} is able to model cross-modal sequences on par with GPT-2 and Flan-T5 models of similar scale.}
\label{tab:cross-modal-transfer}
\end{table}

\textbf{Anomaly detection.} On 44 time series from the UCR anomaly detection archive, \texttt{MOMENT} consistently outperformed both TimesNet and GPT4TS, as well as 2 state-of-the-art deep learning models tailored for anomaly detection, in both zero-shot and linear probing configurations. However, $k$-nearest neighbors performed marginally better in terms of VUS-ROC score, but had a lower adjusted best $F_1$ score.

\begin{table*}[t!]
\centering
\resizebox{\linewidth}{!}{
\begin{tabular}{c|cccc|cccc|cccccccc}
\toprule
\multirow{2}{*}{\textbf{Dataset}} & \multicolumn{2}{c}{$\mathtt{MOMENT_{0}}$} & \multicolumn{2}{c}{$\mathtt{MOMENT_{LP}}$} & \multicolumn{2}{c}{\textbf{GPT4TS}} & \multicolumn{2}{c}{\textbf{TimesNet}} & \multicolumn{2}{c}{\textbf{Naive}} & \multicolumn{2}{c}{\textbf{Linear}} & \multicolumn{2}{c}{\textbf{Nearest}} & \multicolumn{2}{c}{\textbf{Cubic}} \\
 &  MSE & MAE & MSE & MAE & MSE & MAE & MSE & MAE & MSE & MAE & MSE & MAE & MSE & MAE & MSE & MAE \\ \midrule
Weather & 0.082 & 0.130 & 0.035 & 0.075 & \textbf{0.031} & \textbf{0.071} & 0.036 & 0.098 & 0.119 & 0.108 & 0.065 & 0.067 & 0.083 & 0.078 & 0.601 & 0.153 \\ 
ETTh1 & 0.402 &	0.403 & \textbf{0.139} & \textbf{0.234} & 0.227 & 0.254 & 0.175 & 0.264 & 1.185 & 0.658 & 0.775 & 0.534 & 0.900 & 0.579 & 2.178 & 0.916 \\ 
ETTh2 & 0.125 & 0.238 & \textbf{0.061} & \textbf{0.159} & 0.109 & 0.213 & 0.170 & 0.286 & 0.225 & 0.304 & 0.135 & 0.234 & 0.166 & 0.252 & 1.920 & 0.641 \\ 
ETTm1 & 0.202 & 0.288 & \textbf{0.074} & \textbf{0.168} & 0.076 & 0.146 & 0.087 & 0.198 & 0.455 & 0.365 & 0.165 & 0.229 & 0.230 & 0.260 & 0.858 & 0.494 \\ 
ETTm2 & 0.078 & 0.184 & \textbf{0.031} & \textbf{0.108} & 0.052 & 0.133 & 0.112 & 0.220 & 0.113 & 0.191 & 0.062 & 0.138 & 0.079 & 0.152 & 0.534 & 0.356 \\ 
Electricity & 0.250 & 0.371 & 0.094 & 0.211 & \textbf{0.072} & \textbf{0.183} & 0.124 & 0.248 & 1.474 & 0.869 & 0.737 & 0.592 & 0.923 & 0.629 & 2.257 & 0.888 \\ \bottomrule
\end{tabular}}
\caption{\textbf{Imputation Results.} \texttt{MOMENT}
with linear probing achieved the lowest reconstruction error on all ETT datasets. In the zero-shot setting, MOMENT consistently outperformed all statistical interpolation methods with the exception of linear interpolation. Complete results in Tab.~\ref{tab:imputation_appendix}.}
\label{tab:imputation}
\end{table*}

\begin{table*}[t!]
\centering
\resizebox{0.9\linewidth}{!}{
\begin{tabular}{c|c|cc|cc|cc|c}
\toprule
\multicolumn{2}{c}{\textbf{Metric}} & \textbf{$\mathtt{MOMENT_0}$} & \textbf{$\mathtt{MOMENT_{LP}}$} & \textbf{\textbf{GPT4TS}} & \textbf{TimesNet} & \textbf{\begin{tabular}[c]{@{}c@{}}Anomaly\\ Transformer\end{tabular}} & \textbf{DGHL} & \textbf{$k$-NN} \\ \midrule
\multirow{6}{*}{Adj. $F_1$} & Mean & 0.585 & \textbf{0.628} & 0.424 & 0.537 & 0.492 & 0.425 & 0.554 \\
 & Median & 0.683 & \textbf{0.778} & 0.331 & 0.541 & 0.432 & 0.331 & 0.595 \\ 
 & Std. & 0.377 & 0.373 & 0.366 & 0.389 & 0.401 & 0.365 & 0.393 \\ 
 & Mean rank & 3.410 & \textbf{3.005} & 4.862 & 3.642 & 4.326 & 5.071 & 3.681 \\ 
 & Median rank & \textbf{3.00} & \textbf{3.00} & 5.00 & 3.50 & 4.00 & 5.25 & 3.75 \\ 
 & Wins/Losses & 703/472 & 783/393 & 419/757 & 658/518 & 524/652 & 378/798 & 650/525 \\ \midrule
\multirow{6}{*}{VUS ROC} & Mean & 0.670 & 0.684 & 0.611 & 0.679 & 0.661 & 0.646 & \textbf{0.706} \\
 & Median & 0.677 & 0.692 & 0.615 & 0.692 & 0.658 & 0.635 & \textbf{0.727} \\ 
 & Std. & 0.133 & 0.146 & 0.114 & 0.141 & 0.147 & 0.137 & 0.155 \\ 
 & Mean rank & 4.056 & 3.382 & 5.193 & 3.897 & 3.913 & 4.403 & \textbf{3.153} \\
 & Median rank & 4.00 & \textbf{3.00} & 6.00 & 4.00 & 4.00 & 4.50 & 3.00 \\ 
 & Wins/Losses & 577/599 & 709/467 & 354/822 & 608/568 & 605/571 & 509/667 & 754/422 \\ \bottomrule
\end{tabular}
}
\caption{Anomaly detection performance averaged over 248 time series from the UCR Anomaly Archive. $\mathtt{MOMENT_{LP}}$ achieves near state-of-the-art anomaly detection results. Complete results in Tab.~\ref{tab:anomaly_detection_results_appendix}.}
\label{tab:anomaly-detection}
\end{table*}

\textbf{Imputation.} Tab.~\ref{tab:imputation} contains imputation performance of all models averaged over 4 different masking rates. \texttt{MOMENT} with linear probing achieved the lowest reconstruction error on all ETT datasets. In the zero-shot setting, \texttt{MOMENT} consistently outperformed all statistical interpolation methods with the exception of linear interpolation.

\subsection{What is \texttt{MOMENT} Learning?}
We found that \texttt{MOMENT} can capture changes in intuitive time series characteristics such as trend, amplitude, frequencies, and phases of time series. However, it cannot differentiate between vertically shifted time series as it normalizes each signal prior to modeling (Fig.~\ref{fig:interpretability},\ref{fig:interpretability_appendix}). Furthermore, on many classification datasets, \texttt{MOMENT} learns distinct representations of different classes, even in a zero-shot setting without access to labels (Fig.~\ref{fig:clustering_ucr}, \ref{fig:clustering_ucr_appendix}).

\subsection{Properties of Large Time Series Models}
\label{sec:moment_properties}

\begin{figure}[!t]
\begin{minipage}[t]{0.48\textwidth}
\centering
\setlength{\tabcolsep}{0pt}
\begin{tabular}{ccc}
\raisebox{0.25\height}{\includegraphics[trim={0.5cm 0.5cm 0.5cm 0}, width=0.33\linewidth]{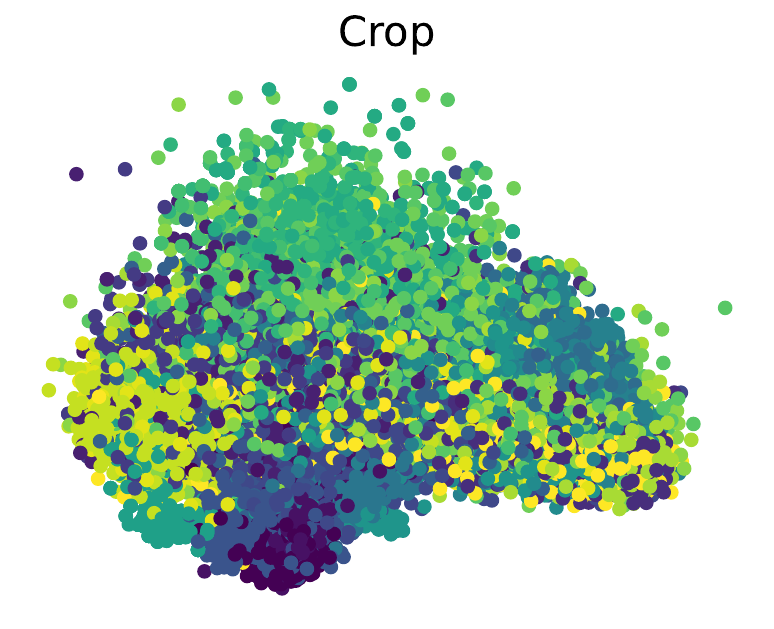}} &
\raisebox{0.25\height}{\includegraphics[trim={0.8cm 0.5cm 0.8cm 0}, width=0.33\linewidth]{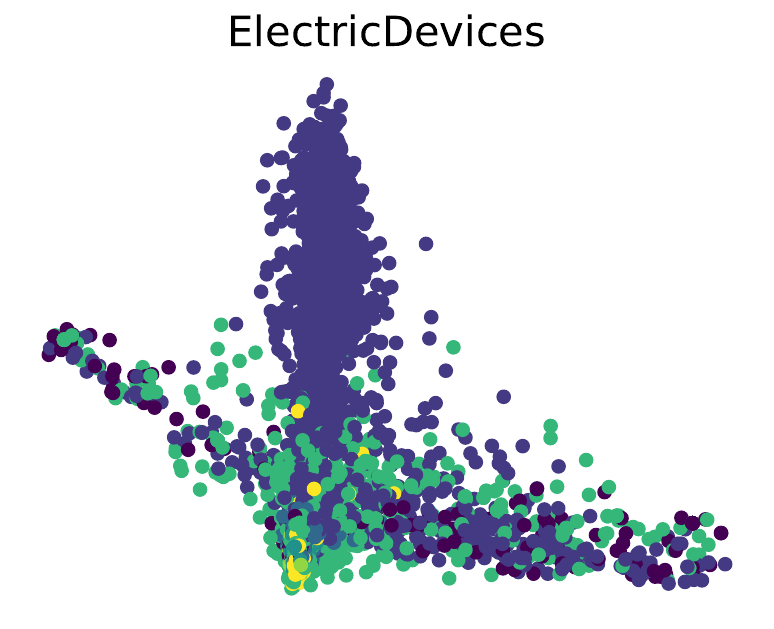}} &
\raisebox{0.25\height}{\includegraphics[trim={0.8cm 0.5cm 0.8cm 0}, width=0.33\linewidth]{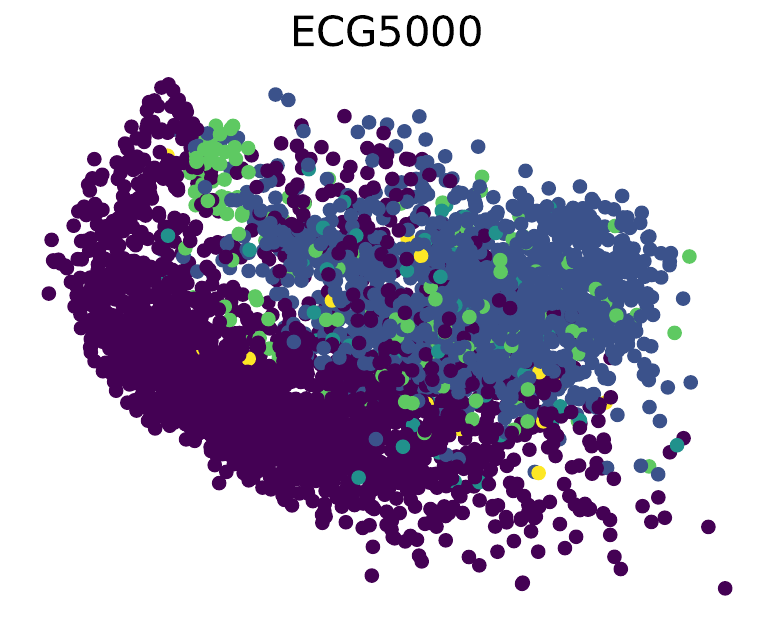}} \\
\end{tabular}
\caption{PCA and t-SNE visualizations of representations learned by \texttt{MOMENT} on the 3 largest UCR datasets. Different colors represent different classes. Even without dataset-specific fine-tuning, \texttt{MOMENT} learns distinct representations for different classes.}
\label{fig:clustering_ucr}
\end{minipage}
\hfill
\begin{minipage}[t]{0.5\textwidth}
\centering
\setlength{\tabcolsep}{0pt}
\begin{tabular}{cc}
\includegraphics[width=0.5\textwidth]{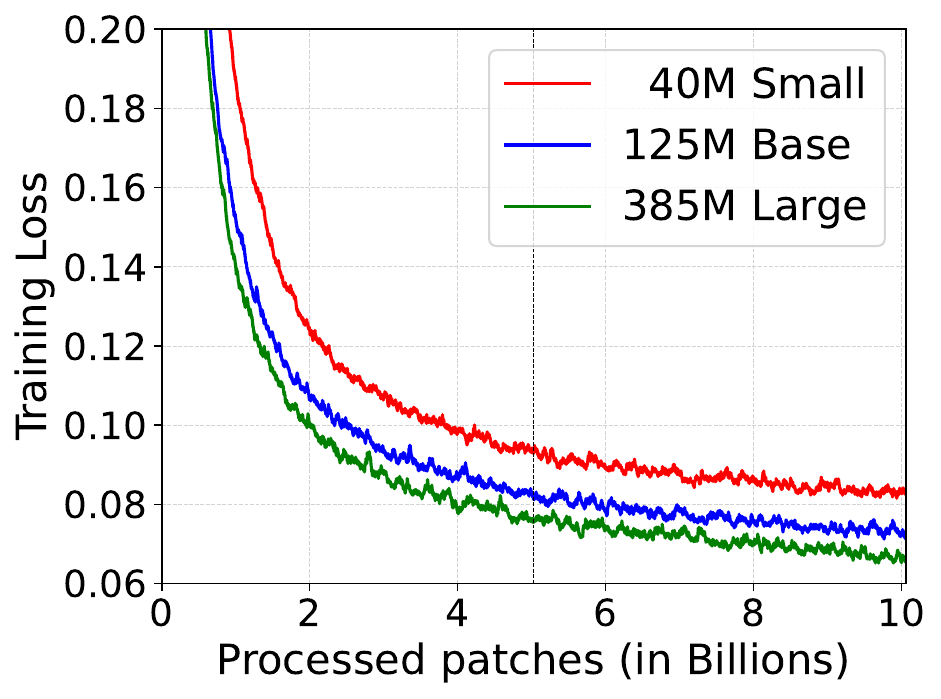} &
\includegraphics[width=0.49\textwidth]{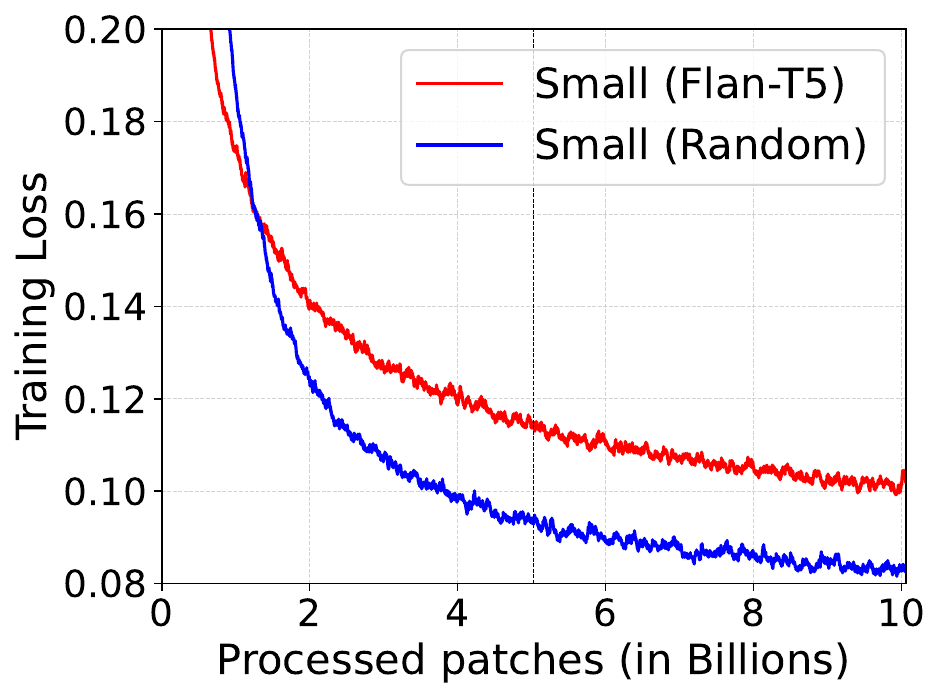} \\
\end{tabular}
\caption{\textbf{Training losses (MSE).} A dashed vertical line denotes the first epoch. All models were trained with a batch size of 131072 patches. (\textit{left}) Larger models obtain lower training loss. \textit{right} Eventually, randomly initialized \texttt{MOMENT-small} outperforms the same model initialized with Flan-T5 weights.}
\label{fig:training_loss}
\end{minipage}
\end{figure}

\textbf{Model scaling improves training loss.} Like LLMs, we found that increasing the size of the model leads to lower training loss, even before the first epoch (Fig.~\ref{fig:training_loss}, left). An immediate next step is to assess how effectively this phenomenon extends to time series modeling tasks under limited supervision.

\textbf{\texttt{MOMENT} can solve cross-modal sequence learning tasks.} \citet{pretrained-transformers-as-universal-computation-engines} first showed that large pre-trained language and vision transformers can solve general sequence learning tasks for modalities outside of text and images with minimal fine-tuning. Several recent studies have leveraged these properties to reprogram LLMs for time series tasks. We explore whether transformers pre-trained on time series can also be used to solve sequence classification tasks on image, text, and binary data. Our results confirm that by freezing the self-attention and feed-forward layers, \texttt{MOMENT} can model sequences comparable to GPT-2 and Flan-T5 models of similar scale (Tab.~\ref{tab:cross-modal-transfer}).

\textbf{\texttt{MOMENT} with randomly initialized weights converges to a lower training loss.} Our observations suggest that with sufficient data, pre-training our model from scratch results in a lower training loss than continually pre-training a model of similar size initialized with language modeling weights (Fig.~\ref{fig:training_loss}, \ref{fig:training_losses_appendix}). This also underscores that there is sufficient publicly accessible pre-training data available in the Time Series Pile to facilitate pre-training time series foundation models from scratch. 

\section{Conclusion and Future Work}
We release the first open-source family of time series foundation models and make contributions at all stages of the development and evaluation process. We first compile a large and diverse collection of public time series, called the Time Series Pile, and demonstrate its efficacy by pre-training high-performing time series foundation models from scratch. Then, we systematically address several time series-specific challenges, which up to now have impeded widespread exploration of extensivelarge-scale multi-dataset pre-training.

We use the Time Series Pile and these strategies to pre-train transformer models of three different sizes. Finally, we design an experimental benchmark to evaluate time series foundation models on multiple practical time series tasks, particularly focusing on scenarios with constrained compute and supervision, building on prior work by \citet{timesnet}. Using this benchmark, we show that \texttt{MOMENT} is effective for the considered tasks with minimal fine-tuning. \texttt{MOMENT}'s superior performance, especially on anomaly detection and classification problems which typically have small datasets, can be attributed to pre-training. Moreover, we demonstrate that across many tasks, smaller statistical and shallower deep learning methods perform reasonably well. Lastly, we make several interesting empirical observations about time series foundation models. Our overarching goal is to push the boundaries of open science by publicly releasing the Time Series Pile, along with code, model weights, and training logs.

We note several interesting directions of future work, including the application of \texttt{MOMENT} to real-world challenges, investigating multi-modal time series and text foundation models \citep{jolt}, and enhancing forecasting performance by pre-training \texttt{MOMENT} using causal attention and forecasting objectives.

\section*{Acknowledgments}

\textbf{Funding.} This work was partially supported by the National Institutes of Health (NIH) under awards R01HL144692 and 1R01NS124642-01, and also by the U.S. Army Research Office and the U.S. Army Futures Command under Contract No. W911NF-20-D-0002.  The content of the information does not necessarily reflect the position or the policy of the government and no official endorsement should be inferred. 

\textbf{Discussions.} We would like to express our sincerest gratitude to Bar{\i}\c{s} Kurt, Andrey Kan, Laurent Callot, Gauthier Guinet, Jingchao Ni, and Jonas M. K\"ubler for insightful discussions regarding the problem setting and experimental design. Their unwavering support was instrumental in the development of \texttt{MOMENT}. We are also thankful to Laurent, Bar{\i}\c{s}, Jingchao and Andrey for their constructive feedback on the writing of this manuscript. Additionally, we acknowledge the insightful exchanges with Yuyang (Bernie) Wang, Abdul Fatir Ansari, Ingo Guering, Xiyuan Zhang, and Anoop Deoras. Special thanks to Cherie Ho for suggesting a creative and befitting name for our model. Lastly, we would like to thank Cecilia Morales for her insightful comments, especially on the broader impacts of this work, and for helping us proofread this manuscript. Finally, we would like to thank the anonymous ICML reviewers for their comments which helped improve our work.

\textbf{Data.} We extend our gratitude to the authors and data curators whose meticulous efforts were instrumental in curating the datasets utilized for both pre-training and evaluation purposes: UCR Time Series Classification Archive \citep{UCR-Archive}, TSB-UAD Anomaly Benchmark \citep{tsb-uad}, Monash Forecasting Archive \citep{monash}, and the long-horizon forecasting datasets \citep{Informer}.

\paragraph{Software and Models.} Our training and evaluation library was inspired from \href{https://github.com/thuml/Time-Series-Library}{\texttt{Time-Series-Library}}. We would also like to thank the authors of the following libraries for their implementations: \href{https://github.com/kzl/universal-computation}{\texttt{universal-computation}}, \href{https://github.com/thuml/Anomaly-Transformer}{\texttt{Anomaly-Transformer}}, \href{https://github.com/TheDatumOrg/VUS}{\texttt{VUS}}, \href{https://github.com/mononitogoswami/tsad-model-selection}{\texttt{tsad-model-selection}}, \href{https://github.com/DAMO-DI-ML/NeurIPS2023-One-Fits-All}{\texttt{One-Fits-All}} and \href{https://github.com/Nixtla/statsforecast}{\texttt{Statsforecast}}~\citep{statsforecast}.

\section*{Reproducibility statement}

All models were trained and evaluated on a computing cluster consisting of 128 AMD EPYC 7502 CPUs, 503 GB of RAM, and 8 NVIDIA RTX A6000 GPUs each with 49 GiB RAM. All \texttt{MOMENT} variants were trained on a single A6000 GPU (with any data or model parallelism). We have made \texttt{MOMENT-large}\footnote{\url{https://huggingface.co/AutonLab/MOMENT-1-large}} and the Time Series Pile\footnote{\url{https://huggingface.co/datasets/AutonLab/Timeseries-PILE}} publicly available on Huggingface. We are working on open-sourcing \texttt{MOMENT-base} and \texttt{MOMENT-small}, and our research code public. The latter is currently available at \url{https://github.com/moment-timeseries-foundation-model/moment-research}. We enlist an exhaustive list of hyper-parameters in App.~\ref{app:experimental-setup-and-results} to aid reproducibility. We would like to emphasize that all datasets used in this study are publicly available.

\section*{Impact statement}
\textbf{Transparency Index.} 
Given the exponential rise in societal reliance on large foundation models, ensuring transparency in their training approach, architecture, and downstream application is crucial for public accountability, scientific advancement, and effective governance. o uphold this objective, we publicly release our training code base, data sources, and evaluation pipeline. We assess the transparency of \texttt{MOMENT} using the criteria outlined by \citet{transparency-index}, focusing on upstream resources utilized during training and model description, encompassing 32 and 33 transparency indicators, respectively. We report expected upstream and model transparency scores for \texttt{MOMENT} in Tab.~\ref{tab:transparency-index-model}. Notably, \texttt{MOMENT} is \textit{expected} to have one of the highest levels of upstream transparency. However, it's model transparency scores are lower, primarily due to comprehensive (external and third-party) harm and trustworthiness evaluations, which are not well understood in the context of time series modeling.

\textbf{Environmental Impact.} We train multiple models over many days resulting in significant energy usage and a sizeable carbon footprint. However, we hope that releasing our models will ensure that future time series modeling efforts are quicker and more efficient, resulting in lower carbon emissions.

We follow prior work \citep{can-lms-be-too-big,carbon-emissions-and-large-nn-training,llama2,sustainable-ai,measuring-carbon-intensity-of-ai} and estimate the carbon footprint of pre-training all variants of \texttt{MOMENT} based on the GPU device used and the carbon efficiency of the electricity grid. Our estimated $\mathtt{CO_2}$ generation estimates are shown in Tab.~\ref{tab:carbon-emissions}.

\begin{table*}[ht!]
\centering
\resizebox{0.8\linewidth}{!}{
\begin{tabular}{c|cccc}
\toprule
Model Variant & \# Parameters (M) & GPU Hours & Power Consumption (W) & Carbon Emission (tCO2eq) \\ \midrule
Small & 40 & 308.378 & 300 & 31.136 \\
Base & 125 & 308.306 & 300 & 31.129 \\
Large & 385 & 404.389 & 300 & 40.831 \\ \midrule
Upper Bound Total & - & 1021.073 & 300 & 103.096 \\ \midrule
Actual Total & - & 712.767 & 300 & 71.967 \\ \bottomrule
\end{tabular}
}
\caption{Total carbon emission induced upon training the \texttt{MOMENT} family of models. \texttt{MOMENT-small} and \texttt{MOMENT-base} were trained simultaneously on a single GPU, thus the TGP required for each model would likely be much less than 300W, and the total time for both models combined is equal to the maximum of the time required for each model. Actual total power consumption and carbon emission values account for this.}
\label{tab:carbon-emissions}
\end{table*}

We use the Total Graphics Power (TGP) to calculate the total power consumed for training \texttt{MOMENT} models, although the total power consumed by the GPU will likely vary a little based on the GPU utilization while training our model. Our calculations do not account for power demands from other sources of our compute. We use 336.566 Kg $C0_2$/MWH as the standard value of $CO_2$ emission per megawatt hour of energy consumed for Pittsburgh\footnote{\url{https://emissionsindex.org/}}.

We share an upper limit of the individual $CO_2$ emission for each model, as well as a more realistic actual estimate for the carbon emissions from \texttt{MOMENT-small} and \texttt{MOMENT-base}, since they were trained simultaneously on a single Nvidia RTX A6000 GPU, and thus the power consumed by the GPU was shared for the training of both variants. \texttt{MOMENT-large} was trained independently on a single RTX A6000 GPU.

\textbf{Ethical considerations and potential misuse.} Despite \texttt{MOMENT}'s promising performance in limited-data settings, it's important to use its predictions with care, especially in high-stakes settings like healthcare. The reliability of \texttt{MOMENT}'s predictions hinges on the quality and diversity of the data it was trained on. Any dependence on unreliable or biased data could potentially result in skewed outputs, unfairly portraying or discriminating against certain groups or individuals. While \texttt{MOMENT} demonstrates the ability to capture changes time-series characteristics, e.g. trend, amplitude, frequency, it struggles to differentiate between vertically shifted time-series as it normalizes signals prior to modeling. Greater interpretability and explainability of \texttt{MOMENT}'s predictions are necessary to enable more informed decision-making processes. Prior to utilizing \texttt{MOMENT} in high-stakes decision-making, we recommend fine-tuning and evaluating the model with task-specific, in-domain data to address these ethical concerns and ensure equitable outcomes. Additionally, efforts should be made to promote equitable access to \texttt{MOMENT} and other AI technologies, particularly within healthcare systems where disparities in access may exacerbate existing inequities.

\bibliography{icml2024_sections/BB_bibliography}
\bibliographystyle{icml2024}


\appendix
\onecolumn

\section{Changelog}
\begin{itemize}
    \item (Oct'2024) Added links to both research and inference benchmarks, notes on contemporary work in Tab.~\ref{tab:contemporary_work} (Timer \cite{timer} and TOTEM \cite{totem}). We also note differences in positional embeddings (Sec.~\ref{sec:model_architecture}). 
\end{itemize}

\section{Related Work}

\textbf{Transformers and Patching for Time series Modeling.} There is a growing body of work utilizing transformers for various time series analysis tasks, for example PatchTST \citep{patchtst}, Informer \citep{Informer}, Autoformer \citep{autoformer}, FEDformer \citep{fedformer}, Pyraformer \citep{pyraformer} for forecasting; Anomaly Transformer \citep{anomaly-transformer} for anomaly detection, and TST \citep{transformer-multivariate-ts-representation-learning}, TS-TCC \citep{ts-tcc} for representation learning. 

One issue with applying transformers to time series data is the complexity of the self-attention mechanism, which grows quadratically with the size of input tokens (or length of time series). Consequently, the primary focus of most initial applications of transformers to time series, especially for forecasting where longer look-back windows typically improve performance, was to redesign the self-attention mechanism to reduce its complexity \citep{Informer,fedformer,pyraformer}. 
\citet{patchtst} demonstrated that treating time series sub-sequences (or patches) as tokens instead of individual time points is a simple, efficient yet effective mechanism for learning useful representations for forecasting. The authors drew inspiration from language and vision domains where sub-words (vs. characters) \citep{bert} and 2-D patches (vs. raw pixels) \citep{beit,transformers-for-image-recognition-at-scale} are used as inputs to transformers. Drawing inspiration from prior work, we build on top of the transformer architecture which takes disjoint time series sub-sequences (or patches) as input.

\textbf{Masked Representation Learning.} Masked pre-training is a widely-used self-supervised learning task where a model learns to accurately reconstruct masked portions of its input. Masked language \citep{bert,unified-text-to-text-transformer} and image modeling \citep{SimMIM,scaling-lang-image-pretraining-via-masking} have been successfully utilized to learn models from vast quantities of unlabeled data, which can generalize to a variety of downstream tasks. 

For time series data, prior work has primarily focused on contrastive representation learning \citep{TS2Vec,ts-tcc,unsupervised-scalable-representation-learning-for-multivariate-ts}. The goal of contrastive learning is to learn a representation space where ``positive" pairs of time series are close while ``negative" pairs are far apart. However, the notion of positive and negative pairs is subjective and data-dependent, and popular transformations such as flipping and cropping invariance may not be appropriate for time series data \citep{TS2Vec}. In contrast, some studies mask portions of time series using zeros and learn a model to reconstruct them \citep{patchtst,transformer-multivariate-ts-representation-learning,simmtm,ti-mae}. 

Representation learning via masking is well-suited to all the downstream tasks we care about, especially forecasting and imputation, as they are instances of the masked reconstruction problem.
Owing to its simplicity and success in vision and language domains, we use the masked prediction task to pre-train our model, using a special embedding (see \texttt{[MASK]} in Fig.~\ref{fig:MOMENT-overview}) to mask time series patches instead of zeros.

\textbf{Cross-modal transfer learning using language models.} \citet{pretrained-transformers-as-universal-computation-engines} had first shown that transformers pre-trained on text data (LLMs) can effectively solve sequence modeling tasks in other modalities. Some recent studies have leveraged this inherent ability of language pre-trained transformers to ``reprogram" LLMs for time series analysis using parameter efficient fine-tuning and suitable tokenization strategies \citep{one-fits-all, llmtime, timellm, tempo, tinytimemixers}. However, some of these models \citep{timellm, llmtime} with billions of parameters demand significant memory and computational resources to perform well. We complement this line of research with three empirical observations (Sec~\ref{sec:moment_properties}): we show that (1) transformers trained on time series can also model sequences across modalities, (2) during pre-training, randomly initializing weights lead to lower pre-training loss, than initializing with language modeling weights, and (3) models pre-trained on time series outperform LLM-based models such as \citep{one-fits-all, timellm} on many tasks and datasets. 

\textbf{Unanswered Questions.} To the best of our knowledge, two questions remain largely unanswered in prior work on time series modeling. First, all existing time series models are (pre-)trained and fine-tuned on individual datasets \citep{patchtst,TS2Vec,timesnet,one-fits-all}, and the benefits (or drawbacks) of large-scale multi-dataset pre-training remains unexplored \citep{transformers-in-ts-survey}. Second, there is very limited work on time series modeling in limited supervision settings, such as zero-shot forecasting \citep{meta-learning-zero-shot-ts-forecasting}, or few-shot classification \citep{Meta-learning-few-shot-ts-classification}. In our work, we consider both these questions and \textit{show that pre-training a model of sufficient capacity on a large corpus of unlabeled time series data can in fact enable it to provide reasonably accurate predictions in limited-supervision settings.}

\section{Contemporary Work}

\begin{table}[hbt!]
\centering
\resizebox{\linewidth}{!}{
\begin{tabular}{rr|cccccc}
\toprule
\multicolumn{2}{r}{\textbf{Feature}} & \textbf{MOMENT} & \textbf{Moirai} & \textbf{Lag-LLaMa} & \textbf{Chronos} & \textbf{Timer} \\ \midrule
\multicolumn{2}{r|}{\textbf{Citation}} & Ours & \citep{moirai} & \citep{lagllama} & \citep{chronos} & \citep{timer} \\
\multicolumn{2}{r|}{\textbf{Base Architecture}} & T5 encoder & Encoder-only transformer & LLaMa & T5 (encoder-decoder) & Decoder-only \\
\multicolumn{2}{r|}{\textbf{Sizes}} & 40M, 125M, 385M & 14M, 91M, 311M & 200M & 20M, 46M, 200M, 710M & 29M, 50M, 67M \\
\multicolumn{2}{r|}{\textbf{Open Source}} & $\checkmark$ & $\checkmark$ & $\checkmark$ & $\checkmark$ & $\checkmark$ \\
\multicolumn{2}{r|}{\textbf{Evaluation Tasks}} & \begin{tabular}[c]{@{}c@{}}Forecasting, Classification, \\ Anomaly detection, Imputation\end{tabular} & Forecasting & Forecasting & Forecasting & \begin{tabular}[c]{@{}c@{}}Forecasting, Imputation, \\ Anomaly detection\end{tabular} \\ \midrule
\multirow{4}{*}{\textbf{Data}} & \textbf{Total Observations} & 1.13B++ & 27.65B\footnote{These values are taken from Tables 8--17 of \citep{moirai} and denote the \# of observations} & 0.36B\footnote{These values are taken from Table 4 of \citep{lagllama} and denote the \# of observations} & 84B & 28B \\
 & \textbf{\# Time Series} & 13M & 4M & 11K & 11M & 7.4M\\
 & \textbf{Public} & $\checkmark$ & $\checkmark$ & $\checkmark$ & $\checkmark$ & $\checkmark$ \\ 
 & \textbf{Public Release} & May'24 & Feb'2024 & Feb'2024 & Mar'2024 & Jun'2024 \\ \midrule
\multicolumn{2}{r|}{\textbf{Tokenization}} & Fixed-length patches  & Multi-scale patches & Lag features & Scaling \& quantization & Single-series Sequence \\
 \multicolumn{2}{r|}{\textbf{Objective}}
& \begin{tabular}[c]{@{}c@{}}Reconstruction error (MSE) \end{tabular}
& \begin{tabular}[c]{@{}c@{}}Forecast log likelihood \\ of mixture distribution\end{tabular}
& \begin{tabular}[c]{@{}c@{}}Forecast log likelihood \\ of Student's \textit{t} distribution\end{tabular} 
& Forecast Cross-entropy loss & Next token prediction & \\
\multicolumn{2}{r|}{\textbf{Distributional Prediction}}
 & In progress (post hoc) & $\checkmark$ & $\checkmark$ & $\checkmark$ & $\times$ \\
\multicolumn{2}{r|}{\textbf{Interpretability}} & $\checkmark$ & $\times$ & $\times$ & $\checkmark$ & $\times$ \\
\multicolumn{2}{r|}{\textbf{Any-variate Prediction}} & $\checkmark$ (Channel independence) & $\checkmark$ (Flattening) & $\times$ & $\times$ & $\checkmark$ (Uniform sub-sampling)\\
\multicolumn{2}{r|}{\textbf{Context length}} & 512 & 1000--5000 & 1024 & 512 & 1440 \\ 
\midrule
\multirow{2}{*}{\textbf{Scaling}} & \textbf{Model}
 & $\checkmark$ & $\checkmark$ & $\checkmark$ & $\checkmark$ & $\checkmark$\\ 
& \textbf{Data} & $\times$ & $\times$ & $\checkmark$ & $\times$ & $\checkmark$ \\ \midrule
\multicolumn{2}{r|}{\textbf{Impact of Initialization}} & $\checkmark$ & $\times$ & $\times$ & $\checkmark$ & $\times$\\ 
\bottomrule
\end{tabular}}
\bigskip
\resizebox{\linewidth}{!}{
\begin{tabular}{rr|cccc}
\toprule
\multicolumn{2}{r}{\textbf{Feature}} & \textbf{TimesFM} & \textbf{TimeGPT-1} & \textbf{TinyTimeMixers} & \textbf{TOTEM} \\ \midrule
\multicolumn{2}{r|}{\textbf{Citation}} & \citep{timesfm} & \citep{timegpt} & \citep{tinytimemixers} & \citep{totem} \\
\multicolumn{2}{r|}{\textbf{Base Architecture}} &  Decoder-only & Transformer & MLP-Mixer \citep{mlpmixer} & VQ-VAE \citep{vqvae} \\
\multicolumn{2}{r|}{\textbf{Sizes}} & 17M, 70M, 200M & ? & $<$1M & 0.3M \\
\multicolumn{2}{r|}{\textbf{Open Source}} &  $\times$ & $\times$ & $\checkmark$ & $\checkmark$ \\
\multicolumn{2}{r|}{\textbf{Evaluation Tasks}} & Forecasting & Forecasting & Forecasting & \begin{tabular}[c]{@{}c@{}}Forecasting, Classification, \\ Anomaly detection, Imputation\end{tabular} \\ \midrule
\multirow{4}{*}{\textbf{Data}} & \textbf{Total Observations} &  100B & 100B & 244M & N/A \\
 & \textbf{\# Time Series} & ? & ? & N/A & N/A \\
 & \textbf{Public} &  $\times$ & $\times$ & $\checkmark$ & $\checkmark$ \\ 
 & \textbf{Public Release} &  - & - & Apr'2024 & Feb'2024\\ 
 \midrule
\multicolumn{2}{r|}{\textbf{Tokenization}} & Fixed-length patches & ? & Adaptive Patching & VQVAE-based \\
 \multicolumn{2}{r|}{\textbf{Objective}} & Forecasting error (MSE) & ? & Forecasting error (MSE) & \begin{tabular}[c]{@{}c@{}} Reconstruction + \\ Commitment error \end{tabular} \\ 
 \multicolumn{2}{r|}{\textbf{Distributional Prediction}} &  $\times$ & $\checkmark$ (post hoc) & $\times$ & $\checkmark$ \\
\multicolumn{2}{r|}{\textbf{Interpretability}} &  $\checkmark$ & $\times$ & $\times$ & $\times$ \\
\multicolumn{2}{r|}{\textbf{Any-variate Prediction}} & $\times$ & $\checkmark$ (?) & $\checkmark$ (Channel Independence + Mixing) & $\checkmark$ (Flattening)\\
\multicolumn{2}{r|}{\textbf{Context length}} & 512 & ? & 512 & 512\\ 
\midrule
\multirow{2}{*}{\textbf{Scaling}} & \textbf{Model} & $\checkmark$ & $\times$ & $\checkmark$ & $\checkmark$ \\ 
& \textbf{Data} & $\checkmark$ & $\times$ & $\checkmark$ & $\checkmark$\\ \midrule
\multicolumn{2}{r|}{\textbf{Impact of Initialization}} & $\times$ & $\times$ & $\checkmark$ & $\times$ \\ 
\bottomrule
\end{tabular}}
\caption{Time series foundation modeling has a young but growing literature. Latest ArXiv pre-prints of most contemporary work are from \textbf{February 2024} with the exception of TimeGPT-1. None of the contemporary works have evaluated their foundation models in terms of cross modal transfer, environmental impact and transparency. ++ denotes that the Time Series Pile is a living database, is still growing in size and diversity. Moirai \citep{moirai} uses a mixture of Student's \textit{t}, log-normal, negative binomial, and low variance normal distribution.}
\label{tab:contemporary_work}
\end{table}

\section{Interesting directions for future work}
We note some interesting directions for future work:
\begin{itemize}
    \item Study the impact of design choices such as the impact of the choice of the loss function (Huber, $L_1$, $L_2$), patch length $(4, 8)$, and masking percentage $(0.3, 0.6)$ on pre-training loss and time series modeling performance. 
    \item Pre-training data. Two interesting directions include using augmentation and synthetic data \citep{chronos} to improve the quality of pre-training, and looking at tuning dataset mixtures in the Time Series Pile. 
\end{itemize}

\section{The Time Series Pile}
We compiled a large collection of publicly available datasets from diverse domains into the Time Series Pile. It has \textbf{13} unique domains of data, which includes \textbf{20.085} GB worth of \textbf{13M} unique time series and \textbf{1.23} billion timestamps (including channels). 

\begin{table}[!htb]
\centering
\resizebox{0.6\columnwidth}{!}{
\begin{tabular}{r|l}
\toprule
\textbf{Unique domains} & \textbf{Examples of Datasets} \\ \midrule
Healthcare & ECG, EEG, Hospital \\
Human Body & Tongue movement, Finger movement, Muscle Signals \\
Nature & Fish outlines, Flower outlines, River flow \\
Audio & Arabic speech, Japanese Speech, Phonetics \\
Power & Power consumption, Electricity, Home appliance usage \\
Economics & Exchange Rate, Bitcoin, Tourism \\
Traffic & Road Traffic, Pedestrian cross, Line occupancy rate \\
Weather & Temperature, Rain, Wind \\
Facilities & Machine Status, Spacecraft Status, Web traffic \\
Web Services & IOPS, NAB \\
Synthetic & MGAB \\
Sensors & NASA MSL, NASA SMAP \\
Gait & Daphnet \\
\bottomrule
\end{tabular}%
}
\caption{\textbf{The Time Series Pile} covers datasets from 13 distinct domains.}
\end{table}

\begin{table}[htb!]
\resizebox{\columnwidth}{!}{
\begin{tabular}{c|c|c|c|c|c}
\toprule
\textbf{Task} & \textbf{Dataset} & \textbf{Channels} & \textbf{Series Length} & \textbf{Data Size} (Train, Val, Test) & \textbf{Information} (Frequency/Number of Classes) \\ \midrule
\multirow{7}{*}{\begin{tabular}[c]{@{}c@{}}Long\\horizon\\forecasting\\(Informer)\\\end{tabular}} & ETTm1, ETTm2 & 7 & \multirow{6}{*}{\{96, 720\}} & (33953, 11425, 11425) & Electricity (15 mins) \\
 & ETTh1, ETTh2 & 7 &  & (8033, 2785, 2785) & Electricity (15 mins) \\
 & Electricity & 321 &  & (17805, 2537, 5165) & Electricity (Hourly) \\ 
 & Traffic & 862 &  & (11673, 1661, 3413) & Transportation (Hourly) \\
 & Weather & 21 &  & (36280, 5175, 10444) & Weather (10 mins) \\ 
 & Exchange & 8 &  & (4704, 665, 1422) & Exchange rate (Daily) \\ 
 & ILI & 7 & \{24, 60\} & (69, 2, 98) & Illness (Weekly) \\ \midrule
\multirow{6}{*}{\begin{tabular}[c]{@{}c@{}}Short\\horizon\\forecasting\\(Monash)\end{tabular}} & M4-Yearly & \multirow{6}{*}{1} & 6 & (16099, 2301, 4600) & - \\ 
 & M4-Quarterly &  & 8 & (16800, 2400, 4800) & - \\ 
 & M4-Monthly &  & 18 & (33600, 4800, 9600) & - \\ 
 & M3-Yearly &  & 6 & (451, 65, 129) & - \\ 
 & M3-Quarterly &  & 8 & (529, 76, 151) & - \\ 
 & M3-Monthly &  & 18 & (999, 144, 285) & - \\ \hline
\multirow{4}{*}{\begin{tabular}[c]{@{}c@{}}Imputation\\(Informer)\\\end{tabular}} & ETTm1, ETTm2 & 7 & \multirow{4}{*}{512} & (33953, 11425, 11425) & Electricity (15 mins) \\
 & ETTh1, ETTh2 & 7 &  & (8033, 2785, 2785) & Electricity (15 mins) \\ 
 & Electricity & 321 &  & (17805, 2537, 5165) & Electricity (Hourly) \\
 & Weather & 21 &  & (36280, 5175, 10444) & Weather (10 mins) \\ \midrule
\multirow{5}{*}{\begin{tabular}[c]{@{}c@{}}Classification\\(UCR)\end{tabular}} & UWaveGestureLibraryX & \multirow{5}{*}{1} & 315 & (640, 256, 3582) & Motion Gesture (8 classes) \\
 & ECG5000 &  & 140 & (357, 143, 4500) & ECG Record (5 classes) \\
 & OSULeaf &  & 427 & (142, 58, 242) & Leaf Outlines (6 classes) \\
 & MedicalImages &  & 99 & (272, 109, 760) & Pixel Intensity (10 classes) \\ 
 & Ham &  & 431 & (77, 32, 105) & Food spectrographs (2 classes) \\ \midrule
\multirow{5}{*}{\begin{tabular}[c]{@{}c@{}}Anomaly\\detection\\(TSB-UAD)\end{tabular}} & 1sddb40 & \multirow{5}{*}{1} & - & (24489, 9489, 3969) & Beats \\ 
 & BIDMC1 &  & - & (1274, 204, 7988) & PVC \\
 & CIMIS44AirTemperature3 &  & - & (2346, 632, 3672) & Weather Data \\ 
 & CIMIS44AirTemperature5 &  & - & (2346, 632, 3672) & Weather Data \\ 
 & ECG2 &  & - & (10203, 3775, 14488) & ECG2 Lead \\ \bottomrule
\end{tabular}}
\caption{\textbf{Subset of The Time Series Pile used for experiments.} A brief description of datasets that collectively make the Time Series Pile. Due to space constraints, we only include metadata for the subsets of the M3 and M4 datasets in our experiments, as well as 5 classification and anomaly detection datasets. Characteristics of all short-horizon forecasting, classification and anomaly detection datasets in the Time Series Pile can be found in our official repository, and \href{https://forecastingdata.org/}{Monash archive}, \href{https://www.timeseriesclassification.com/}{UCR/UEA classification archive}, and \href{https://github.com/TheDatumOrg/TSB-UAD}{TSB-UAD anomaly benchmark}, respectively.}
\label{tab:time-series-pile}
\end{table}

\begin{table}[htb!]
\centering
\resizebox{\columnwidth}{!}{
\begin{tabular}{ccccccccc}
\toprule
\textbf{Dataset} & \textbf{Domain} & \textbf{\# time series} & \textbf{Min Length} & \textbf{Mean Length} & \textbf{Max Length} & 
\begin{tabular}[c]{@{}c@{}} \textbf{Total} \\ \textbf{Observations}\end{tabular}
& \begin{tabular}[c]{@{}c@{}} \textbf{Average \# of} \\ \textbf{anomalies}\end{tabular} & 
\begin{tabular}[c]{@{}c@{}} \textbf{Average \# of} \\ \textbf{abnormal points}\end{tabular} \\ \midrule
Dodger & Traffic & 1 & 50399 & 50399 & 50399 & 50399 & 133 & 5612 \\
ECG & Healthcare & 74 & 5000 & 311189.82 & 10828799 & 23028047 & 195.6 & 15634 \\
IOPS & Web Services & 29 & 7577 & 100648.89 & 149160 & 2918818 & 46.5 & 2312.3 \\
KDD21 & Multiple & 250 & 1000 & 21215.04 & 250000 & 5303761 & 1 & 196.5 \\
MGAB & Synthetic & 10 & 99999 & 99999 & 99999 & 999990 & 10 & 200 \\
NAB & Web Services & 58 & 1125 & 6300.72 & 22693 & 365442 & 2 & 575.5 \\
SensorScope & Weather & 23 & 20732 & 27037.43 & 30492 & 621861 & 11.2 & 6110.4 \\
YAHOO & Web Services & 367 & 740 & 1560.21 & 1679 & 572599 & 5.9 & 10.7 \\
NASA-MSL & Sensors & 27 & 438 & 2158.88 & 4307 & 58290 & 1.33 & 286.3 \\
NASA-SMAP & Sensors & 54 & 311 & 2554.62 & 2880 & 137950 & 1.26 & 1032.4 \\
Daphnet & Gait & 45 & 9599 & 21759 & 55039 & 979155 & 7.6 & 2841 \\
GHL & Sensors & 126 & 200000 & 200000 & 200000 & 25200000 & 1.2 & 388.8 \\
Genesis & Sensors & 6 & 16219 & 16219 & 16219 & 97314 & 3 & 50 \\
MITDB & Healthcare & 32 & 649999 & 649999 & 649999 & 20799968 & 210.1 & 72334.3 \\
OPP & Sensors & 465 & 22229 & 31615.85 & 51115 & 14701373 & 2 & 1267.3 \\
Occupancy & Sensors & 7 & 2664 & 4688.85 & 9751 & 32822 & 18.3 & 1414.5 \\
SMD & Web Services & 281 & 23686 & 25561.30 & 28742 & 7182727 & 10.4 & 900.2 \\
SVDB & Healthcare & 115 & 230399 & 230399 & 230399 & 26495885 & 208 & 27144.5 \\ \midrule
\end{tabular}}
\caption{\textbf{Anomaly detection datasets in Time Series Pile taken from \citep{tsb-uad}}. The total number of observations is \textbf{129,546,401}, while the total number of unique time series is \textbf{1,970}.}
\end{table}

\begin{table}[hbt!]
\centering
\resizebox{\columnwidth}{!}{
\begin{tabular}{ccccccccc}
\toprule
\textbf{Dataset} & \textbf{Domain} & \textbf{\# Time Series} & \textbf{Min Length} & \textbf{Mean Length} & \textbf{Max Length} & \textbf{Total Observations} & \textbf{Frequency} & \multicolumn{1}{c}{\textbf{Forecast Horizon}} \\ \midrule
Australian Electricity Demand & Power & 5 & 230736 & 231052.8 & 232272 & 1155264 & Half Hourly &  \\ 
Bitcoin & Economics & 18 & 2659 & 4186.88 & 4581 & 75364 & Daily &  \\
Car Parts & Economics & 2674 & 51 & 51 & 51 & 136374 & monthly &  \\ 
CIF 2016 & Economics & 72 & 28 & 98.72 & 120 & 7108 & monthly &  \\ 
Covid Deaths & Nature & 266 & 212 & 212 & 212 & 56392 & Daily &  \\ 
Dominick & Economics & 115704 & 28 & 165.01 & 393 & 19092987 & weekly &  \\ 
\midrule
\multirow{2}{*}{Electricity} & \multirow{2}{*}{Power} & 321 & 26304 & 26304 & 26304 & 8443584 & hourly &  \\
 &  & 321 & 156 & 156 & 156 & 50076 & weekly & \multicolumn{1}{c}{8} \\ \midrule
FRED MD & Economics & 107 & 728 & 728 & 728 & 77896 & monthly &  \\
Hospital & Healthcare & 767 & 84 & 84 & 84 & 64428 & monthly &  \\ 
\midrule
\multirow{2}{*}{Kaggle Web Traffic} & \multirow{2}{*}{Web Services} & 145063 & 803 & 803 & 803 & 116485589 & Daily & \multicolumn{1}{c}{59} \\
 &  & 145063 & 114 & 114 & 114 & 16537182 & weekly & \multicolumn{1}{c}{8} \\ \midrule
KDD Cup 2028 & Nature & 270 & 9504 & 10897.64 & 10920 & 2942364 & hourly &  \\ 
London Smart Meters & Power & 5560 & 288 & 29951.24 & 39648 & 166528896 & half\_hourly &  \\
\midrule
\multirow{3}{*}{M1} & \multirow{3}{*}{Multiple} & 617 & 48 & 90.75 & 150 & 55998 & monthly & \multicolumn{1}{c}{18} \\
 &  & 203 & 18 & 48.98 & 114 & 9944 & quarterly & \multicolumn{1}{c}{8} \\
 &  & 181 & 15 & 24.94 & 58 & 4515 & yearly & \multicolumn{1}{c}{6} \\ \midrule
\multirow{4}{*}{M3} & \multirow{4}{*}{Multiple} & 1428 & 66 & 117.34 & 144 & 167562 & monthly & \multicolumn{1}{c}{18} \\
 &  & 174 & 71 & 76.58 & 104 & 13325 & \multicolumn{1}{l}{} & \multicolumn{1}{c}{8} \\ 
 &  & 756 & 24 & 48.94 & 72 & 37004 & quarterly & \multicolumn{1}{c}{8} \\
 &  & 645 & 20 & 28.40 & 47 & 18319 & yearly & \multicolumn{1}{c}{6} \\ \midrule
\multirow{6}{*}{M4} & \multirow{6}{*}{Multiple} & 4227 & 107 & 2371.38 & 9933 & 10023836 & daily & \multicolumn{1}{c}{14} \\
 &  & 414 & 748 & 901.86 & 1008 & 373372 & hourly & \multicolumn{1}{c}{48} \\
 &  & 48000 & 60 & 234.30 & 2812 & 11246411 & monthly & \multicolumn{1}{c}{18} \\
 &  & 24000 & 24 & 100.25 & 874 & 2406108 & quarterly & \multicolumn{1}{c}{8} \\
 &  & 359 & 93 & 1035.03 & 2610 & 371579 & weekly & \multicolumn{1}{c}{13} \\
 &  & 23000 & 19 & 37.32 & 841 & 858458 & yearly & \multicolumn{1}{c}{6} \\ \midrule
\multirow{2}{*}{NN5} & \multirow{2}{*}{Economics} & 111 & 791 & 791 & 791 & 87801 & daily & \multicolumn{1}{c}{56} \\
 &  & 111 & 113 & 113 & 113 & 12543 & weekly & \multicolumn{1}{c}{8} \\ \midrule
Pedestrian Counts & Traffic & 66 & 576 & 47459.78 & 96424 & 3132346 & hourly &  \\ 
Rideshare & Traffic & 2304 & 541 & 541 & 541 & 1246464 & hourly &  \\
Saugeenday & Nature & 1 & 23741 & 23741 & 23741 & 23741 & daily &  \\ \midrule
\multirow{3}{*}{Solar} & \multirow{3}{*}{Power} & 137 & 52560 & 52560 & 52560 & 7200720 & 10\_minutes &  \\
 &  & 1 & 7397222 & 7397222 & 7397222 & 7397222 & 4\_seconds &  \\
 &  & 137 & 52 & 52 & 52 & 7124 & weekly & \multicolumn{1}{c}{5} \\ \midrule
Sunspot & Nature & 1 & 73931 & 73931 & 73931 & 73931 & daily &  \\ 
Temperature Rain & Nature & 32072 & 725 & 725 & 725 & 23252200 & daily &  \\ \midrule
\multirow{3}{*}{Tourism} & \multirow{3}{*}{Economics} & 366 & 91 & 298.57 & 333 & 109280 & monthly & \multicolumn{1}{c}{24} \\
 &  & 427 & 30 & 99.63 & 130 & 42544 & quarterly & \multicolumn{1}{c}{8} \\
 &  & 518 & 11 & 24.62 & 47 & 12757 & yearly & \multicolumn{1}{c}{4} \\ \midrule
\multirow{2}{*}{Traffic} & \multirow{2}{*}{Traffic} & 862 & 17544 & 17544 & 17544 & 15122928 & hourly &  \\
 &  & 862 & 104 & 104 & 104 & 89648 & weekly & \multicolumn{1}{c}{8} \\ \midrule
US Births & Nature & 1 & 7305 & 7305 & 7305 & 7305 & daily &  \\ 
Vehicle Trips & Traffic & 329 & 70 & 128.82 & 243 & 42382 & daily &  \\ 
Weather & Weather & 3010 & 1332 & 14296.34 & 65981 & 43032000 & daily &  \\ 
Wind & Power & 1 & 7397147 & 7397147 & 7397147 & 7397147 & 4\_seconds &  \\ 
Wind Farms & Power & 339 & 6345 & 507899.88 & 527040 & 172178060 & minutely & \\ \bottomrule
\end{tabular}}
\caption{\textbf{Short horizon forecasting datasets in Time Series Pile} taken from \citep{monash}. The total number of observations is \textbf{281,326,601}, while the total number of unique time series is \textbf{559,102}.}
\end{table}

\begin{table}[!htb]
\centering
\resizebox{\columnwidth}{!}{%
\begin{tabular}{ccccccccc}
\toprule
\textbf{Dataset} & \textbf{Domain} & \textbf{\# Time Series} & \textbf{Min Length} & \textbf{Mean Length} & \textbf{Max Length} & \textbf{Total Observations} & \textbf{Frequency} & \textbf{Forecast Horizon} \\ \midrule
\multirow{5}{*}{FRED} & \multirow{5}{*}{Economics} & 464 & 791 & 1570.08 & 25289 & 728520 & Daily & 14 \\
 &  & 188028 & 19 & 34.76 & 339 & 6535989 & Yearly & 6 \\
 &  & 1630 & 94 & 1261.12 & 5396 & 2055634 & Weekly & 13 \\
 &  & 133411 & 60 & 302.49 & 3867 & 40356476 & Monthly & 18 \\
 &  & 58908 & 24 & 120.11 & 1288 & 7075821 & Quarterly & 8 \\ \bottomrule
\end{tabular}
}
\caption{\textbf{Short horizon forecasting dataset in Time Series Pile taken from the Federal Reserve Economic Data [\href{https://fred.stlouisfed.org/}{(FRED)}] \citep{meta-learning-zero-shot-ts-forecasting}}. The total number of observations is \textbf{56,752,440}, while the total number of unique time series is \textbf{382,441}.}
\end{table}

\begin{table}[thb!]
\centering
\resizebox{\columnwidth}{!}{
\begin{tabular}{ccccccc}
\toprule
\textbf{Dataset} & \textbf{Domain} & \textbf{\# Channels} & \textbf{Time Series Length} & \textbf{Total Observations} & \textbf{Frequency} & \textbf{Forecast Horizon} \\ \midrule
Electricity & Power & 321 & 26304 & 8443584 & Hourly & \{96, 192, 336, 720\} \\
ETTh1 & Power & 7 & 17420 & 121940 & Hourly & \{96, 192, 336, 720\} \\
ETTh2 & Power & 7 & 17420 & 121940 & Hourly & \{96, 192, 336, 720\} \\
ETTm1 & Power & 7 & 69680 & 487760 & 15 Minute & \{96, 192, 336, 720\} \\
ETTm2 & Power & 7 & 69680 & 487760 & 15 Minute & \{96, 192, 336, 720\} \\
Exchange & Finance & 8 & 7588 & 60704 & Daily & \{96, 192, 336, 720\} \\
Illness & Epidemiology & 7 & 966 & 6762 & Weekly & \{24, 36, 48, 60\} \\
Traffic & Traffic & 862 & 17544 & 15122928 & Hourly & \{96, 192, 336, 720\} \\
Weather & Weather & 21 & 52696 & 1106616 & 10 Minute & \{96, 192, 336, 720\} \\ \bottomrule
\end{tabular}}
\caption{\textbf{Long-horizon forecasting datasets in the Time Series Pile taken from \citet{Informer}}. The total number of observations is \textbf{25,959,994}, while the total number of unique time series is \textbf{1,247}.}
\end{table}

\newpage

\footnotesize
\begin{longtable}{wc{3.9cm}|wc{1.7cm}|wc{1cm}|wc{0.8cm}|wc{1cm}|wc{1.3cm}|wc{1.6cm}|wc{1.7cm}}
\toprule
\textbf{Dataset} & \textbf{Domain} & \textbf{\begin{tabular}[c]{@{}c@{}}\# Time-\\ series\end{tabular}} & \textbf{\begin{tabular}[c]{@{}c@{}}\#\\ Classes\end{tabular}} & \textbf{\begin{tabular}[c]{@{}c@{}}\#\\ Channels\end{tabular}} & \textbf{\begin{tabular}[c]{@{}c@{}}Time Series\\ Length\end{tabular}} & \textbf{\begin{tabular}[c]{@{}c@{}}\# Time Series\\ (w. channels)\end{tabular}} & \textbf{\begin{tabular}[c]{@{}c@{}}Total\\ Observations\end{tabular}} \\ \midrule
SemgHandGenderCh2 & Human Body & 900 & 2 & 1 & 1500 & 900 & 1350000 \\
GestureMidAirD2 & Human Body & 338 & 26 & 1 & 360 & 338 & 121680 \\
UWaveGestureLibraryAll & Human Body & 4478 & 8 & 1 & 945 & 4478 & 4231710 \\
SelfRegulationSCP1 & Healthcare & 561 & 2 & 6 & 896 & 3366 & 3015936 \\
UWaveGestureLibraryX & Human Body & 4478 & 8 & 1 & 315 & 4478 & 1410570 \\
GesturePebbleZ2 & Human Body & 304 & 6 & 1 & 455 & 304 & 138320 \\
Healthcare5000 & Healthcare & 5000 & 5 & 1 & 140 & 5000 & 700000 \\
OSULeaf & Nature & 442 & 6 & 1 & 427 & 442 & 188734 \\
MedicalImages & Healthcare & 1141 & 10 & 1 & 99 & 1141 & 112959 \\
Haptics & Human Body & 463 & 5 & 1 & 1092 & 463 & 505596 \\
LargeKitchenAppliances & Power & 750 & 3 & 1 & 720 & 750 & 540000 \\
JapaneseVowels & Audio & 640 & 9 & 12 & 26 & 7680 & 199680 \\
Worms & Nature & 258 & 5 & 1 & 900 & 258 & 232200 \\
Ham & Facilities & 214 & 2 & 1 & 431 & 214 & 92234 \\
DistalPhalanxTW & Nature & 539 & 6 & 1 & 80 & 539 & 43120 \\
ProximalPhalanxOutlineCorrect & Nature & 891 & 2 & 1 & 80 & 891 & 71280 \\
SemgHandMovementCh2 & Human Body & 900 & 6 & 1 & 1500 & 900 & 1350000 \\
RefrigerationDevices & Power & 750 & 3 & 1 & 720 & 750 & 540000 \\
FreezerRegularTrain & Facilities & 3000 & 2 & 1 & 301 & 3000 & 903000 \\
PigAirwayPressure & Nature & 312 & 52 & 1 & 2000 & 312 & 624000 \\
TwoLeadECG & Healthcare & 1162 & 2 & 1 & 82 & 1162 & 95284 \\
GunPointMaleVersusFemale & Human Body & 451 & 2 & 1 & 150 & 451 & 67650 \\
Trace & Power & 200 & 4 & 1 & 275 & 200 & 55000 \\
SmoothSubspace & Generated & 300 & 3 & 1 & 15 & 300 & 4500 \\
MiddlePhalanxTW & Nature & 553 & 6 & 1 & 80 & 553 & 44240 \\
AtrialFibrillation & Healthcare & 30 & 3 & 2 & 640 & 60 & 38400 \\
SyntheticControl & Generated & 600 & 6 & 1 & 60 & 600 & 36000 \\
ShapesAll & Generated & 1200 & 60 & 1 & 512 & 1200 & 614400 \\
Human BodyVerticalSignal & Human Body & 724 & 12 & 1 & 1250 & 724 & 905000 \\
PLAID & Facilities & 1074 & 11 & 1 & 1344 & 1074 & 1443456 \\
AllGestureWiimoteX & Human Body & 1000 & 10 & 1 & 385 & 1000 & 385000 \\
Heartbeat & Healthcare & 409 & 2 & 61 & 405 & 24949 & 10104345 \\
Wafer & Facilities & 7164 & 2 & 1 & 152 & 7164 & 1088928 \\
FaceFour & Generated & 112 & 4 & 1 & 350 & 112 & 39200 \\
Phoneme & Audio & 2110 & 39 & 1 & 1024 & 2110 & 2160640 \\
InlineSkate & Human Body & 650 & 7 & 1 & 1882 & 650 & 1223300 \\
CricketX & Human Body & 780 & 12 & 1 & 300 & 780 & 234000 \\
SelfRegulationSCP2 & Healthcare & 380 & 2 & 7 & 1152 & 2660 & 3064320 \\
DistalPhalanxOutlineCorrect & Nature & 876 & 2 & 1 & 80 & 876 & 70080 \\
ChlorineConcentration & Nature & 4307 & 3 & 1 & 166 & 4307 & 714962 \\
Chinatown & Traffic & 363 & 2 & 1 & 24 & 363 & 8712 \\
GestureMidAirD1 & Human Body & 338 & 26 & 1 & 360 & 338 & 121680 \\
MiddlePhalanxOutlineAgeGroup & Nature & 554 & 3 & 1 & 80 & 554 & 44320 \\
UMD & Generated & 180 & 3 & 1 & 150 & 180 & 27000 \\
Crop & Nature & 24000 & 24 & 1 & 46 & 24000 & 1104000 \\
PenDigits & Facilities & 10992 & 10 & 2 & 8 & 21984 & 175872 \\
GesturePebbleZ1 & Human Body & 304 & 6 & 1 & 455 & 304 & 138320 \\
Handwriting & Facilities & 1000 & 26 & 3 & 152 & 3000 & 456000 \\
Mallat & Generated & 2400 & 8 & 1 & 1024 & 2400 & 2457600 \\
ERing & Human Body & 300 & 6 & 4 & 65 & 1200 & 78000 \\
StarLightCurves & Nature & 9236 & 3 & 1 & 1024 & 9236 & 9457664 \\
WordSynonyms & Audio & 905 & 25 & 1 & 270 & 905 & 244350 \\
PEMS-SF & Traffic & 440 & 7 & 963 & 144 & 423720 & 61015680 \\
FaceDetection & Human Body & 9414 & 2 & 144 & 62 & 1355616 & 84048192 \\
Computers & Power & 500 & 2 & 1 & 720 & 500 & 360000 \\
ArrowHead & Generated & 211 & 3 & 1 & 251 & 211 & 52961 \\
Wine & Nature & 111 & 2 & 1 & 234 & 111 & 25974 \\
Coffee & Nature & 56 & 2 & 1 & 286 & 56 & 16016 \\
Earthquakes & Nature & 461 & 2 & 1 & 512 & 461 & 236032 \\
Herring & Nature & 128 & 2 & 1 & 512 & 128 & 65536 \\
Lightning2 & Nature & 121 & 2 & 1 & 637 & 121 & 77077 \\
Beef & Nature & 60 & 5 & 1 & 470 & 60 & 28200 \\
MiddlePhalanxOutlineCorrect & Nature & 891 & 2 & 1 & 80 & 891 & 71280 \\
HealthcareFiveDays & Healthcare & 884 & 2 & 1 & 136 & 884 & 120224 \\
Yoga & Human Body & 3300 & 2 & 1 & 426 & 3300 & 1405800 \\
Adiac & Nature & 781 & 37 & 1 & 176 & 781 & 137456 \\
HandMovementDirection & Human Body & 234 & 4 & 10 & 400 & 2340 & 936000 \\
MoteStrain & Facilities & 1272 & 2 & 1 & 84 & 1272 & 106848 \\ 
Rock & Nature & 70 & 4 & 1 & 2844 & 70 & 199080 \\
Strawberry & Nature & 983 & 2 & 1 & 235 & 983 & 231005 \\
InsectWingbeatSound & Nature & 2200 & 11 & 1 & 256 & 2200 & 563200 \\
DodgerLoopWeekend & Traffic & 158 & 2 & 1 & 288 & 158 & 45504 \\
MixedShapesSmallTrain & Generated & 2525 & 5 & 1 & 1024 & 2525 & 2585600 \\
EthanolConcentration & Power & 524 & 4 & 3 & 1751 & 1572 & 2752572 \\
OliveOil & Nature & 60 & 4 & 1 & 570 & 60 & 34200 \\
Meat & Nature & 120 & 3 & 1 & 448 & 120 & 53760 \\
MelbournePedestrian & Traffic & 3633 & 10 & 1 & 24 & 3633 & 87192 \\
Car & Facilities & 120 & 4 & 1 & 577 & 120 & 69240 \\
FaceAll & Human Body & 2250 & 14 & 1 & 131 & 2250 & 294750 \\
FacesUCR & Human Body & 2250 & 14 & 1 & 131 & 2250 & 294750 \\
AllGestureWiimoteY & Human Body & 1000 & 10 & 1 & 369 & 1000 & 369000 \\
NATOPS & Human Body & 360 & 6 & 24 & 51 & 8640 & 440640 \\
SemgHandSubjectCh2 & Human Body & 900 & 5 & 1 & 1500 & 900 & 1350000 \\
ShakeGestureWiimoteZ & Human Body & 100 & 10 & 1 & 385 & 100 & 38500 \\
Cricket & Human Body & 180 & 12 & 6 & 1197 & 1080 & 1292760 \\
BME & Generated & 180 & 3 & 1 & 128 & 180 & 23040 \\
EigenWorms & Nature & 259 & 5 & 6 & 17984 & 1554 & 27947136 \\
FordB & Facilities & 4446 & 2 & 1 & 500 & 4446 & 2223000 \\
NonInvasiveFetalHealthcareThorax1 & Healthcare & 3765 & 42 & 1 & 750 & 3765 & 2823750 \\
UWaveGestureLibrary & Human Body & 440 & 8 & 3 & 315 & 1320 & 415800 \\
CinCHealthcareTorso & Healthcare & 1420 & 4 & 1 & 1639 & 1420 & 2327380 \\
PigArtPressure & Nature & 312 & 52 & 1 & 2000 & 312 & 624000 \\
Fish & Nature & 350 & 7 & 1 & 463 & 350 & 162050 \\
SonyAIBORobotSurface2 & Facilities & 980 & 2 & 1 & 65 & 980 & 63700 \\
FiftyWords & Facilities & 905 & 50 & 1 & 270 & 905 & 244350 \\
MotorImagery & Healthcare & 378 & 2 & 64 & 3000 & 24192 & 72576000 \\
ToeSegmentation1 & Human Body & 268 & 2 & 1 & 277 & 268 & 74236 \\
PhonemeSpectra & Audio & 6668 & 39 & 11 & 217 & 73348 & 15916516 \\
FreezerSmallTrain & Facilities & 2878 & 2 & 1 & 301 & 2878 & 866278 \\
TwoPatterns & Generated & 5000 & 4 & 1 & 128 & 5000 & 640000 \\
ShapeletSim & Generated & 200 & 2 & 1 & 500 & 200 & 100000 \\
Plane & Generated & 210 & 7 & 1 & 144 & 210 & 30240 \\
GestureMidAirD3 & Human Body & 338 & 26 & 1 & 360 & 338 & 121680 \\
DiatomSizeReduction & Generated & 322 & 4 & 1 & 345 & 322 & 111090 \\
Human BodyHorizontalSignal & Human Body & 724 & 12 & 1 & 1250 & 724 & 905000 \\
CricketZ & Human Body & 780 & 12 & 1 & 300 & 780 & 234000 \\
StandWalkJump & Human Body & 27 & 3 & 4 & 2500 & 108 & 270000 \\
WormsTwoClass & Human Body & 258 & 2 & 1 & 900 & 258 & 232200 \\
Lightning7 & Nature & 143 & 7 & 1 & 319 & 143 & 45617 \\
UWaveGestureLibraryY & Human Body & 4478 & 8 & 1 & 315 & 4478 & 1410570 \\
GunPointAgeSpan & Human Body & 451 & 2 & 1 & 150 & 451 & 67650 \\
DistalPhalanxOutlineAgeGroup & Nature & 539 & 3 & 1 & 80 & 539 & 43120 \\
SwedishLeaf & Nature & 1125 & 15 & 1 & 128 & 1125 & 144000 \\
LSST & Nature & 4925 & 14 & 6 & 36 & 29550 & 1063800 \\
CBF & Generated & 930 & 3 & 1 & 128 & 930 & 119040 \\
BeetleFly & Nature & 40 & 2 & 1 & 512 & 40 & 20480 \\
Libras & Human Body & 360 & 15 & 2 & 45 & 720 & 32400 \\
HouseTwenty & Facilities & 159 & 2 & 1 & 2000 & 159 & 318000 \\
ScreenType & Facilities & 750 & 3 & 1 & 720 & 750 & 540000 \\
InsectEPGSmallTrain & Nature & 266 & 3 & 1 & 601 & 266 & 159866 \\
AllGestureWiimoteZ & Sensor & 1000 & 10 & 1 & 326 & 1000 & 326000 \\
DodgerLoopDay & Sensor & 158 & 7 & 1 & 288 & 158 & 45504 \\
NonInvasiveFetalHealthcareThorax2 & Healthcare & 3765 & 42 & 1 & 750 & 3765 & 2823750 \\
BasicHuman Bodys & Human Body & 80 & 4 & 6 & 100 & 480 & 48000 \\
GunPointOldVersusYoung & Human Body & 451 & 2 & 1 & 150 & 451 & 67650 \\
FordA & Sensor & 4921 & 2 & 1 & 500 & 4921 & 2460500 \\
InsectWingbeat & Nature & 50000 & 10 & 200 & 22 & 10000000 & 220000000 \\
ItalyPowerDemand & Power & 1096 & 2 & 1 & 24 & 1096 & 26304 \\
ProximalPhalanxOutlineAgeGroup & Nature & 605 & 3 & 1 & 80 & 605 & 48400 \\
ACSF1 & Power & 200 & 10 & 1 & 1460 & 200 & 292000 \\
GunPoint & Human Body & 200 & 2 & 1 & 150 & 200 & 30000 \\
RacketSports & Human Body & 303 & 4 & 6 & 30 & 1818 & 54540 \\
SmallKitchenAppliances & Power & 750 & 3 & 1 & 720 & 750 & 540000 \\
ProximalPhalanxTW & Nature & 605 & 6 & 1 & 80 & 605 & 48400 \\
DuckDuckGeese & Facilities & 100 & 5 & 1345 & 270 & 134500 & 36315000 \\
PickupGestureWiimoteZ & Human Body & 100 & 10 & 1 & 361 & 100 & 36100 \\
EthanolLevel & Power & 1004 & 4 & 1 & 1751 & 1004 & 1758004 \\
SpokenArabicDigits & Audio & 8798 & 10 & 13 & 93 & 114374 & 10636782 \\
SonyAIBORobotSurface1 & Facilities & 621 & 2 & 1 & 70 & 621 & 43470 \\
HandOutlines & Human Body & 1370 & 2 & 1 & 2709 & 1370 & 3711330 \\
PowerCons & Power & 360 & 2 & 1 & 144 & 360 & 51840 \\
PhalangesOutlinesCorrect & Nature & 2658 & 2 & 1 & 80 & 2658 & 212640 \\
BirdChicken & Nature & 40 & 2 & 1 & 512 & 40 & 20480 \\
ToeSegmentation2 & Human Body & 166 & 2 & 1 & 343 & 166 & 56938 \\
PigCVP & Healthcare & 312 & 52 & 1 & 2000 & 312 & 624000 \\
CricketY & Human Body & 780 & 12 & 1 & 300 & 780 & 234000 \\
FingerMovements & Human Body & 416 & 2 & 28 & 50 & 11648 & 582400 \\
ElectricDevices & Power & 16637 & 7 & 1 & 96 & 16637 & 1597152 \\
InsectEPGRegularTrain & Nature & 311 & 3 & 1 & 601 & 311 & 186911 \\
DodgerLoopGame & Traffic & 158 & 2 & 1 & 288 & 158 & 45504 \\
Fungi & Nature & 204 & 18 & 1 & 201 & 204 & 41004 \\
Symbols & Generated & 1020 & 6 & 1 & 398 & 1020 & 405960 \\
MixedShapesRegularTrain & Generated & 2925 & 5 & 1 & 1024 & 2925 & 2995200 \\
ArticularyWordRecognition & Human Body & 575 & 25 & 9 & 144 & 5175 & 745200 \\
UWaveGestureLibraryZ & Human Body & 4478 & 8 & 1 & 315 & 4478 & 1410570 \\
Epilepsy & Human Body & 275 & 4 & 3 & 206 & 825 & 169950 \\
Healthcare200 & Healthcare & 200 & 2 & 1 & 96 & 200 & 19200 \\ \bottomrule
\caption{\textbf{Classification datasets in Time Series Pile taken from \citep{dl-for-ts-classification} [\protect\hyperlink{https://www.cs.ucr.edu/Eeamonn/time_series_data_2018/}{UCR Archive}]}. The total number of observations is 634,084,943, while the total number of unique time series is 290,226.}
\end{longtable}

\section{Experimental Setup and Results}
\label{app:experimental-setup-and-results}

Through our experiments, our goal is to answer the following research questions.

\paragraph{Is \texttt{MOMENT} effective for multiple time series analysis tasks in limited and rich supervision settings?} We conduct large-scale experiments on widely used benchmarks to evaluate \texttt{MOMENT} on forecasting, classification, anomaly detection, and imputation as outlined in Table \color{blue}8\color{black}. The \textit{limited supervision} setting mimics practical scenarios in which it is infeasible to train (or fine-tune) a deep neural network due to limited compute and, little or inadequately characterized data. In these settings, \texttt{MOMENT} provides predictions without any explicit (re)training on target data\footnote{For classification, the quality of \texttt{MOMENT}’s representations is measured using the accuracy of a Support Vector Machine trained on them, as is common in prior work on unsupervised representation learning \citep{TS2Vec,unsupervised-scalable-representation-learning-for-multivariate-ts}. However, unlike prior work, \texttt{MOMENT} embeds time series without any data-specific training.}. On the other hand, the rich supervision setting allows us to examine whether \texttt{MOMENT} can utilize task-specific data to improve its performance via end-to-end fine-tuning or linear probing.

\paragraph{What does \texttt{MOMENT} learn?} We evaluated \texttt{MOMENT}’s ability to model time series characteristics such as varying frequencies, trends, and scales. Structure in the PCA and t-SNE (Fig. 9) visualizations of the embeddings of synthetically generated sinusoids suggest that MOMENT can capture subtle trend, scale, frequency, and auto-correlation information. $\epsilon$ denotes gaussian noise with 0 mean and 0.1 standard deviation. $c$ controls the factor of interest, i.e. the power of the trend polynomial, amplitude, and frequency of the sine waves in experiments (i), (ii) \& (iii), respectively.

\paragraph{Hyper-parameter Tuning.} We do not perform extensive hyper-parameter tuning. In all experiments that follow, unless mentioned otherwise, we fine-tune \texttt{MOMENT-Base} with a batch size of 16, and cosine learning rate schedule with an initial learning rate of $1e^{-5}$. For baseline methods, we capture recommended settings from their respective papers and public repositories. We report all hyper-parameters settings for \texttt{MOMENT} and baselines in Appendix \color{blue}D\color{black}.

\subsection{Forecasting}

\paragraph{Task description.} Given a time series $\mathcal{T} = [x_1, . . . , x_L]$ where $x_i \in \mathbb{R}$, the univariate forecasting problem is to predict the next $H$ time-steps $[x_{L+1}, . . . , x_{L+H}]$. Depending on the length of the horizon, forecasting can be categorized as \textit{short} or \textit{long-horizon}\footnote{The distinction between long and short-horizon forecasting is rather arbitrary. For instance, most of the default forecasting horizons for the long-horizon forecasting benchmark Influenza-like Illness (24, 36, 48, 60) are shorter than the Hourly subset of the M4 dataset, a popular short-horizon forecasting benchmark.}. We consider both tasks in our experiments. We propose two configurations of \texttt{MOMENT} for the forecasting problem: (1) we can produce short-horizon forecasts without any explicit training or fine-tuning, by appending masked patches and predicting them using the default reconstruction head (Fig. \color{blue}4 (ii)\color{black}); (2) alternatively, we can replace the reconstruction head to a forecasting head and then fine-tune it (Fig. \color{blue}4 (i)\color{black}).

\subsubsection{Long-Horizon Forecasting}

\paragraph{Datasets.} We use all the long-horizon forecasting datasets (Sec \ref{subsec:timeseries_pile}). But to speed up our experiments, we drop all exogenous variables from multi-variate datasets and only consider the target time series for forecasting.

\paragraph{Baselines.} We compare our methods with various transformer-based and deep learning baselines. These models can be found in Table \ref{tab:long_horizon_forecasting_appendix}. For Time-LLM we could not run experiments on Weather, electricity, and traffic datasets, due to time constraints, and since we could not fit them into a single GPU. 

\paragraph{Experimental Setting.} We train all models with a look-back window of length $L =$ 512 to forecast $T =$ {24, 60} time-steps for the \texttt{ILI} dataset and $T =$ {96, 720} for the rest. We evaluate the Mean Squared Error (MSE) and Mean Absolute Error (MAE) as metrics.

\paragraph{Hyperparameters.} The hyperparameters used for training all models in our long-horizon forecasting experiments are shown in Table \ref{tab:hyperparameters-long-horizon-forecasting}.

\begin{table}[htb!]
\centering
\resizebox{0.5\columnwidth}{!}{
\begin{tabular}{r|rl}
\toprule
\textbf{Model} & \multicolumn{2}{c}{\textbf{Hyper-parameters}} \\ \midrule
\multirow{5}{*}{\texttt{MOMENT}} & \texttt{sequence length} : & $512$ \\
 & \texttt{patch length} : & $8$ \\
 & \texttt{patch stride length} : & $8$ \\
 & \texttt{initial learning rate} : & $0.0001$ \\
 & \texttt{forecast horizon} : & $\{96, 720\}$ \\ \midrule
\multirow{7}{*}{\texttt{Time-LLM}} & \texttt{sequence length} : & $512$ \\
 & \texttt{patch length} : & $16$ \\
 & \texttt{patch stride length} : & $8$ \\
 & \texttt{initial learning rate} : & $0.001$ \\
 & \texttt{dimension of feedforward layer} : & $2048$ \\
 & \texttt{llm layers} : & $32$ \\
 & \texttt{number of heads} : & $8$ \\ \midrule
\multirow{5}{*}{\texttt{N-BEATS}} & \texttt{sequence length} : & $512$ \\
 & \texttt{stack types} : & \{trend, seasonality\} \\
 & \texttt{number of blocks per stack} : & $3$ \\
 & \texttt{thetas dimensions} : & $\{4,8\}$ \\
 & \texttt{hidden layer units} : & $256$ \\ \midrule
\end{tabular}}
\caption{Hyper-parameter values for long-horizon forecasting models.}
\label{tab:hyperparameters-long-horizon-forecasting}
\end{table}

\begin{table*}[htb!]
\centering
\large
\resizebox{\linewidth}{!}{
\begin{tabular}{cc|cc|cc|cc|cc|cc|cc|cc|cc|cc|cc|cc|cc|cc|cc|cc|cc|cc}
\toprule
\multicolumn{2}{c}{Methods} & \multicolumn{2}{c}{$\mathtt{MOMENT_{LP}}$} & \multicolumn{2}{c}{\textbf{Time-LLM}} & \multicolumn{2}{c}{\textbf{GPT4TS}} & \multicolumn{2}{c}{\textbf{PatchTST}} & \multicolumn{2}{c}{\textbf{DLinear}} & \multicolumn{2}{c}{\textbf{TimesNet}} & \multicolumn{2}{c}{\textbf{FEDformer}} & \multicolumn{2}{c}{\textbf{Pyraformer}} & \multicolumn{2}{c}{\textbf{Autoformer}} & \multicolumn{2}{c}{\textbf{Stationary}} & \multicolumn{2}{c}{\textbf{ETSformer}} & \multicolumn{2}{c}{\textbf{LightTS}} & \multicolumn{2}{c}{\textbf{Informer}} & \multicolumn{2}{c}{\textbf{Reformer}} & \multicolumn{2}{c}{\textbf{LogTrans}} & \multicolumn{2}{c}{\textbf{N-BEATS}} \\ 
\multicolumn{2}{c}{Metric} & MSE & MAE & MSE & MAE & MSE & MAE & MSE & MAE & MSE & MAE & MSE & MAE & MSE & MAE & MSE & MAE & MSE & MAE & MSE & MAE & MSE & MAE & MSE & MAE & MSE & MAE & MSE & MAE & MSE & MAE & MSE & MAE \\ \midrule
\multirow{2}{*}{Weather} & 96 & 0.154 & 0.209 & - & - & 0.162 & 0.212 & 0.149 & 0.198 & 0.176 & 0.237 & 0.172 & 0.220 & 0.217 & 0.296 & 0.896 & 0.556 & 0.266 & 0.336 & 0.173 & 0.223 & 0.197 & 0.281 & 0.182 & 0.242 & 0.300 & 0.384 & 0.689 & 0.596 & 0.458 & 0.490 & 0.152 & 0.210 \\
 & 720 & 0.315 & 0.336 & - & - & 0.326 & 0.337 & 0.314 & 0.334 & 0.333 & 0.362 & 0.365 & 0.359 & 0.403 & 0.428 & 1.004 & 0.934 & 0.419 & 0.428 & 0.414 & 0.410 & 0.352 & 0.288 & 0.352 & 0.386 & 1.059 & 0.741 & 1.130 & 0.792 & 0.869 & 0.675 & 0.331 & 0.359 \\ \midrule
\multirow{2}{*}{ETTh1} & 96 & 0.387 & 0.410 & 0.408 & 0.429 & 0.376 & 0.397 & 0.370 & 0.399 & 0.375 & 0.399 & 0.384 & 0.402 & 0.376 & 0.419 & 0.664 & 0.612 & 0.449 & 0.459 & 0.513 & 0.491 & 0.494 & 0.479 & 0.424 & 0.432 & 0.865 & 0.713 & 0.837 & 0.728 & 0.878 & 0.740 & 0.399 & 0.428 \\
 & 720 & 0.454 & 0.472 & 0.523 & 0.514 & 0.477 & 0.456 & 0.447 & 0.466 & 0.472 & 0.490 & 0.521 & 0.500 & 0.506 & 0.507 & 0.963 & 0.782 & 0.514 & 0.512 & 0.643 & 0.616 & 0.562 & 0.535 & 0.547 & 0.533 & 1.181 & 0.865 & 1.257 & 0.889 & 1.135 & 0.852 & 0.608 &	0.573 \\ \midrule
\multirow{2}{*}{ETTh2} & 96 & 0.288 & 0.345 & 0.285 & 0.348 & 0.285 & 0.342 & 0.274 & 0.336 & 0.289 & 0.353 & 0.340 & 0.374 & 0.358 & 0.397 & 0.645 & 0.597 & 0.346 & 0.388 & 0.476 & 0.458 & 0.340 & 0.391 & 0.397 & 0.437 & 3.755 & 1.525 & 2.626 & 1.317 & 2.116 & 1.197 & 0.327 & 0.387 \\
 & 720 & 0.403 & 0.439 & 0.399 & 0.435 & 0.406 & 0.441 & 0.379 & 0.422 & 0.605 & 0.551 & 0.462 & 0.468 & 0.463 & 0.474 & 0.963 & 0.783 & 0.515 & 0.511 & 0.562 & 0.560 & 0.500 & 0.497 & 0.863 & 0.672 & 3.647 & 1.625 & 3.874 & 1.697 & 3.188 & 1.540 & 1.454 & 0.847 \\ \midrule
\multirow{2}{*}{ETTm1} & 96 & 0.293 & 0.349 & 0.384 & 0.403 & 0.292 & 0.346 & 0.290 & 0.342 & 0.299 & 0.343 & 0.338 & 0.375 & 0.379 & 0.419 & 0.543 & 0.510 & 0.505 & 0.475 & 0.386 & 0.398 & 0.375 & 0.398 & 0.374 & 0.400 & 0.672 & 0.571 & 0.538 & 0.528 & 0.600 & 0.546 & 0.318 & 0.367\\ 
 & 720 & 0.405 & 0.416 & 0.437 & 0.429 & 0.417 & 0.421 & 0.416 & 0.420 & 0.425 & 0.421 & 0.478 & 0.450 & 0.543 & 0.490 & 0.908 & 0.724 & 0.671 & 0.561 & 0.585 & 0.516 & 0.499 & 0.462 & 0.527 & 0.502 & 1.166 & 0.823 & 1.102 & 0.841 & 1.153 & 0.820 & 0.448 &	0.448\\ \midrule
\multirow{2}{*}{ETTm2} & 96 & 0.170 & 0.260 & 0.181 & 0.269 & 0.173 & 0.262 & 0.165 & 0.255 & 0.167 & 0.269 & 0.187 & 0.267 & 0.203 & 0.287 & 0.435 & 0.507 & 0.255 & 0.339 & 0.192 & 0.274 & 0.189 & 0.280 & 0.209 & 0.308 & 0.365 & 0.453 & 0.658 & 0.619 & 0.768 & 0.642 & 0.197 & 0.271\\
 & 720 & 0.363 & 0.387 & 0.366 & 0.388 & 0.378 & 0.401 & 0.362 & 0.385 & 0.397 & 0.421 & 0.408 & 0.403 & 0.421 & 0.415 & 3.625 & 1.451 & 0.433 & 0.432 & 0.417 & 0.413 & 0.414 & 0.413 & 0.675 & 0.587 & 3.379 & 1.338 & 2.631 & 1.242 & 3.048 & 1.328 & 0.395 &	0.419 \\ \midrule
\multirow{2}{*}{ILI} & 24 & {2.728} & {1.114} & 3.025 & 1.195 & 2.063 & 0.881 & 1.319 & 0.754 & 2.215 & 1.081 & 2.317 & 0.934 & 3.228 & 1.260 & 1.420 & 2.012 & 3.483 & 1.287 & 2.294 & 0.945 & 2.527 & 1.020 & 8.313 & 2.144 & 5.764 & 1.677 & 4.400 & 1.382 & 4.480 & 1.444 & 4.539 & 1.528 \\
& 60 & {2.893} & {1.132} & 3.245 & 1.221 & 1.979 & 0.957 & 1.470 & 0.788 & 2.368 & 1.096 & 2.027 & 0.928 & 2.857 & 1.157 & 7.662 & 2.100 & 2.770 & 1.125 & 2.178 & 0.963 & 2.487 & 1.016 & 7.283 & 1.985 & 5.264 & 1.564 & 4.882 & 1.483 & 5.278 & 1.560 & 5.429 & 1.661 \\ \midrule
\multirow{2}{*}{ECL} & 96 & 0.138 & 0.242 & - & - & 0.139 & 0.238 & 0.129 & 0.222 & 0.140 & 0.237 & 0.168 & 0.272 & 0.193 & 0.308 & 0.386 & 0.449 & 0.201 & 0.317 & 0.169 & 0.273 & 0.187 & 0.304 & 0.207 & 0.307 & 0.274 & 0.368 & 0.312 & 0.402 & 0.258 & 0.357 & 0.131 & 0.228 \\
 & 720 & {0.211} & {0.305} & - & - & 0.206 & 0.297 & 0.197 & 0.290 & 0.203 & 0.301 & 0.220 & 0.320 & 0.246 & 0.355 & 0.376 & 0.445 & 0.254 & 0.361 & 0.222 & 0.321 & 0.233 & 0.345 & 0.265 & 0.360 & 0.373 & 0.439 & 0.340 & 0.420 & 0.283 & 0.376 & 0.208 & 0.298 \\ \midrule
\multirow{2}{*}{Traffic} & 96 & 0.391 & 0.282 & - & - & 0.388 & 0.282 & 0.360 & 0.249 & 0.410 & 0.282 & 0.593 & 0.321 & 0.587 & 0.366 & 2.085 & 0.468 & 0.613 & 0.388 & 0.612 & 0.338 & 0.607 & 0.392 & 0.615 & 0.391 & 0.719 & 0.391 & 0.732 & 0.423 & 0.684 & 0.384 & 0.375 & 0.259 \\
 & 720 & 0.450 & 0.310 & - & - & 0.450 & 0.312 & 0.432 & 0.286 & 0.466 & 0.315 & 0.640 & 0.350 & 0.626 & 0.382 & 0.881 & 0.473 & 0.660 & 0.408 & 0.653 & 0.355 & 0.632 & 0.396 & 0.658 & 0.407 & 0.864 & 0.472 & 0.755 & 0.423 & 0.717 & 0.396 & 0.508 & 0.335 \\ \bottomrule
\end{tabular}}
\caption{Long-term forecasting performance measured using Mean Squared Error (MSE) and Mean Absolute Error (MAE).}
\label{tab:long_horizon_forecasting_appendix}
\end{table*}

\begin{table}[htb!]
\centering
\resizebox{\linewidth}{!}{
\begin{tabular}{cccccccccccccccccccccccccc}
\toprule
\multicolumn{2}{c}{\textbf{Models}} & \multicolumn{2}{c}{\textbf{MOMENT}} & \multicolumn{2}{c}{\textbf{iTransformer}} & \multicolumn{2}{c}{\textbf{RLinear}} & \multicolumn{2}{c}{\textbf{PatchTST}} & \multicolumn{2}{c}{\textbf{Crossformer}} & \multicolumn{2}{c}{\textbf{TiDE}} & \multicolumn{2}{c}{\textbf{TimesNet}} & \multicolumn{2}{c}{\textbf{DLinear}} & \multicolumn{2}{c}{\textbf{SCINet}} & \multicolumn{2}{c}{\textbf{FEDFormer}} & \multicolumn{2}{c}{\textbf{Stationary}} & \multicolumn{2}{c}{\textbf{Autoformer}} \\ \midrule
Dataset & Pred\_horizon & MSE & MAE & MSE & MAE & MSE & MAE & MSE & MAE & MSE & MAE & MSE & MAE & MSE & MAE & MSE & MAE & MSE & MAE & MSE & MAE & MSE & MAE & MSE & MAE \\ \midrule
PEMS08 & 12 & 0.132 & 0.249 & 0.079 & 0.182 & 0.133 & 0.247 & 0.168 & 0.232 & 0.165 & 0.214 & 0.227 & 0.343 & 0.112 & 0.212 & 0.154 & 0.276 & 0.087 & 0.184 & 0.173 & 0.273 & 0.109 & 0.207 & 0.436 & 0.485 \\
 & 24 & 0.212 & 0.320 & 0.115 & 0.219 & 0.249 & 0.343 & 0.224 & 0.281 & 0.215 & 0.260 & 0.318 & 0.409 & 0.141 & 0.238 & 0.248 & 0.353 & 0.122 & 0.221 & 0.210 & 0.301 & 0.140 & 0.236 & 0.467 & 0.502 \\
 & 36 & 0.309 & 0.393 & 0.186 & 0.235 & 0.569 & 0.544 & 0.321 & 0.354 & 0.315 & 0.355 & 0.497 & 0.510 & 0.198 & 0.283 & 0.440 & 0.470 & 0.189 & 0.270 & 0.320 & 0.394 & 0.211 & 0.294 & 0.966 & 0.733 \\
 \bottomrule
\end{tabular}}
\caption{Long-term forecasting performance with a look-back window of 96 time-steps. Results are taken from iTransformer \citep{itransformer}.}
\end{table}

\subsubsection{Zero-shot Short-Horizon Forecasting}
\label{app:zero_shot_short_horizon_forecasting}
\paragraph{Datasets.} To evaluate zero-shot forecasting performance, we conduct experiments on the M3 and M4 datasets (Sec. \ref{subsec:timeseries_pile}).

\paragraph{Baselines.} We compare \texttt{MOMENT} with GPT4TS \citep{one-fits-all}, TimesNet \citep{timesnet}, N-BEATS \citep{N-BEATS}, 3 statistical and 3 benchmarking forecasting methods: AutoARIMA, AutoTheta, AutoETS, Naive, Seasonal Naive, and Random Walk (Makridakis et al., 2020).

\paragraph{Experimental Setting.} Each statistical method is \textit{fit} on individual time series before producing a forecast. We follow the same train-test split and forecasting horizons from the M3 and M4 competitions, and report sMAPE as is common in prior work \citep{N-BEATS,timesnet}\footnote{The definitions of sMAPE were different in the M3 and M4 competitions. In our experiments, we used the same definition as the M4 competition.}. We follow the same experimental procedure as outlined in \citep{meta-learning-zero-shot-ts-forecasting} with two exceptions: our results are reported only (1) on 40\% of the M3 and M4 datasets that were unseen during pre-training, (2) a subset of frequencies with largest support in the datasets. Daily, hourly, and weekly frequencies had very little data and we could not get promising zero-shot performance for any of the deep learning models. Some ways that prior work \citep{meta-learning-zero-shot-ts-forecasting} had overcome this issue was by leveraging data from frequencies with plenty of data. We also believe that ensembling played an important part in N-BEATS promising zero-shot performance.    

\paragraph{Hyperparameters.} The hyperparameters used for training all models in our short-horizon forecasting experiments are shown in Table \ref{tab:hyperparameters-short-horizon-forecasting}.

\begin{table}[htb!]
\centering
\resizebox{0.6\columnwidth}{!}{
\begin{tabular}{r|rl}
\toprule
\textbf{Model} & \multicolumn{2}{c}{\textbf{Hyper-parameters}} \\ \midrule
\multirow{5}{*}{$\mathtt{MOMENT_{LP}}$} & \texttt{sequence length} : & $512$ \\
 & \texttt{patch length} : & $8$ \\
 & \texttt{patch stride length} : & $8$ \\
 & \texttt{initial learning rate} : & $0.002$ \\
 & \texttt{max epochs} : & $\{5, 10\}$ \\ \midrule
\multirow{4}{*}{$\mathtt{MOMENT_0}$} & \texttt{sequence length} : & $512$ \\
 & \texttt{patch length} : & $8$ \\
 & \texttt{patch stride length} : & $8$ \\
 & \texttt{initial learning rate} : & $0.001$ \\ \midrule
\multirow{5}{*}{\texttt{N-BEATS}} & \texttt{sequence length} : & $512$ \\
 & \texttt{stack types} : & \{'trend', 'seasonality'\} \\
 & \texttt{number of blocks per stack} : & $3$ \\
 & \texttt{thetas dimensions} : & $\{4,8\}$ \\
 & \texttt{hidden layer units} : & $256$ \\ \midrule
\multirow{5}{*}{\texttt{GPT4TS}} & \texttt{forecast horizon} : & $0$ \\
 & \texttt{gpt layers} : & $3$ \\
 & \texttt{patch length} : & $1$ \\
 & \texttt{patch stride length} : & $1$ \\
 & \texttt{sequence length} : & $512$ \\ \midrule
\multirow{4}{*}{\texttt{TimesNet}} & \texttt{sequence length} : & $512$ \\
 & \texttt{model dimension} : & $32$ \\
 & \texttt{dimension of feedforward layer} : & $32$ \\
 & \texttt{top-$k$} : & $5$ \\ \midrule
\end{tabular}}
\caption{Hyper-parameter values for short-horizon forecasting models.}
\label{tab:hyperparameters-short-horizon-forecasting}
\end{table}

\begin{table}[htb!]
\centering
\resizebox{0.3\linewidth}{!}{
\begin{tabular}{c|cc}
\toprule
Source Dataset $\rightarrow$ & \multirow{2}{*}{M4} & \multirow{2}{*}{Fred} \\
Target Dataset $\downarrow$ &  &  \\ \midrule
\textbf{M4} & \textbf{} &  \\
Yearly & \textbf{-} & Yearly \\
Quarterly & \textbf{-} & Quarterly \\
Monthly & \textbf{-} & Monthly \\
\textbf{M3} & \textbf{} &  \\
Yearly & Yearly & Yearly \\
Quarterly & Quarterly & Quarterly \\
Monthly & Monthly & Monthly \\
\bottomrule
\end{tabular}}
\caption{Experimental settings for short-horizon forecasting experiments for varying source and target datasets.}
\label{tab:short-horizon-forecasting-dataset-settings}
\end{table}

\begin{table}[!tbh]
\centering
\resizebox{\linewidth}{!}{
\begin{tabular}{r|cccccc|cccccc}
\toprule
 & \multicolumn{6}{c}{Adjusted Best $F_1$} & \multicolumn{6}{c}{VUS-ROC} \\
\textbf{Model name} & \textbf{Anomaly Transformer} & $\mathtt{MOMENT_{0}}$ & $\mathtt{MOMENT_{LP}}$ & \textbf{DGHL} & \textbf{GPT4TS} & \textbf{TimesNet} & AnomalyTransformer & $\mathtt{MOMENT_{0}}$ & $\mathtt{MOMENT_{LP}}$ & \textbf{DGHL} & \textbf{GPT4TS} & \textbf{TimesNet} \\
Dataset name &  &  &  &  &  &  &  &  &  &  &  &  \\
\midrule
1sddb40 & 0.030 & 0.560 & 0.540 & 0.390 & 0.190 & 0.680 & 0.640 & 0.740 & 0.750 & 0.640 & 0.660 & 0.720 \\
BIDMC1 & 0.990 & 1.000 & 1.000 & 1.000 & 1.000 & 1.000 & 0.690 & 0.560 & 0.650 & 0.720 & 0.630 & 0.740 \\
CHARISfive & 0.010 & 0.070 & 0.130 & 0.020 & 0.020 & 0.080 & 0.360 & 0.430 & 0.400 & 0.510 & 0.450 & 0.460 \\
CHARISten & 0.020 & 0.060 & 0.110 & 0.040 & 0.100 & 0.030 & 0.430 & 0.500 & 0.540 & 0.520 & 0.510 & 0.530 \\
CIMIS44AirTemperature3 & 0.060 & 1.000 & 0.980 & 0.500 & 0.180 & 0.470 & 0.640 & 0.740 & 0.750 & 0.740 & 0.620 & 0.740 \\
CIMIS44AirTemperature5 & 0.390 & 0.990 & 0.990 & 0.960 & 0.200 & 0.710 & 0.780 & 0.750 & 0.810 & 0.920 & 0.560 & 0.720 \\
ECG2 & 1.000 & 1.000 & 1.000 & 0.620 & 0.900 & 1.000 & 0.830 & 0.740 & 0.840 & 0.630 & 0.780 & 0.600 \\
ECG3 & 0.360 & 0.810 & 0.980 & 0.800 & 0.840 & 0.480 & 0.540 & 0.700 & 0.770 & 0.680 & 0.450 & 0.610 \\
Fantasia & 0.750 & 1.000 & 0.950 & 0.660 & 0.870 & 0.550 & 0.730 & 0.630 & 0.640 & 0.710 & 0.650 & 0.610 \\
GP711MarkerLFM5z4 & 0.930 & 0.810 & 1.000 & 0.500 & 0.640 & 0.950 & 0.540 & 0.630 & 0.730 & 0.600 & 0.620 & 0.720 \\
GP711MarkerLFM5z5 & 0.760 & 0.690 & 0.970 & 0.310 & 0.480 & 0.900 & 0.690 & 0.760 & 0.720 & 0.520 & 0.630 & 0.840 \\
InternalBleeding4 & NaN & 1.000 & NaN & NaN & NaN & NaN & NaN & 0.650 & NaN & NaN & NaN & NaN \\
InternalBleeding5 & 0.940 & 1.000 & 1.000 & 1.000 & 0.920 & 1.000 & 0.460 & 0.600 & 0.690 & 0.760 & 0.630 & 0.940 \\
Italianpowerdemand & 0.010 & 0.390 & 0.740 & 0.590 & 0.010 & 0.440 & 0.450 & 0.800 & 0.770 & 0.700 & 0.480 & 0.710 \\
Lab2Cmac011215EPG5 & 0.990 & 0.970 & 0.980 & 0.340 & 0.600 & 0.990 & 0.770 & 0.620 & 0.630 & 0.710 & 0.640 & 0.610 \\
Lab2Cmac011215EPG6 & 0.410 & 0.090 & 0.100 & 0.260 & 0.100 & 0.170 & 0.700 & 0.480 & 0.480 & 0.600 & 0.520 & 0.450 \\
MesoplodonDensirostris & 1.000 & 0.910 & 0.840 & 0.790 & 1.000 & 1.000 & 0.850 & 0.730 & 0.720 & 0.740 & 0.690 & 0.790 \\
PowerDemand1 & 0.870 & 0.260 & 0.440 & 0.490 & 0.760 & 0.950 & 0.720 & 0.520 & 0.540 & 0.530 & 0.600 & 0.750 \\
TkeepFirstMARS & 0.010 & 0.080 & 0.150 & 0.020 & 0.020 & 0.230 & 0.520 & 0.570 & 0.760 & 0.460 & 0.500 & 0.790 \\
TkeepSecondMARS & 0.830 & 0.950 & 1.000 & 0.160 & 0.120 & 0.950 & 0.720 & 0.950 & 0.910 & 0.970 & 0.810 & 0.980 \\
WalkingAceleration5 & 0.990 & 1.000 & 1.000 & 0.910 & 0.870 & 0.930 & 0.940 & 0.860 & 0.870 & 0.930 & 0.910 & 0.850 \\
apneaecg & 0.400 & 0.210 & 0.200 & 0.250 & 0.310 & 0.260 & 0.580 & 0.690 & 0.690 & 0.590 & 0.580 & 0.760 \\
apneaecg2 & 0.650 & 0.940 & 1.000 & 1.000 & 1.000 & 0.650 & 0.790 & 0.750 & 0.740 & 0.730 & 0.650 & 0.610 \\
gait1 & 0.180 & 0.710 & 0.360 & 0.070 & 0.410 & 0.520 & 0.630 & 0.650 & 0.570 & 0.600 & 0.580 & 0.600 \\
gaitHunt1 & 0.080 & 0.500 & 0.430 & 0.020 & 0.100 & 0.300 & 0.810 & 0.640 & 0.680 & 0.570 & 0.710 & 0.840 \\
insectEPG2 & 0.120 & 0.110 & 0.230 & 0.140 & 0.810 & 0.960 & 0.650 & 0.570 & 0.820 & 0.650 & 0.560 & 0.730 \\
insectEPG4 & 0.980 & 1.000 & 1.000 & 0.460 & 0.210 & 0.850 & 0.690 & 0.700 & 0.720 & 0.730 & 0.490 & 0.650 \\
ltstdbs30791AS & 1.000 & 1.000 & 1.000 & 1.000 & 1.000 & 1.000 & 0.780 & 0.760 & 0.810 & 0.770 & 0.740 & 0.670 \\
mit14046longtermecg & 0.450 & 0.560 & 0.590 & 0.530 & 0.580 & 0.600 & 0.790 & 0.660 & 0.660 & 0.640 & 0.610 & 0.840 \\
park3m & 0.150 & 0.560 & 0.640 & 0.200 & 0.630 & 0.930 & 0.630 & 0.750 & 0.780 & 0.650 & 0.540 & 0.780 \\
qtdbSel1005V & 0.410 & 0.570 & 0.650 & 0.400 & 0.390 & 0.530 & 0.520 & 0.640 & 0.640 & 0.490 & 0.610 & 0.540 \\
qtdbSel100MLII & 0.420 & 0.780 & 0.840 & 0.410 & 0.600 & 0.870 & 0.620 & 0.580 & 0.620 & 0.590 & 0.580 & 0.650 \\
resperation1 & 0.000 & 0.040 & 0.150 & 0.030 & 0.010 & 0.030 & 0.750 & 0.500 & 0.670 & 0.740 & 0.470 & 0.670 \\
s20101mML2 & 0.690 & 0.650 & 0.710 & 0.150 & 0.050 & 0.080 & 0.640 & 0.760 & 0.720 & 0.690 & 0.640 & 0.690 \\
sddb49 & 0.890 & 1.000 & 1.000 & 0.880 & 0.940 & 1.000 & 0.660 & 0.730 & 0.730 & 0.740 & 0.580 & 0.680 \\
sel840mECG1 & 0.160 & 0.610 & 0.660 & 0.280 & 0.210 & 0.360 & 0.620 & 0.720 & 0.720 & 0.870 & 0.650 & 0.600 \\
sel840mECG2 & 0.150 & 0.360 & 0.390 & 0.320 & 0.280 & 0.210 & 0.590 & 0.710 & 0.690 & 0.490 & 0.520 & 0.520 \\
tilt12744mtable & 0.070 & 0.110 & 0.240 & 0.100 & 0.000 & 0.030 & 0.480 & 0.670 & 0.740 & 0.660 & 0.510 & 0.640 \\
tilt12754table & 0.230 & 0.590 & 0.640 & 0.040 & 0.060 & 0.050 & 0.600 & 0.750 & 0.820 & 0.790 & 0.550 & 0.750 \\
tiltAPB2 & 0.920 & 0.960 & 0.980 & 0.360 & 0.830 & 0.380 & 0.770 & 0.750 & 0.770 & 0.710 & 0.600 & 0.700 \\
tiltAPB3 & 0.170 & 0.480 & 0.850 & 0.030 & 0.050 & 0.090 & 0.680 & 0.610 & 0.650 & 0.540 & 0.440 & 0.580 \\
weallwalk & 0.000 & 0.520 & 0.580 & 0.070 & 0.130 & 0.170 & 0.730 & 0.930 & 0.930 & 0.860 & 0.870 & 0.850 \\
\bottomrule
\end{tabular}}
\caption{Anomaly detection performance measured using adj. best $F_1$ and VUS-ROC for a subset of 45 datasets sampled from the UCR Anomaly archive.}
\label{tab:anomaly_detection_results_appendix}
\end{table}

\begin{table}[!tbh]
\centering
\resizebox{\linewidth}{!}{
\begin{tabular}{r|c|cc|ccccc|cccccccc|c}
\toprule
\textbf{Dataset} & $\mathtt{MOMENT_{0}}$ & \textbf{TimesNet} & \textbf{GPT4TS} & \textbf{TS2Vec} & \textbf{T-Loss} & \textbf{TNC} & \textbf{TS-TCC} & \textbf{TST} & \textbf{CNN} & \textbf{Encoder} & \textbf{FCN} & \textbf{MCDNN} & \textbf{MLP} & \textbf{ResNet} & \textbf{t-LeNet} & \textbf{TWIESN} & \textbf{DTW} \\ \midrule
GestureMidAirD2 & 0.608 & 0.131 & 0.200 & 0.469 & 0.546 & 0.254 & 0.254 & 0.138 & 0.518 & 0.480 & 0.631 & 0.500 & 0.545 & 0.668 & 0.038 & 0.575 & 0.608 \\
UWaveGestureLibraryX & 0.821 & 0.688 & 0.749 & 0.795 & 0.785 & 0.733 & 0.733 & 0.569 & 0.721 & 0.771 & 0.754 & 0.726 & 0.768 & 0.781 & 0.127 & 0.608 & 0.728 \\
GesturePebbleZ2 & 0.816 & 0.310 & 0.285 & 0.873 & 0.899 & 0.430 & 0.430 & 0.380 & 0.778 & 0.796 & 0.781 & 0.720 & 0.701 & 0.777 & 0.184 & 0.843 & 0.671 \\
ECG5000 & 0.942 & 0.584 & 0.584 & 0.935 & 0.933 & 0.941 & 0.941 & 0.928 & 0.928 & 0.941 & 0.940 & 0.933 & 0.930 & 0.935 & 0.584 & 0.922 & 0.924 \\
OSULeaf & 0.785 & 0.397 & 0.231 & 0.851 & 0.760 & 0.723 & 0.723 & 0.545 & 0.482 & 0.554 & 0.979 & 0.419 & 0.560 & 0.980 & 0.182 & 0.628 & 0.591 \\
MedicalImages & 0.762 & 0.571 & 0.496 & 0.789 & 0.750 & 0.747 & 0.747 & 0.632 & 0.671 & 0.664 & 0.778 & 0.627 & 0.719 & 0.770 & 0.514 & 0.649 & 0.737 \\
Ham & 0.581 & 0.686 & 0.781 & 0.714 & 0.724 & 0.743 & 0.743 & 0.524 & 0.720 & 0.682 & 0.707 & 0.718 & 0.699 & 0.758 & 0.514 & 0.768 & 0.467 \\
DistalPhalanxTW & 0.612 & 0.604 & 0.619 & 0.698 & 0.676 & 0.676 & 0.676 & 0.568 & 0.671 & 0.694 & 0.695 & 0.685 & 0.610 & 0.663 & 0.285 & 0.591 & 0.590 \\
ProximalPhalanxOutlineCorrect & 0.856 & 0.869 & 0.801 & 0.887 & 0.859 & 0.873 & 0.873 & 0.770 & 0.807 & 0.768 & 0.907 & 0.866 & 0.730 & 0.920 & 0.684 & 0.817 & 0.784 \\
FreezerRegularTrain & 0.982 & 0.926 & 0.829 & 0.986 & 0.956 & 0.989 & 0.989 & 0.922 & 0.987 & 0.760 & 0.997 & 0.973 & 0.906 & 0.998 & 0.500 & 0.946 & 0.899 \\
TwoLeadECG & 0.847 & 0.633 & 0.658 & 0.986 & 0.999 & 0.976 & 0.976 & 0.871 & 0.877 & 0.784 & 0.999 & 0.806 & 0.753 & 1.000 & 0.500 & 0.949 & 0.905 \\
GunPointMaleVersusFemale & 0.991 & 0.601 & 0.475 & 1.000 & 0.997 & 0.997 & 0.997 & 1.000 & 0.977 & 0.978 & 0.997 & 0.952 & 0.980 & 0.992 & 0.525 & 0.988 & 0.997 \\
Trace & 1.000 & 0.760 & 0.710 & 1.000 & 0.990 & 1.000 & 1.000 & 1.000 & 0.952 & 0.740 & 1.000 & 0.902 & 0.806 & 1.000 & 0.240 & 0.934 & 1.000 \\
SmoothSubspace & 0.820 & 0.440 & 0.453 & 0.980 & 0.960 & 0.953 & 0.953 & 0.827 & 0.976 & 0.964 & 0.975 & 0.963 & 0.980 & 0.980 & 0.333 & 0.849 & 0.827 \\
MiddlePhalanxTW & 0.532 & 0.506 & 0.571 & 0.584 & 0.591 & 0.610 & 0.610 & 0.506 & 0.551 & 0.597 & 0.501 & 0.562 & 0.536 & 0.495 & 0.286 & 0.569 & 0.506 \\
SyntheticControl & 0.990 & 0.467 & 0.437 & 0.997 & 0.987 & 0.990 & 0.990 & 0.490 & 0.987 & 0.973 & 0.989 & 0.953 & 0.973 & 0.997 & 0.167 & 0.879 & 0.993 \\
ShapesAll & 0.815 & 0.238 & 0.237 & 0.902 & 0.848 & 0.773 & 0.773 & 0.733 & 0.617 & 0.679 & 0.894 & 0.599 & 0.776 & 0.926 & 0.017 & 0.643 & 0.768 \\
AllGestureWiimoteX & 0.607 & 0.209 & 0.237 & 0.777 & 0.763 & 0.697 & 0.697 & 0.259 & 0.411 & 0.475 & 0.713 & 0.261 & 0.477 & 0.741 & 0.100 & 0.522 & 0.716 \\
Wafer & 0.997 & 0.989 & 0.994 & 0.998 & 0.992 & 0.994 & 0.994 & 0.991 & 0.961 & 0.998 & 0.997 & 0.992 & 0.996 & 0.998 & 0.892 & 0.916 & 0.980 \\
FaceFour & 0.852 & 0.830 & 0.659 & 0.932 & 0.920 & 0.773 & 0.773 & 0.511 & 0.905 & 0.852 & 0.930 & 0.711 & 0.836 & 0.955 & 0.295 & 0.857 & 0.830 \\
CricketX & 0.749 & 0.523 & 0.531 & 0.782 & 0.713 & 0.731 & 0.731 & 0.385 & 0.535 & 0.644 & 0.794 & 0.513 & 0.591 & 0.799 & 0.074 & 0.627 & 0.754 \\
DistalPhalanxOutlineCorrect & 0.717 & 0.786 & 0.659 & 0.761 & 0.775 & 0.754 & 0.754 & 0.728 & 0.772 & 0.724 & 0.760 & 0.759 & 0.727 & 0.770 & 0.583 & 0.711 & 0.717 \\
ChlorineConcentration & 0.765 & 0.618 & 0.565 & 0.832 & 0.749 & 0.753 & 0.753 & 0.562 & 0.608 & 0.583 & 0.817 & 0.662 & 0.800 & 0.853 & 0.533 & 0.554 & 0.648 \\
Chinatown & 0.965 & 0.274 & 0.857 & 0.965 & 0.951 & 0.983 & 0.983 & 0.936 & 0.977 & 0.966 & 0.980 & 0.945 & 0.872 & 0.978 & 0.726 & 0.825 & 0.957 \\
GestureMidAirD1 & 0.646 & 0.285 & 0.292 & 0.608 & 0.608 & 0.369 & 0.369 & 0.208 & 0.534 & 0.528 & 0.695 & 0.518 & 0.575 & 0.698 & 0.038 & 0.549 & 0.569 \\
MiddlePhalanxOutlineAgeGroup & 0.461 & 0.344 & 0.526 & 0.636 & 0.656 & 0.630 & 0.630 & 0.617 & 0.534 & 0.577 & 0.535 & 0.558 & 0.522 & 0.545 & 0.571 & 0.578 & 0.500 \\
UMD & 0.993 & 0.681 & 0.368 & 1.000 & 0.993 & 0.986 & 0.986 & 0.910 & 0.960 & 0.771 & 0.988 & 0.842 & 0.949 & 0.990 & 0.333 & 0.835 & 0.993 \\
Crop & 0.734 & 0.388 & 0.341 & 0.756 & 0.722 & 0.742 & 0.742 & 0.710 & 0.670 & 0.760 & 0.738 & 0.687 & 0.618 & 0.743 & 0.042 & 0.489 & 0.665 \\
GesturePebbleZ1 & 0.849 & 0.512 & 0.605 & 0.930 & 0.919 & 0.395 & 0.395 & 0.500 & 0.844 & 0.821 & 0.880 & 0.769 & 0.792 & 0.901 & 0.163 & 0.840 & 0.791 \\
WordSynonyms & 0.688 & 0.335 & 0.451 & 0.676 & 0.691 & 0.531 & 0.531 & 0.422 & 0.568 & 0.557 & 0.561 & 0.470 & 0.599 & 0.617 & 0.219 & 0.506 & 0.649 \\
ArrowHead & 0.743 & 0.360 & 0.429 & 0.857 & 0.766 & 0.737 & 0.737 & 0.771 & 0.717 & 0.630 & 0.843 & 0.678 & 0.784 & 0.838 & 0.303 & 0.689 & 0.703 \\
Wine & 0.537 & 0.519 & 0.611 & 0.870 & 0.815 & 0.778 & 0.778 & 0.500 & 0.519 & 0.556 & 0.611 & 0.500 & 0.541 & 0.722 & 0.500 & 0.744 & 0.574 \\
Coffee & 0.893 & 0.964 & 0.679 & 1.000 & 1.000 & 1.000 & 1.000 & 0.821 & 1.000 & 0.886 & 1.000 & 0.979 & 0.993 & 1.000 & 0.507 & 0.979 & 1.000 \\
Earthquakes & 0.748 & 0.741 & 0.748 & 0.748 & 0.748 & 0.748 & 0.748 & 0.748 & 0.709 & 0.740 & 0.725 & 0.748 & 0.727 & 0.712 & 0.748 & 0.748 & 0.719 \\
Herring & 0.594 & 0.531 & 0.578 & 0.641 & 0.594 & 0.594 & 0.594 & 0.594 & 0.531 & 0.512 & 0.644 & 0.572 & 0.491 & 0.600 & 0.594 & 0.625 & 0.531 \\
Beef & 0.833 & 0.400 & 0.167 & 0.767 & 0.667 & 0.600 & 0.600 & 0.500 & 0.767 & 0.707 & 0.680 & 0.507 & 0.713 & 0.753 & 0.200 & 0.527 & 0.633 \\
MiddlePhalanxOutlineCorrect & 0.467 & 0.512 & 0.519 & 0.838 & 0.825 & 0.818 & 0.818 & 0.753 & 0.744 & 0.752 & 0.795 & 0.796 & 0.755 & 0.826 & 0.570 & 0.743 & 0.698 \\
ECGFiveDays & 0.804 & 0.519 & 0.561 & 1.000 & 1.000 & 0.878 & 0.878 & 0.763 & 0.874 & 0.842 & 0.985 & 0.800 & 0.973 & 0.966 & 0.497 & 0.723 & 0.768 \\
Yoga & 0.834 & 0.672 & 0.691 & 0.887 & 0.837 & 0.791 & 0.791 & 0.830 & 0.786 & 0.753 & 0.837 & 0.741 & 0.856 & 0.867 & 0.536 & 0.626 & 0.837 \\
Adiac & 0.688 & 0.565 & 0.598 & 0.762 & 0.675 & 0.767 & 0.767 & 0.550 & 0.393 & 0.318 & 0.841 & 0.620 & 0.391 & 0.833 & 0.023 & 0.428 & 0.604 \\
MoteStrain & 0.774 & 0.700 & 0.681 & 0.861 & 0.851 & 0.843 & 0.843 & 0.768 & 0.885 & 0.872 & 0.936 & 0.691 & 0.855 & 0.924 & 0.539 & 0.809 & 0.835 \\
Strawberry & 0.951 & 0.946 & 0.935 & 0.962 & 0.954 & 0.965 & 0.965 & 0.916 & 0.952 & 0.959 & 0.975 & 0.958 & 0.959 & 0.980 & 0.643 & 0.911 & 0.941 \\
InsectWingbeatSound & 0.607 & 0.529 & 0.598 & 0.630 & 0.597 & 0.415 & 0.415 & 0.266 & 0.585 & 0.630 & 0.392 & 0.587 & 0.604 & 0.499 & 0.091 & 0.435 & 0.355 \\
DodgerLoopWeekend & 0.826 & 0.638 & 0.804 & 0.964 & NaN & NaN & NaN & 0.732 & 0.974 & 0.983 & 0.904 & 0.978 & 0.978 & 0.952 & 0.739 & 0.954 & 0.949 \\
Meat & 0.917 & 0.433 & 0.667 & 0.950 & 0.950 & 0.883 & 0.883 & 0.900 & 0.913 & 0.787 & 0.803 & 0.787 & 0.893 & 0.990 & 0.333 & 0.970 & 0.933 \\
MelbournePedestrian & 0.876 & 0.718 & 0.207 & 0.959 & 0.944 & 0.949 & 0.949 & 0.741 & 0.813 & 0.884 & 0.912 & 0.840 & 0.863 & 0.909 & 0.100 & 0.730 & 0.791 \\
FaceAll & 0.791 & 0.177 & 0.147 & 0.771 & 0.786 & 0.813 & 0.813 & 0.504 & 0.774 & 0.794 & 0.938 & 0.720 & 0.794 & 0.867 & 0.080 & 0.673 & 0.808 \\
FacesUCR & 0.811 & 0.679 & 0.462 & 0.924 & 0.884 & 0.863 & 0.863 & 0.543 & 0.873 & 0.867 & 0.943 & 0.775 & 0.831 & 0.954 & 0.143 & 0.641 & 0.905 \\
AllGestureWiimoteY & 0.666 & 0.223 & 0.160 & 0.793 & 0.726 & 0.741 & 0.741 & 0.423 & 0.479 & 0.509 & 0.784 & 0.420 & 0.571 & 0.794 & 0.100 & 0.600 & 0.729 \\
ShakeGestureWiimoteZ & 0.960 & 0.020 & 0.080 & 0.940 & 0.920 & 0.860 & 0.860 & 0.760 & 0.580 & 0.756 & 0.884 & 0.516 & 0.548 & 0.880 & 0.100 & 0.864 & 0.860 \\
BME & 0.960 & 0.467 & 0.367 & 0.993 & 0.993 & 0.933 & 0.933 & 0.760 & 0.947 & 0.827 & 0.836 & 0.896 & 0.905 & 0.999 & 0.333 & 0.819 & 0.900 \\
FordB & 0.798 & 0.754 & 0.677 & 0.794 & 0.793 & 0.815 & 0.815 & 0.507 & 0.749 & 0.777 & 0.772 & 0.698 & 0.707 & 0.813 & 0.503 & 0.512 & 0.620 \\
Fish & 0.800 & 0.726 & 0.731 & 0.926 & 0.891 & 0.817 & 0.817 & 0.720 & 0.855 & 0.734 & 0.961 & 0.720 & 0.848 & 0.981 & 0.126 & 0.878 & 0.823 \\
SonyAIBORobotSurface2 & 0.829 & 0.646 & 0.650 & 0.871 & 0.889 & 0.907 & 0.907 & 0.745 & 0.831 & 0.844 & 0.980 & 0.804 & 0.831 & 0.975 & 0.617 & 0.635 & 0.831 \\
FiftyWords & 0.802 & 0.499 & 0.492 & 0.771 & 0.732 & 0.653 & 0.653 & 0.525 & 0.624 & 0.658 & 0.646 & 0.611 & 0.708 & 0.740 & 0.125 & 0.518 & 0.690 \\
ToeSegmentation1 & 0.925 & 0.456 & 0.561 & 0.917 & 0.939 & 0.930 & 0.930 & 0.807 & 0.598 & 0.706 & 0.961 & 0.559 & 0.589 & 0.957 & 0.526 & 0.882 & 0.772 \\
FreezerSmallTrain & 0.902 & 0.704 & 0.500 & 0.870 & 0.933 & 0.979 & 0.979 & 0.920 & 0.739 & 0.676 & 0.683 & 0.688 & 0.686 & 0.832 & 0.500 & 0.917 & 0.753 \\
TwoPatterns & 0.994 & 0.989 & 0.923 & 1.000 & 0.999 & 0.999 & 0.999 & 0.466 & 0.991 & 1.000 & 0.870 & 0.976 & 0.948 & 1.000 & 0.259 & 0.875 & 1.000 \\
ShapeletSim & 0.961 & 0.500 & 0.489 & 1.000 & 0.672 & 0.683 & 0.683 & 0.489 & 0.497 & 0.510 & 0.706 & 0.498 & 0.513 & 0.782 & 0.500 & 0.546 & 0.650 \\
Plane & 0.990 & 0.981 & 0.924 & 1.000 & 0.990 & 1.000 & 1.000 & 0.933 & 0.962 & 0.964 & 1.000 & 0.952 & 0.977 & 1.000 & 0.143 & 1.000 & 1.000 \\
GestureMidAirD3 & 0.369 & 0.085 & 0.162 & 0.292 & 0.285 & 0.177 & 0.177 & 0.154 & 0.317 & 0.368 & 0.326 & 0.278 & 0.382 & 0.340 & 0.038 & 0.275 & 0.323 \\
DiatomSizeReduction & 0.879 & 0.967 & 0.987 & 0.984 & 0.984 & 0.977 & 0.977 & 0.961 & 0.954 & 0.880 & 0.346 & 0.646 & 0.909 & 0.301 & 0.301 & 0.914 & 0.967 \\
CricketZ & 0.731 & 0.459 & 0.397 & 0.792 & 0.708 & 0.713 & 0.713 & 0.403 & 0.501 & 0.651 & 0.810 & 0.484 & 0.629 & 0.809 & 0.062 & 0.643 & 0.754 \\
Lightning7 & 0.726 & 0.575 & 0.562 & 0.863 & 0.795 & 0.685 & 0.685 & 0.411 & 0.647 & 0.696 & 0.825 & 0.559 & 0.616 & 0.827 & 0.260 & 0.608 & 0.726 \\
UWaveGestureLibraryY & 0.738 & 0.547 & 0.648 & 0.719 & 0.710 & 0.641 & 0.641 & 0.348 & 0.626 & 0.676 & 0.642 & 0.639 & 0.699 & 0.666 & 0.121 & 0.497 & 0.634 \\
GunPointAgeSpan & 0.962 & 0.494 & 0.494 & 0.987 & 0.994 & 0.994 & 0.994 & 0.991 & 0.912 & 0.890 & 0.996 & 0.887 & 0.934 & 0.997 & 0.494 & 0.965 & 0.918 \\
DistalPhalanxOutlineAgeGroup & 0.669 & 0.597 & 0.489 & 0.727 & 0.727 & 0.755 & 0.755 & 0.741 & 0.758 & 0.761 & 0.718 & 0.729 & 0.647 & 0.718 & 0.433 & 0.705 & 0.770 \\
SwedishLeaf & 0.923 & 0.894 & 0.899 & 0.941 & 0.914 & 0.923 & 0.923 & 0.738 & 0.884 & 0.902 & 0.967 & 0.841 & 0.845 & 0.963 & 0.064 & 0.837 & 0.792 \\
CBF & 0.960 & 0.761 & 0.830 & 1.000 & 0.983 & 0.998 & 0.998 & 0.898 & 0.959 & 0.977 & 0.994 & 0.908 & 0.869 & 0.996 & 0.332 & 0.896 & 0.997 \\
BeetleFly & 0.900 & 0.400 & 0.700 & 0.900 & 0.800 & 0.800 & 0.800 & 1.000 & 0.900 & 0.620 & 0.910 & 0.630 & 0.880 & 0.850 & 0.500 & 0.790 & 0.700 \\
AllGestureWiimoteZ & 0.537 & 0.221 & 0.116 & 0.746 & 0.723 & 0.689 & 0.689 & 0.447 & 0.375 & 0.396 & 0.692 & 0.287 & 0.439 & 0.726 & 0.100 & 0.516 & 0.643 \\
DodgerLoopDay & 0.438 & 0.237 & 0.200 & 0.562 & NaN & NaN & NaN & 0.200 & 0.312 & 0.487 & 0.143 & 0.305 & 0.160 & 0.150 & 0.160 & 0.593 & 0.500 \\
GunPointOldVersusYoung & 0.981 & 0.508 & 0.524 & 1.000 & 1.000 & 1.000 & 1.000 & 1.000 & 0.922 & 0.923 & 0.989 & 0.926 & 0.941 & 0.989 & 0.524 & 0.975 & 0.838 \\
FordA & 0.936 & 0.913 & 0.914 & 0.936 & 0.928 & 0.930 & 0.930 & 0.568 & 0.896 & 0.928 & 0.914 & 0.863 & 0.816 & 0.937 & 0.510 & 0.555 & 0.555 \\
ItalyPowerDemand & 0.911 & 0.837 & 0.880 & 0.925 & 0.954 & 0.955 & 0.955 & 0.845 & 0.954 & 0.964 & 0.963 & 0.966 & 0.953 & 0.962 & 0.499 & 0.871 & 0.950 \\
ProximalPhalanxOutlineAgeGroup & 0.863 & 0.868 & 0.839 & 0.834 & 0.844 & 0.839 & 0.839 & 0.854 & 0.812 & 0.872 & 0.825 & 0.839 & 0.849 & 0.847 & 0.488 & 0.839 & 0.805 \\
GunPoint & 0.927 & 0.887 & 0.847 & 0.980 & 0.980 & 0.993 & 0.993 & 0.827 & 0.948 & 0.784 & 1.000 & 0.907 & 0.928 & 0.991 & 0.493 & 0.989 & 0.907 \\
ProximalPhalanxTW & 0.712 & 0.800 & 0.712 & 0.824 & 0.771 & 0.800 & 0.800 & 0.780 & 0.777 & 0.791 & 0.761 & 0.775 & 0.767 & 0.773 & 0.341 & 0.784 & 0.761 \\
PickupGestureWiimoteZ & 0.620 & 0.100 & 0.080 & 0.820 & 0.740 & 0.600 & 0.600 & 0.240 & 0.608 & 0.496 & 0.744 & 0.412 & 0.604 & 0.704 & 0.100 & 0.616 & 0.660 \\
SonyAIBORobotSurface1 & 0.729 & 0.542 & 0.589 & 0.903 & 0.902 & 0.899 & 0.899 & 0.724 & 0.690 & 0.729 & 0.958 & 0.655 & 0.692 & 0.961 & 0.429 & 0.725 & 0.725 \\
PowerCons & 0.894 & 0.956 & 0.989 & 0.961 & 0.900 & 0.961 & 0.961 & 0.911 & 0.960 & 0.971 & 0.863 & 0.929 & 0.977 & 0.879 & 0.500 & 0.852 & 0.878 \\
PhalangesOutlinesCorrect & 0.652 & 0.614 & 0.663 & 0.809 & 0.784 & 0.804 & 0.804 & 0.773 & 0.799 & 0.745 & 0.818 & 0.795 & 0.756 & 0.845 & 0.613 & 0.656 & 0.728 \\
BirdChicken & 0.850 & 0.450 & 0.550 & 0.800 & 0.850 & 0.650 & 0.650 & 0.650 & 0.710 & 0.510 & 0.940 & 0.540 & 0.740 & 0.880 & 0.500 & 0.620 & 0.750 \\
ToeSegmentation2 & 0.915 & 0.731 & 0.731 & 0.892 & 0.900 & 0.877 & 0.877 & 0.615 & 0.752 & 0.702 & 0.889 & 0.649 & 0.745 & 0.894 & 0.815 & 0.794 & 0.838 \\
CricketY & 0.746 & 0.531 & 0.521 & 0.749 & 0.728 & 0.718 & 0.718 & 0.467 & 0.582 & 0.639 & 0.793 & 0.521 & 0.598 & 0.810 & 0.085 & 0.652 & 0.744 \\
ElectricDevices & 0.646 & 0.552 & 0.506 & 0.721 & 0.707 & 0.686 & 0.686 & 0.676 & 0.686 & 0.702 & 0.706 & 0.653 & 0.593 & 0.728 & 0.242 & 0.605 & 0.602 \\
DodgerLoopGame & 0.623 & 0.471 & 0.717 & 0.841 & NaN & NaN & NaN & 0.696 & 0.816 & 0.810 & 0.768 & 0.877 & 0.865 & 0.710 & 0.478 & 0.716 & 0.877 \\
Fungi & 0.898 & 0.043 & 0.054 & 0.957 & 1.000 & 0.753 & 0.753 & 0.366 & 0.961 & 0.934 & 0.018 & 0.051 & 0.863 & 0.177 & 0.063 & 0.439 & 0.839 \\
Symbols & 0.936 & 0.864 & 0.694 & 0.976 & 0.963 & 0.916 & 0.916 & 0.786 & 0.808 & 0.754 & 0.955 & 0.644 & 0.836 & 0.893 & 0.174 & 0.798 & 0.950 \\
UWaveGestureLibraryZ & 0.765 & 0.632 & 0.643 & 0.770 & 0.757 & 0.690 & 0.690 & 0.655 & 0.630 & 0.684 & 0.727 & 0.645 & 0.697 & 0.749 & 0.121 & 0.573 & 0.658 \\
ECG200 & 0.760 & 0.830 & 0.790 & 0.920 & 0.940 & 0.880 & 0.880 & 0.830 & 0.816 & 0.884 & 0.888 & 0.838 & 0.914 & 0.874 & 0.640 & 0.874 & 0.770 \\
\bottomrule
\end{tabular}}
\caption{Classification accuracy of methods across 91 UCR datasets. \texttt{MOMENT} without fine-tuning on individual datasets demonstrates promising accuracy.}
\label{tab:classification_results_appendix}
\end{table}

\begin{table}[!htb]
    \centering
    \begin{tabular}{r|c|ccccc|c}
\toprule
 Dataset & \texttt{MOMENT$_0$} & TS2Vec & T-Loss & TNC & TS-TCC & TST & DTW \\
\midrule
ArticularyWordRecognition & 0.990 & 0.987 & 0.943 & 0.973 & 0.953 & 0.977 & 0.987 \\
AtrialFibrillation & 0.200 & 0.200 & 0.133 & 0.133 & 0.267 & 0.067 & 0.200 \\
BasicMotions & 1.000 & 0.975 & 1.000 & 0.975 & 1.000 & 0.975 & 0.975 \\
Cricket & 0.986 & 0.972 & 0.972 & 0.958 & 0.917 & 1.000 & 1.000 \\
DuckDuckGeese & 0.600 & 0.680 & 0.650 & 0.460 & 0.380 & 0.620 & 0.600 \\
EigenWorms & 0.809 & 0.847 & 0.840 & 0.840 & 0.779 & 0.748 & 0.618 \\
Epilepsy & 0.993 & 0.964 & 0.971 & 0.957 & 0.957 & 0.949 & 0.964 \\
ERing & 0.959 & 0.874 & 0.133 & 0.852 & 0.904 & 0.874 & 0.133 \\
EthanolConcentration & 0.357 & 0.308 & 0.205 & 0.297 & 0.285 & 0.262 & 0.323 \\
FaceDetection & 0.633 & 0.501 & 0.513 & 0.536 & 0.544 & 0.534 & 0.529 \\
FingerMovements & 0.490 & 0.480 & 0.580 & 0.470 & 0.460 & 0.560 & 0.530 \\
HandMovementDirection & 0.324 & 0.338 & 0.351 & 0.324 & 0.243 & 0.243 & 0.231 \\
Handwriting & 0.308 & 0.515 & 0.451 & 0.249 & 0.498 & 0.225 & 0.286 \\
Heartbeat & 0.722 & 0.683 & 0.741 & 0.746 & 0.751 & 0.746 & 0.717 \\
JapaneseVowels & 0.716 & 0.984 & 0.989 & 0.978 & 0.930 & 0.978 & 0.949 \\
Libras & 0.850 & 0.867 & 0.883 & 0.817 & 0.822 & 0.656 & 0.870 \\
LSST & 0.411 & 0.537 & 0.509 & 0.595 & 0.474 & 0.408 & 0.551 \\
MotorImagery & 0.500 & 0.510 & 0.580 & 0.500 & 0.610 & 0.500 & 0.500 \\
NATOPS & 0.828 & 0.928 & 0.917 & 0.911 & 0.822 & 0.850 & 0.883 \\
PEMS-SF & 0.896 & 0.682 & 0.676 & 0.699 & 0.734 & 0.740 & 0.711 \\
PenDigits & 0.972 & 0.989 & 0.981 & 0.979 & 0.974 & 0.560 & 0.977 \\
PhonemeSpectra & 0.233 & 0.233 & 0.222 & 0.207 & 0.252 & 0.085 & 0.151 \\
RacketSports & 0.796 & 0.855 & 0.855 & 0.776 & 0.816 & 0.809 & 0.803 \\
SelfRegulationSCP1 & 0.840 & 0.812 & 0.843 & 0.799 & 0.823 & 0.754 & 0.775 \\
SelfRegulationSCP2 & 0.478 & 0.578 & 0.539 & 0.550 & 0.533 & 0.550 & 0.539 \\
SpokenArabicDigits & 0.981 & 0.988 & 0.905 & 0.934 & 0.970 & 0.923 & 0.963 \\
StandWalkJump & 0.400 & 0.467 & 0.333 & 0.400 & 0.333 & 0.267 & 0.200 \\
UWaveGestureLibrary & 0.909 & 0.906 & 0.875 & 0.759 & 0.753 & 0.575 & 0.903 \\
InsectWingbeat & 0.246 & 0.466 & 0.156 & 0.469 & 0.264 & 0.105 & NaN \\
\midrule
\textbf{Mean} & 0.670 & 0.694 & 0.646 & 0.660 & 0.657 & 0.605 & 0.638 \\
\textbf{Median} & 0.722 & 0.683 & 0.676 & 0.746 & 0.751 & 0.620 & 0.664 \\
\textbf{Std.} & 0.274 & 0.255 & 0.296 & 0.267 & 0.263 & 0.294 & 0.296 \\
\textbf{Mean Rank} & 3.466	& 2.862 & 3.603 & 4.362 & 4.121 & 5.069 & 4.429 \\
\textbf{Median Rank} & 3.0 & 2.5 & 4.0 & 5.0 &	5.0 & 5.5 &	4.5 \\
\textbf{Wins/Losses} & 101.5/71.5 & 119.0/54.0 & 97.5/75.5 & 75.5/97.5 & 82.5/90.5 & 55.0/118.0 & 72.0/96.0 \\
\bottomrule
\end{tabular}
\caption{Classification accuracy of methods across 29 UEA datasets. \texttt{MOMENT} without fine-tuning on individual datasets demonstrates promising accuracy.}
\label{tab:multivariate_classification_results_appendix}
\end{table}

\subsection{Classification}
\label{app:classification_app}
\paragraph{Task Description.} The classification problem comprises of learning a mapping $f : \mathcal{T} \to \{1, . . . , C\}$ from a time series to a finite set of classes, using a training dataset of the form $\{(\mathcal{T}_0, c_0), . . . ,(\mathcal{T}_n, c_n)\}$, $c_i \in \{1, . . . , C\}$. One straightforward way to use \texttt{MOMENT} to learn $f$ is to replace its reconstruction head with a linear head that maps patch representations to the $C$ logits. Another way would be to learn $f$ in two stages, as is common in prior work on unsupervised representation learning \citep{TS2Vec,unsupervised-scalable-representation-learning-for-multivariate-ts}: in the first stage, we obtain sequence-level representations for each time series without access to labels. The second stage involves learning any ML classifier (e.g., Support Vector Machine with RBF kernel) using these representations and labels.

\paragraph{Datasets.} We conduct experiments on a subset of 95 datasets from the UCR Classification Archive \citep{UCR-Archive}. These datasets (listed in Table \color{blue}10\color{black}) comprise of equal-length univariate time series shorter than 512 time steps.

\paragraph{Baselines.} We compare \texttt{MOMENT} against 5 \textbf{unsupervised representation learning} methods (TS2Vec \citep{TS2Vec}, TST \citep{transformer-multivariate-ts-representation-learning}, TS-TCC \citep{ts-tcc}, TNC \citep{tnc}, and T-Loss \citep{unsupervised-scalable-representation-learning-for-multivariate-ts}), 8 \textbf{supervised deep learning} (CNN \citep{cnn-for-ts-classification}, Encoder \citep{encoder-for-ts}, FCN \citep{ts-classification-using-dnn-baselines}, MCNN \citep{mcnn-for-ts-classification}, MLP \citep{ts-classification-using-dnn-baselines}, ResNet \citep{ts-classification-using-dnn-baselines}, t-LeNet \citep{data-augmentation-ts-classification-using-cnn}, TWIESN \citep{TWIESN}), 1 \textbf{supervised statistical learning} method DTW \citep{UCR-Archive}), TimesNet \citep{timesnet} and GPT4TS \citep{one-fits-all}.

\paragraph{Experimental Setting.} All models except for \texttt{MOMENT} were trained on each dataset individually, either with labels for supervised deep and statistical learning methods), or without labels for representation learning methods. We collect baseline results for deep learning methods from \citet{dl-for-ts-classification}, representation learning methods from \citet{TS2Vec}, and DTW from \citet{UCR-Archive}. We report accuracy as the evaluation metric.

\paragraph{Hyperparameters.} The hyperparameters used for evaluating classification experiments are shown in Table \ref{tab:hyperparameters-classification}.

\begin{table}[htb!]
\centering
\resizebox{0.7\columnwidth}{!}{
\begin{tabular}{c|rl}
\toprule
\textbf{Model} & \multicolumn{2}{c}{\textbf{Hyper-parameters}} \\ \midrule
\multirow{3}{*}{$\mathtt{MOMENT_{0}}$} & \texttt{sequence length} : & $512$ \\
 & \texttt{patch length} : & $8$ \\
 & \texttt{patch stride length} : & $8$ \\ \midrule
\multirow{6}{*}{$\mathtt{SVM}$} & \texttt{C} : & $\{0.0001, 0.001, 0.01, 0.1, 1, 10, 100, 1000, 10000\}$ \\
 & \texttt{kernel} : & RBF \\
 & \texttt{degree} : & $3$ \\
 & \texttt{cache size} : & $200$ \\ 
 & \texttt{max iterations} : & $10000000$ \\ 
 & \texttt{decision function shape} : & One versus rest \\ \midrule
\end{tabular}}
\caption{Hyper-parameter values for classification.}
\label{tab:hyperparameters-classification}
\end{table}

\subsection{Anomaly Detection}

\paragraph{Task Description.} Given a time series $\mathcal{T}$, anomaly detection is a binary classification problem, where the goal is to detect whether a time step $x_i$ is indicative of an anomaly or not. As shown in Fig. \color{blue}4 (v)\color{black}, to detect anomalies in $\mathcal{T}$, we retain \texttt{MOMENT}’s reconstruction head and use it to reconstruct the input time series. Then, time steps where observations and predictions differ beyond a certain threshold are classified as anomalies\footnote{Estimating good thresholds for anomaly detection is
beyond the scope of this study and an active area of research \citep{goswami2023unsupervised,anomaly-detection-in-ts-evaluation}.}.

\paragraph{Datasets.} We conduct experiments on a subset of 46 univariate time series from the UCR Anomaly Archive \citep{ts-anomaly-detection-benchmarks-are-flawed}, as enumerated in Table \color{blue}11\color{black}. When choosing the subset of time series, we prioritized coverage over different domains and data sources represented in the archive.

\paragraph{Baselines.} We compare \texttt{MOMENT} with 2 state-of-the-art anomaly detection methods DGHL \citep{dghl} and Anomaly Transformer \citep{anomaly-transformer} along with TimesNet and GPT4TS. We also include $k$-Nearest Neighbors (with $k =$ 5) \citep{efficient-algorithms-for-mining-outliers-from-datasets}, a classical anomaly detection method in our experiments. In the zero-shot setting, we compare \texttt{MOMENT} to randomly initialized DGHL ($\mathtt{DGHL_0}$)\footnote{Randomly initialized DGHL is not a trivial zero-shot baseline, since it performs gradient descent to find the best latent $z$ that minimizes reconstruction error during inference time \citep{dghl}.} and $k$-NN.

\paragraph{Experimental Setting.} All algorithms use a fixed anomaly detection window size (= 512). Based on prior work \citep{timesnet,one-fits-all}, we use the mean squared error between predictions and observations as the anomaly criterion\footnote{To ensure a fair comparison, we do not use Anomaly Transformer’s joint criterion as the anomaly score. We believe that this might put the Anomaly Transformer at some disadvantage in our experiments.}. Following prior work \citep{goswami2023unsupervised}, we downsample all time series longer than 2560 timesteps by a factor of 10 to speed up the training and evaluation process.

We report two anomaly detection metrics: adjusted best $F_1$ which is frequently used in practice \citep{goswami2023unsupervised,dghl}, and the recently proposed volume under ROC surface (VUS-ROC) metric \citep{VUS-ROC}. For both metrics, higher scores are better.

\paragraph{Hyperparameters.} The hyperparameters used for training all models in our anomaly detection experiments are shown in Table \ref{tab:hyperparameters-anomaly-detection}.

\begin{table}[htb!]
\centering
\resizebox{0.9\columnwidth}{!}{
\begin{tabular}{r|rl}
\toprule
\textbf{Model} & \multicolumn{2}{c}{\textbf{Hyper-parameters}} \\ \midrule
\multirow{3}{*}{$\mathtt{MOMENT_{0}}$} & \texttt{sequence length} : & $512$ \\
 & \texttt{patch length} : & $8$ \\
 & \texttt{patch stride length} : & $8$ \\ \midrule
\multirow{4}{*}{$\mathtt{MOMENT_{LP}}$} & \texttt{sequence length} : & $512$ \\
 & \texttt{patch length} : & $8$ \\
 & \texttt{patch stride length} : & $8$ \\
 & \texttt{initial lr} : & $5e-5$ \\ \midrule
\multirow{8}{*}{\texttt{Anomaly Transformer}} & \texttt{sequence length} : & $512$ \\
 & \texttt{number of channels} : & $1$ \\
 & \texttt{k} : & $3$ \\
 & \texttt{anomaly ratio} : & $4.00$ \\
 & \texttt{model dimensions} : & $512$ \\
 & \texttt{number of heads} : & $8$ \\
 & \texttt{embedding layers} : & $3$ \\
 & \texttt{dimension of feedforward layer} : & $512$ \\ \midrule
\multirow{10}{*}{\texttt{DGHL}} & \texttt{sequence length} : & $512$ \\
 & \texttt{number of channels} : & $1$ \\
 & \texttt{hidden multiplier} : & $32$ \\
 & \texttt{max filters} : & $256$ \\
 & \texttt{kernel multiplier} : & $1$ \\
 & \texttt{sub-windows} : & $4$ \\
 & \texttt{size of latent z vector} : & $50$ \\
 & \texttt{number of iteration in the Langevyn dynamics inference formula} : & $100$ \\
 & \texttt{z step size} : & $0.1$ \\
 & \texttt{noise std} : & $0.001$ \\ \midrule
\multirow{5}{*}{\texttt{GPT4TS}} & \texttt{sequence length} : & $512$ \\
 & \texttt{gpt layers} : & $3$ \\
 & \texttt{patch length} : & $1$ \\
 & \texttt{patch stride length} : & $1$ \\
 & \texttt{transformer backbone} : & GPT-2 \\ \midrule
\multirow{5}{*}{\texttt{TimesNet}} & \texttt{sequence length} : & $512$ \\
 & \texttt{dimension of model} : & $16$ \\
 & \texttt{dimension of feedforward layer} : & $16$ \\
 & \texttt{top k} : & $3$ \\ 
 & \texttt{number of kernels} : & $6$ \\ \midrule
\texttt{$k$-NN} & \texttt{k} : & $5$ \\ \midrule

\end{tabular}}
\caption{Hyperparameter values for anomaly detection.}
\label{tab:hyperparameters-anomaly-detection}
\end{table}

\subsection{Imputation}

\paragraph{Task Description.} Consider a time series $\mathcal{T} = [x_1, . . . , x_L]$ and an observation mask $\mathcal{M} = [m_1, . . . , m_L]$, where $m_i =$ 0 if $x_i$ is missing and $m_i =$ 1 if $x_i$ is observed. Then imputation is the task of estimating the missing values $\mathcal{T}$ by exploiting its observed values. We treat a patch as observed only if all its time steps are observed. For the remaining patches, we replace their patch embeddings with \texttt{[MASK]} and use \texttt{MOMENT}’s default reconstruction head to impute its values (Fig. \color{blue}4 (iv)\color{black}).

\paragraph{Datasets.} We evaluate imputation performance on 6 real-world datasets from domains where missing data is a common problem: 4 subsets of Electricity Transformer Temperature (ETT), Weather, and Electricity \citep{timesnet,one-fits-all}.

\paragraph{Baselines.} We compare the two variants of \texttt{MOMENT} with 3 state-of-the-art deep learning methods, TimesNet, FPT, and DGHL; and 3 statistical interpolation methods, Cubic Spline, Linear, and 1-D Nearest Neighbor interpolation.

\paragraph{Experimental Setting.} To evaluate the models’ ability to interpolate missing values, we randomly mask contiguous sub-sequences of length 8. Instead of masking contiguous sub-sequences, previous studies \citep{timesnet,one-fits-all} mask individual time points, making the imputation task much easier. The results from prior studies are shown in Table \ref{tab:imputation_full}. We observe that the statistical methods perform similarly to transformer methods, owing to the ease of the task. For our experiments involving randomly masking patches of length 8, our results are shown in Table \ref{tab:imputation_appendix}. We measure the imputation performance of models using mean squared error, over 4 different masking rates: 12.5\%, 25\%, 37.5\%, and 50\%.
f
\paragraph{Hyperparameters.} The hyperparameters used for training all models in our imputation experiments are shown in Table \ref{tab:hyperparameters-imputation}.

\begin{table}[htb!]
\centering
\resizebox{0.5\columnwidth}{!}{
\begin{tabular}{r|rl}
\toprule
\textbf{Model} & \multicolumn{2}{c}{\textbf{Hyper-parameters}} \\ \midrule
\multirow{3}{*}{$\mathtt{MOMENT_{0}}$} & \texttt{sequence length} : & $512$ \\
 & \texttt{patch length} : & $8$ \\
 & \texttt{patch stride length} : & $8$ \\ \midrule
\multirow{4}{*}{$\mathtt{MOMENT_{LP}}$} & \texttt{sequence length} : & $512$ \\
 & \texttt{patch length} : & $8$ \\
 & \texttt{patch stride length} : & $8$ \\
 & \texttt{initial lr} : & $0.0001$ \\ \midrule
\multirow{6}{*}{\texttt{GPT4TS}} & \texttt{sequence length} : & $512$ \\
 & \texttt{gpt layers} : & $3$ \\
 & \texttt{patch length} : & $1$ \\
 & \texttt{patch stride length} : & $1$ \\
 & \texttt{transformer backbone} : & GPT-2 \\
 & \texttt{dimension of feedforward layer} : & $16$ \\ \midrule
\multirow{5}{*}{\texttt{TimesNet}} & \texttt{sequence length} : & $512$ \\
 & \texttt{dimension of model} : & $64$ \\
 & \texttt{dimension of feedforward layer} : & $64$ \\
 & \texttt{top k} : & $3$ \\ 
 & \texttt{number of kernels} : & $6$ \\ \midrule
\end{tabular}}
\caption{Hyperparameter values for imputation.}
\label{tab:hyperparameters-imputation}
\end{table}

\begin{table}[htb!]
\centering
\resizebox{\linewidth}{!}{
\begin{tabular}{cc|cc|cc|cc|cc|cc|cc|cc|cc|cc|cc|cc|cc|cc|cc|cc}
\toprule
\multicolumn{2}{c|}{Methods} 
&\multicolumn{2}{c|}{GPT4TS} & \multicolumn{2}{c|}{TimesNet}& \multicolumn{2}{c|}{PatchTST}&\multicolumn{2}{c|}{ETSformer}&\multicolumn{2}{c|}{LightTS}&\multicolumn{2}{c|}{DLinear}&\multicolumn{2}{c|}{FEDformer}&\multicolumn{2}{c|}{Stationary}&\multicolumn{2}{c|}{Autoformer}&\multicolumn{2}{c|}{Informer}&\multicolumn{2}{c|}{Reformer}&\multicolumn{2}{c|}{Naive}&\multicolumn{2}{c|}{Linear}&\multicolumn{2}{c|}{Nearest}&\multicolumn{2}{c|}{Cubic} \\
Dataset & Mask Ratio & MSE & MAE & MSE & MAE & MSE & MAE & MSE & MAE & MSE & MAE & MSE & MAE & MSE & MAE & MSE & MAE & MSE & MAE & MSE & MAE & MSE & MAE & MSE & MAE & MSE & MAE & MSE & MAE & MSE & MAE \\

\midrule
\multirow{5}{*}{\rotatebox{90}{$ETTm1$}}
& 12.5\% &{\bf 0.017}&{\bf0.085}&0.023&0.101&0.041&0.130&0.096&0.229&0.093&0.206&0.080&0.193&0.052&0.166&0.032&0.119&0.046&0.144&0.063&0.180&0.042&0.146 & 0.059 & 0.145 & 0.034 & 0.109 & 0.055 & 0.140 & 0.052 & 0.135 \\
& 25\% &{\bf0.022}&{\bf0.096}&0.023&0.101&0.044&0.135&0.096&0.229&0.093&0.206&0.080&0.193&0.052&0.166&0.032&0.119&0.046&0.144&0.063&0.180&0.042&0.146 & 0.066 & 0.153 & 0.036 & 0.112 & 0.056 & 0.142 & 0.060 & 0.142 \\
& 37.5\% &{\bf0.029}&{\bf0.111}&0.029&0.111&0.049&0.143&0.133&0.271&0.113&0.231&0.103&0.219&0.069&0.191&0.039&0.131&0.057&0.161&0.079&0.200&0.063&0.182 & 0.077 & 0.164 & 0.038 & 0.117 & 0.060 & 0.146 & 0.071 & 0.151\\
& 50\% &0.040&0.128&{\bf0.036}&{\bf0.124}&0.055&0.151&0.186&0.323&0.134&0.255&0.132&0.248&0.089&0.218&0.047&0.145&0.067&0.174&0.093&0.218&0.082&0.208 & 0.094 & 0.178 & 0.042 & 0.123 & 0.066 & 0.153 & 0.100 & 0.164 \\
& Avg &0.028&{\bf0.105}&{\bf0.027}&0.107&0.047&0.140&0.120&0.253&0.104&0.218&0.093&0.206&0.062&0.177&0.036&0.126&0.051&0.150&0.071&0.188&0.055&0.166 & 0.074 & 0.160 & 0.038 & 0.115 & 0.059 & 0.145 & 0.071 & 0.148\\
\midrule

\multirow{5}{*}{\rotatebox{90}{$ETTm2$}}
& 12.5\% &{\bf0.017}&{\bf0.076}&0.018&0.080&0.026&0.094&0.108&0.239&0.034&0.127&0.062&0.166&0.056&0.159&0.021&0.088&0.023&0.092&0.133&0.270&0.108&0.228 & 0.038 & 0.095 & 0.023 & 0.077 & 0.035 & 0.091 & 0.033 & 0.097 \\
& 25\% &{\bf0.020}&{\bf0.080}&0.020&0.085&0.028&0.099&0.164&0.294&0.042&0.143&0.085&0.196&0.080&0.195&0.024&0.096&0.026&0.101&0.135&0.272&0.136&0.262 & 0.041 & 0.100 & 0.025 & 0.081 & 0.036 & 0.093 & 0.039 & 0.103\\
& 37.5\% &{\bf0.022}&{\bf0.087}&0.023&0.091&0.030&0.104&0.237&0.356&0.051&0.159&0.106&0.222&0.110&0.231&0.027&0.103&0.030&0.108&0.155&0.293&0.175&0.300& 0.046 & 0.106 & 0.027 & 0.085 & 0.038 & 0.095 & 0.047 & 0.111 \\
& 50\% &{\bf0.025}&{\bf0.095}&0.026&0.098&0.034&0.110&0.323&0.421&0.059&0.174&0.131&0.247&0.156&0.276&0.030&0.108&0.035&0.119&0.200&0.333&0.211&0.329& 0.051 & 0.115 & 0.030 & 0.090 & 0.041 & 0.100 & 0.062 & 0.122 \\
& Avg &{\bf0.021}&{\bf0.084}&0.022&0.088&0.029&0.102&0.208&0.327&0.046&0.151&0.096&0.208&0.101&0.215&0.026&0.099&0.029&0.105&0.156&0.292&0.157&0.280& 0.044 & 0.104 & 0.026 & 0.083 & 0.038 & 0.095 & 0.045 & 0.109 \\
\midrule

\multirow{5}{*}{\rotatebox{90}{$ETTh1$}}
& 12.5\% &{\bf0.043}&{\bf0.140}&0.057&0.159&0.093&0.201&0.126&0.263&0.240&0.345&0.151&0.267&0.070&0.190&0.060&0.165&0.074&0.182&0.114&0.234&0.074&0.194& 0.211 & 0.275 & 0.083 & 0.183 & 0.181 & 0.260 & 0.107 & 0.207 \\
& 25\% &{\bf0.054}&{\bf0.156}&0.069&0.178&0.107&0.217&0.169&0.304&0.265&0.364&0.180&0.292&0.106&0.236&0.080&0.189&0.090&0.203&0.140&0.262&0.102&0.227& 0.259 & 0.298 & 0.098 & 0.197 & 0.192 & 0.266 & 0.127 & 0.224 \\
& 37.5\% &{\bf0.072}&{\bf0.180}&0.084&0.196&0.120&0.230&0.220&0.347&0.296&0.382&0.215&0.318&0.124&0.258&0.102&0.212&0.109&0.222&0.174&0.293&0.135&0.261& 0.323 & 0.326 & 0.119 & 0.215 & 0.215 & 0.277 & 0.160 & 0.245 \\
& 50\% &0.107&0.216&{\bf0.102}&{\bf0.215}&0.141&0.248&0.293&0.402&0.334&0.404&0.257&0.347&0.165&0.299&0.133&0.240&0.137&0.248&0.215&0.325&0.179&0.298& 0.423 & 0.366 & 0.158 & 0.242 & 0.257 & 0.297 & 0.235 & 0.279 \\
& Avg &{\bf0.069}&{\bf0.173}&0.078&0.187&0.115&0.224&0.202&0.329&0.284&0.373&0.201&0.306&0.117&0.246&0.094&0.201&0.103&0.214&0.161&0.279&0.122&0.245& 0.304 & 0.317 & 0.114 & 0.209 & 0.211 & 0.275 & 0.157 & 0.239 \\
\midrule

\multirow{5}{*}{\rotatebox{90}{$ETTh2$}}
& 12.5\% &{\bf0.039}&{\bf0.125}&0.040&0.130&0.057&0.152&0.187&0.319&0.101&0.231&0.100&0.216&0.095&0.212&0.042&0.133&0.044&0.138&0.305&0.431&0.163&0.289& 0.090 & 0.167 & 0.058 & 0.134 & 0.085 & 0.162 & 0.091 & 0.172 \\
& 25\% &{\bf0.044}&{\bf0.135}&0.046&0.141&0.061&0.158&0.279&0.390&0.115&0.246&0.127&0.247&0.137&0.258&0.049&0.147&0.050&0.149&0.322&0.444&0.206&0.331& 0.097 & 0.176 & 0.060 & 0.138 & 0.088 & 0.165 & 0.101 & 0.179 \\
& 37.5\% &{\bf0.051}&{\bf0.147}&0.052&0.151&0.067&0.166&0.400&0.465&0.126&0.257&0.158&0.276&0.187&0.304&0.056&0.158&0.060&0.163&0.353&0.462&0.252&0.370& 0.105 & 0.185 & 0.064 & 0.144 & 0.091 & 0.169 & 0.118 & 0.190 \\
& 50\% &{\bf0.059}&{\bf0.158}&0.060&0.162&0.073&0.174&0.602&0.572&0.136&0.268&0.183&0.299&0.232&0.341&0.065&0.170&0.068&0.173&0.369&0.472&0.316&0.419& 0.118 & 0.199 & 0.071 & 0.153 & 0.097 & 0.176 & 0.150 & 0.205 \\
& Avg &{\bf0.048}&{\bf0.141}&0.049&0.146&0.065&0.163&0.367&0.436&0.119&0.250&0.142&0.259&0.163&0.279&0.053&0.152&0.055&0.156&0.337&0.452&0.234&0.352& 0.102 & 0.182 & 0.063 & 0.142 & 0.090 & 0.168 & 0.115 & 0.186 \\
\midrule

\multirow{5}{*}{\rotatebox{90}{$ECL$}}
& 12.5\% &{\bf0.080}&{\bf0.194}&0.085&0.202&0.055&0.160&0.196&0.321&0.102&0.229&0.092&0.214&0.107&0.237&0.093&0.210&0.089&0.210&0.218&0.326&0.190&0.308& 0.214 & 0.293 & 0.079 & 0.182 & 0.181 & 0.271 & 0.091 & 0.196 \\
& 25\% &{\bf0.087}&{\bf0.203}&0.089&0.206&0.065&0.175&0.207&0.332&0.121&0.252&0.118&0.247&0.120&0.251&0.097&0.214&0.096&0.220&0.219&0.326&0.197&0.312& 0.266 & 0.324 & 0.094 & 0.199 & 0.194 & 0.280 & 0.115 & 0.217 \\
& 37.5\%  &0.094&{\bf0.211}&{\bf0.094}&0.213&0.076&0.189&0.219&0.344&0.141&0.273&0.144&0.276&0.136&0.266&0.102&0.220&0.104&0.229&0.222&0.328&0.203&0.315& 0.339 & 0.366 & 0.117 & 0.223 & 0.220 & 0.296 & 0.152 & 0.245\\
& 50\% &0.101&{\bf0.220}&{\bf0.100}&0.221&0.091&0.208&0.235&0.357&0.160&0.293&0.175&0.305&0.158&0.284&0.108&0.228&0.113&0.239&0.228&0.331&0.210&0.319& 0.447 & 0.424 & 0.156 & 0.259 & 0.264 & 0.324 & 0.222 & 0.286 \\
& Avg &{\bf0.090}&{\bf0.207}&0.092&0.210&0.072&0.183&0.214&0.339&0.131&0.262&0.132&0.260&0.130&0.259&0.100&0.218&0.101&0.225&0.222&0.328&0.200&0.313& 0.316 & 0.352 & 0.112 & 0.216 & 0.215 & 0.293 & 0.145 & 0.236 \\
\midrule

\multirow{5}{*}{\rotatebox{90}{$Weather$}}
& 12.5\% &0.026&0.049&{\bf0.025}&{\bf0.045}&0.029&0.049&0.057&0.141&0.047&0.101&0.039&0.084&0.041&0.107&0.027&0.051&0.026&0.047&0.037&0.093&0.031&0.076& 0.041 & 0.042 & 0.026 & 0.031 & 0.038 & 0.041 & 0.038 & 0.036 \\
& 25\% &{\bf0.028}&{\bf0.052}&0.029&0.052&0.031&0.053&0.065&0.155&0.052&0.111&0.048&0.103&0.064&0.163&0.029&0.056&0.030&0.054&0.042&0.100&0.035&0.082& 0.045 & 0.045 & 0.027 & 0.032 & 0.040 & 0.041 & 0.043 & 0.039 \\
& 37.5\% &0.033&0.060&{\bf0.031}&{\bf0.057}&0.035&0.058&0.081&0.180&0.058&0.121&0.057&0.117&0.107&0.229&0.033&0.062&0.032&0.060&0.049&0.111&0.040&0.091& 0.048 & 0.049 & 0.030 & 0.034 & 0.042 & 0.043 & 0.055 & 0.043 \\
& 50\% &0.037&0.065&{\bf0.034}&{\bf0.062}&0.038&0.063&0.102&0.207&0.065&0.133&0.066&0.134&0.183&0.312&0.037&0.068&0.037&0.067&0.053&0.114&0.046&0.099& 0.054 & 0.054 & 0.033 & 0.037 & 0.044 & 0.045 & 0.069 & 0.048 \\
& Avg &0.031&0.056&{\bf0.030}&{\bf0.054}&0.060&0.144&0.076&0.171&0.055&0.117&0.052&0.110&0.099&0.203&0.032&0.059&0.031&0.057&0.045&0.104&0.038&0.087& 0.047 & 0.048 & 0.029 & 0.034 & 0.041 & 0.043 & 0.051 & 0.042 \\
\bottomrule
\end{tabular}}
\caption{Results for the imputation task in the time-steps missing at random setting. Results averaged across 4 different masking rates: \{12.5\%, 25\%, 37.5\%, 50\%\}. Statistical interpolation methods such as forward and backward fill (naive), linear, nearest, and cubic interpolation perform better than many transformer-based baselines. Therefore, we consider the much harder, patches missing at random setting in our experiments.} 
\label{tab:imputation_full}
\end{table}

\begin{table*}[htb!]
\centering
\resizebox{\linewidth}{!}{
\begin{tabular}{c|c|cccc|cccc|cccccccc}
\toprule
\multirow{2}{*}{\textbf{Dataset}} & \multirow{2}{*}{\textbf{Mask Ratio}} & \multicolumn{2}{c}{$\mathtt{MOMENT_{0}}$} & \multicolumn{2}{c}{$\mathtt{MOMENT_{LP}}$} & \multicolumn{2}{c}{\textbf{GPT4TS}} & \multicolumn{2}{c}{\textbf{TimesNet}} & \multicolumn{2}{c}{\textbf{Naive}} & \multicolumn{2}{c}{\textbf{Linear}} & \multicolumn{2}{c}{\textbf{Nearest}} & \multicolumn{2}{c}{\textbf{Cubic}} \\
 &  & \textbf{MSE} & \textbf{MAE} & \textbf{MSE} & \textbf{MAE} & \textbf{MSE} & \textbf{MAE} & \textbf{MSE} & \textbf{MAE} & \multicolumn{1}{c}{\textbf{MSE}} & \multicolumn{1}{c}{\textbf{MAE}} & \multicolumn{1}{c}{\textbf{MSE}} & \multicolumn{1}{c}{\textbf{MAE}} & \multicolumn{1}{c}{\textbf{MSE}} & \multicolumn{1}{c}{\textbf{MAE}} & \multicolumn{1}{c}{\textbf{MSE}} & \multicolumn{1}{c}{\textbf{MAE}} \\ \midrule
\multirow{5}{*}{Weather} 
 & 0.125 &0.085 & 0.131 & 0.033 & 0.073 & 0.036 & 0.076 & 0.035 & 0.096 & 0.105 & 0.089 & 0.050 & 0.055 & 0.069 & 0.067 & 0.373 & 0.115 \\
 & 0.250 & 0.079 & 0.130 & 0.036 & 0.078 & 0.030 & 0.071 & 0.037 & 0.100 & 0.127 & 0.104 & 0.075 & 0.065 & 0.094 & 0.076 & 0.297 & 0.125 \\
 & 0.375 & 0.081 & 0.128 & 0.034 & 0.075 & 0.030 & 0.070 & 0.035 & 0.099 & 0.120 & 0.111 & 0.066 & 0.069 & 0.082 & 0.080 & 0.904 & 0.176 \\
 & 0.500 & 0.081 & 0.129 & 0.035 & 0.075 & 0.026 & 0.069 & 0.035 & 0.098 & 0.124 & 0.127 & 0.066 & 0.078 & 0.086 & 0.090 & 0.831 & 0.197 \\ \midrule
 & Mean & 0.082 & 0.130 & 0.035 & 0.075 & 0.031 & 0.071 & 0.036 & 0.098 & 0.119 & 0.108 & 0.065 & 0.067 & 0.083 & 0.078 & 0.601 & 0.153 \\ \midrule
\multirow{5}{*}{ETTh1} 
 & 0.125 & 0.430 & 0.417 & 0.160 & 0.239 & 0.183 & 0.242 & 0.158 & 0.254 & 1.008 & 0.602 & 0.583 & 0.466 & 0.712 & 0.523 & 0.985 & 0.661 \\
 & 0.250 & 0.373 & 0.392 & 0.142 & 0.238 & 0.278 & 0.267 & 0.154 & 0.261 & 1.311 & 0.686 & 0.833 & 0.540 & 0.954 & 0.581 & 1.433 & 0.772 \\
 & 0.375 & 0.398 & 0.398 & 0.121 & 0.228 & 0.232 & 0.263 & 0.195 & 0.274 & 1.317 & 0.703 & 0.843 & 0.572 & 0.973 & 0.613 & 2.615 & 1.028 \\
 & 0.500 & 0.408 & 0.403 & 0.132 & 0.231 & 0.213 & 0.243 & 0.192 & 0.267 & 1.103 & 0.643 & 0.840 & 0.559 & 0.963 & 0.601 & 3.681 & 1.204 \\ \midrule
 & Mean & 0.402 & 0.403 & 0.139 & 0.234 & 0.227 & 0.254 & 0.175 & 0.264 & 1.185 & 0.658 & 0.775 & 0.534 & 0.900 & 0.579 & 2.178 & 0.916 \\ \midrule
\multirow{5}{*}{ETTh2} 
 & 0.125 & 0.122 & 0.235 & 0.051 & 0.150 & 0.115 & 0.215 & 0.163 & 0.277 & 0.196 & 0.285 & 0.105 & 0.208 & 0.134 & 0.225 & 0.452 & 0.404 \\
 & 0.250 & 0.127 & 0.237 & 0.079 & 0.177 & 0.114 & 0.216 & 0.161 & 0.280 & 0.210 & 0.291 & 0.120 & 0.220 & 0.154 & 0.240 & 0.831 & 0.524 \\
 & 0.375 & 0.124 & 0.237 & 0.056 & 0.155 & 0.110 & 0.216 & 0.170 & 0.291 & 0.229 & 0.310 & 0.142 & 0.243 & 0.175 & 0.262 & 1.571 & 0.672 \\
 & 0.500 & 0.127 & 0.242 & 0.056 & 0.154 & 0.098 & 0.207 & 0.186 & 0.295 & 0.265 & 0.329 & 0.171 & 0.264 & 0.199 & 0.279 & 4.823 & 0.966 \\ \midrule
 & Mean & 0.125 & 0.238 & 0.061 & 0.159 & 0.109 & 0.213 & 0.170 & 0.286 & 0.225 & 0.304 & 0.135 & 0.234 & 0.166 & 0.252 & 1.920 & 0.641 \\ \midrule
\multirow{5}{*}{ETTm1} 
 & 0.125 & 0.179 & 0.278 & 0.069 & 0.170 & 0.078 & 0.147 & 0.089 & 0.200 & 0.273 & 0.293 & 0.094 & 0.183 & 0.147 & 0.217 & 0.334 & 0.345 \\
 & 0.250 & 0.206 & 0.290 & 0.071 & 0.169 & 0.071 & 0.144 & 0.080 & 0.194 & 0.395 & 0.341 & 0.114 & 0.202 & 0.171 & 0.234 & 0.539 & 0.424 \\
 & 0.375 & 0.209 & 0.289 & 0.069 & 0.163 & 0.076 & 0.146 & 0.091 & 0.199 & 0.475 & 0.378 & 0.188 & 0.242 & 0.257 & 0.274 & 0.842 & 0.528 \\
 & 0.500 & 0.215 & 0.294 & 0.086 & 0.169 & 0.081 & 0.149 & 0.088 & 0.197 & 0.679 & 0.448 & 0.265 & 0.291 & 0.346 & 0.316 & 1.715 & 0.680 \\ \midrule
 & Mean & 0.202 & 0.288 & 0.074 & 0.168 & 0.076 & 0.146 & 0.087 & 0.198 & 0.455 & 0.365 & 0.165 & 0.229 & 0.230 & 0.260 & 0.858 & 0.494 \\ \midrule
\multirow{5}{*}{ETTm2} 
 & 0.125 & 0.076 & 0.183 & 0.032 & 0.108 & 0.043 & 0.126 & 0.128 & 0.233 & 0.087 & 0.164 & 0.049 & 0.117 & 0.062 & 0.132 & 0.237 & 0.262 \\
 & 0.250 & 0.084 & 0.187 & 0.029 & 0.105 & 0.046 & 0.129 & 0.101 & 0.207 & 0.104 & 0.182 & 0.057 & 0.132 & 0.073 & 0.146 & 0.373 & 0.309 \\
 & 0.375 & 0.076 & 0.181 & 0.032 & 0.109 & 0.059 & 0.137 & 0.116 & 0.225 & 0.115 & 0.196 & 0.063 & 0.141 & 0.078 & 0.154 & 0.626 & 0.376 \\
 & 0.500 & 0.077 & 0.183 & 0.031 & 0.110 & 0.059 & 0.140 & 0.103 & 0.212 & 0.144 & 0.222 & 0.080 & 0.162 & 0.102 & 0.177 & 0.899 & 0.477 \\ \midrule
 & Mean & 0.078 & 0.184 & 0.031 & 0.108 & 0.052 & 0.133 & 0.112 & 0.220 & 0.113 & 0.191 & 0.062 & 0.138 & 0.079 & 0.152 & 0.534 & 0.356 \\ \midrule
\multirow{5}{*}{Electricity} 
 & 0.125 & 0.251 & 0.370 & 0.095 & 0.211 & 0.069 & 0.180 & 0.126 & 0.248 & 1.350 & 0.818 & 0.458 & 0.466 & 0.608 & 0.492 & 0.924 & 0.610 \\
 & 0.250 & 0.249 & 0.372 & 0.093 & 0.211 & 0.073 & 0.184 & 0.121 & 0.246 & 1.447 & 0.857 & 0.654 & 0.554 & 0.815 & 0.582 & 1.619 & 0.769 \\
 & 0.375 & 0.250 & 0.371 & 0.094 & 0.211 & 0.071 & 0.181 & 0.125 & 0.248 & 1.518 & 0.888 & 0.836 & 0.637 & 1.031 & 0.675 & 2.507 & 0.959 \\
 & 0.500 & 0.250 & 0.371 & 0.092 & 0.210 & 0.075 & 0.185 & 0.126 & 0.249 & 1.581 & 0.915 & 1.002 & 0.712 & 1.239 & 0.766 & 3.978 & 1.213 \\ \midrule
 & Mean & 0.250 & 0.371 & 0.094 & 0.211 & 0.072 & 0.183 & 0.124 & 0.248 & 1.474 & 0.869 & 0.737 & 0.592 & 0.923 & 0.629 & 2.257 & 0.888 \\ \toprule
\end{tabular}}
\caption{\textbf{Imputation Results.} \texttt{MOMENT} achieves state-of-the-art imputation results in both zero-shot and linear probe fine-tuning settings.}
\label{tab:imputation_appendix}
\end{table*}

\subsection{What is \texttt{MOMENT} Learning?}

\begin{figure*}[htb!]
\centering
\setlength{\tabcolsep}{0pt}
\begin{tabular}{ccccc}
\includegraphics[width=0.18\linewidth]{figures/trend_artifacts_pca.pdf} & 
\includegraphics[width=0.18\linewidth]{figures/amplitude_artifacts_pca.pdf} & 
\includegraphics[width=0.18\linewidth]{figures/frequency_artifacts_pca.pdf} & 
\includegraphics[width=0.18\linewidth]{figures/baseline_artifacts_pca.pdf} & 
\includegraphics[width=0.18\linewidth]{figures/autocorrelation_artifacts_pca.pdf} \\
\includegraphics[width=0.18\linewidth]{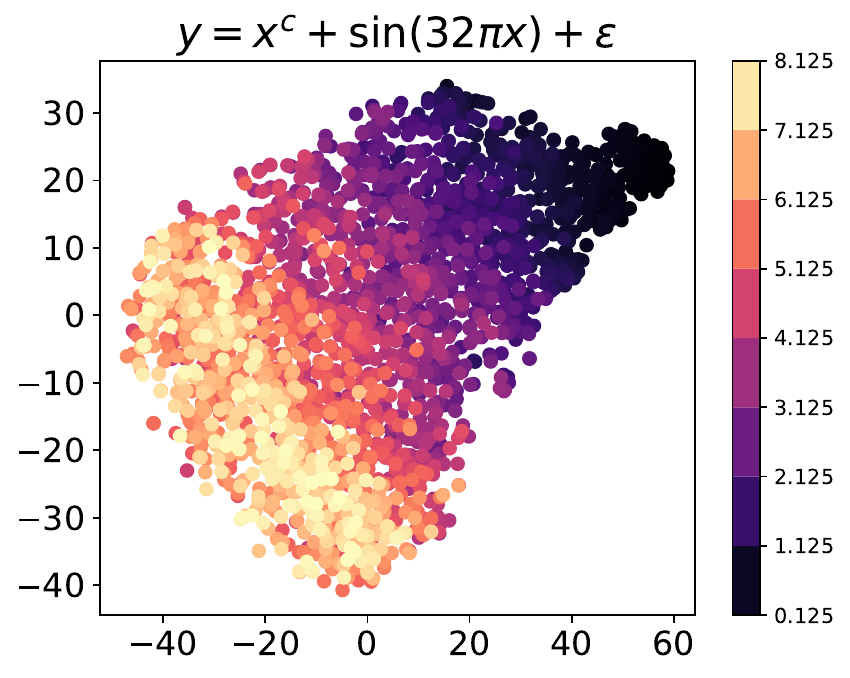} & 
\includegraphics[width=0.18\linewidth]{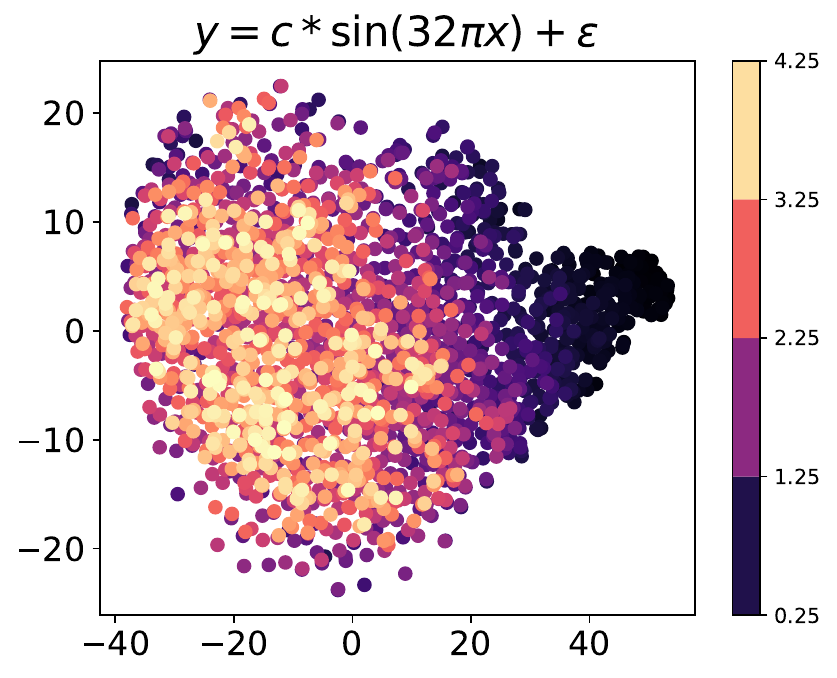} & 
\includegraphics[width=0.18\linewidth]{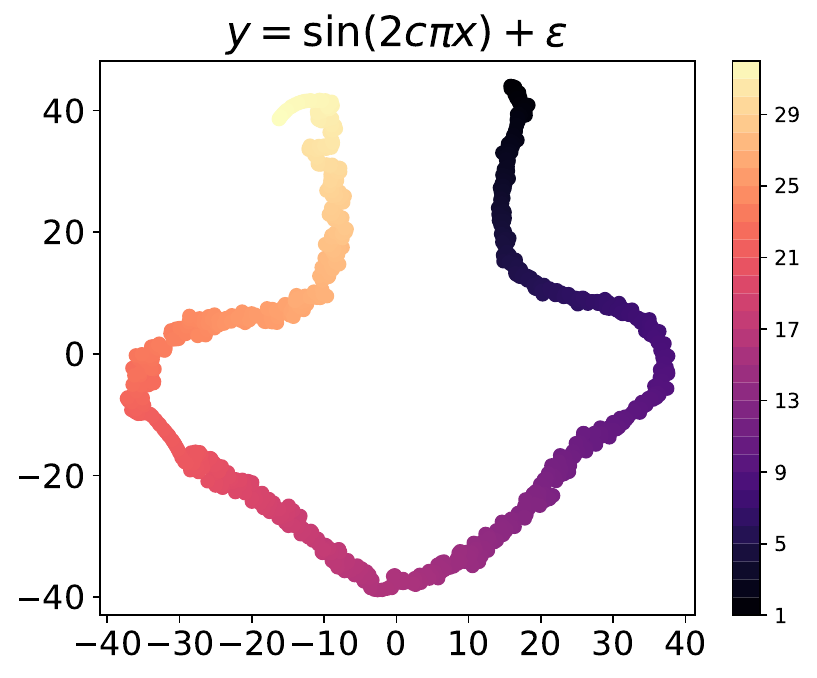} & 
\includegraphics[width=0.18\linewidth]{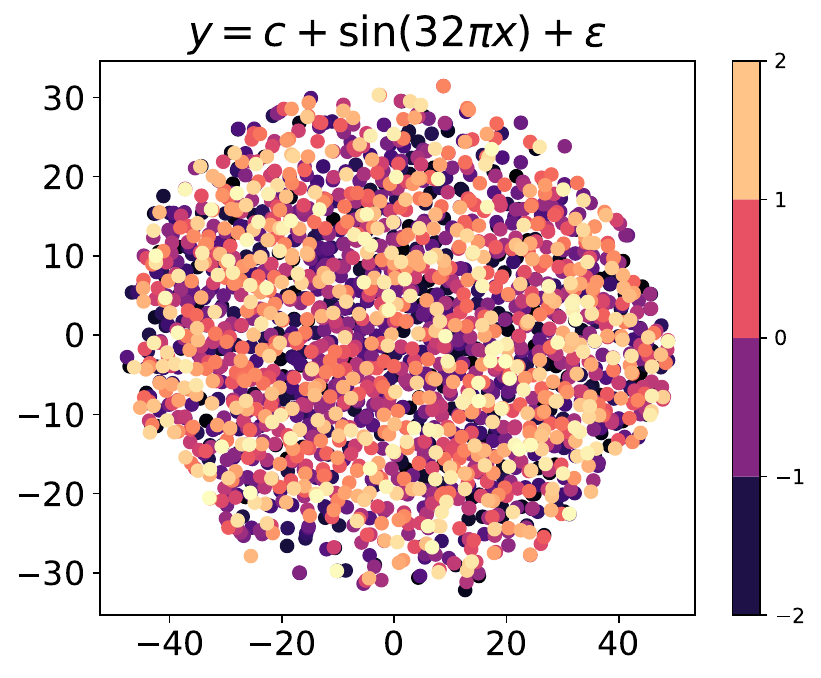} & 
\includegraphics[width=0.18\linewidth]{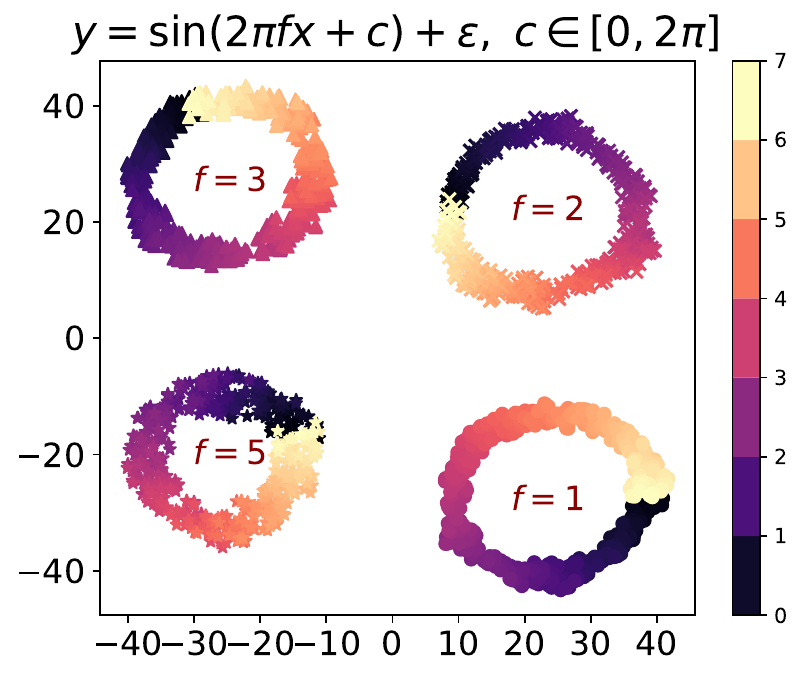} \\
\textit{(i)} Trend & 
\textit{(ii)} Amplitude & 
\textit{(iii)} Frequency & 
\textit{(iv)} Basline Shift & 
\textit{(v)} Auto-correlation 
\end{tabular}
\caption{What is \texttt{MOMENT} learning? Structure in the PCA (top) and t-SNE (bottom) visualizations of the embeddings of synthetically generated sinusoids suggest that MOMENT can capture subtle trend, scale, frequency, and auto-correlation information. $\epsilon$ denotes gaussian
noise with $0$ mean and $0.1$ standard deviation. $c$ controls the factor of interest, i.e. the power of the trend polynomial, amplitude, and frequency of the sine waves in experiments (i), (ii) \& (iii), respectively.}
\label{fig:interpretability_appendix}
\end{figure*}

To investigate what \texttt{MOMENT} is learning, we conducted a series of experiments using synthetically generated sine waves to evaluate \texttt{MOMENT}’s ability to capture changes in trend, amplitude, frequencies, baselines, and phase of time series. In each experiment, $c$ controls the factor of interest, for example the power of the trend polynomial $c \in (\frac{1}{8}, 8)$ \citep{N-BEATS} (Fig.~\ref{fig:interpretability_timeseries}), and frequency $c \in (1, 32)$ of the generated sine waves (Fig.~\ref{fig:interpretability_timeseries}). We generate multiple sine waves by varying $c$, derive their sequence-level representations using \texttt{MOMENT} (Sec. \ref{subsec:finetuning-MOMENT-on-downstream-tasks}), and visualize them in a 2- dimensional space using PCA and t-SNE \citep{tsne} in Fig.~\ref{fig:interpretability} and Fig.~\ref{fig:interpretability_appendix}. 

We also study the composition of the learnable mask embedding and the relationship between frequency and reconstruction error in a zero-shot setting. We find that the learned mask embedding is approximately composed of numbers drawn from the standard normal and that \texttt{MOMENT} can reconstruct lower frequency signals better. We observed a curious spike in reconstruction error around time series of frequency $c = 64$. 
(Fig.~\ref{fig:interpretability-experiment-mask-frequency-reconstruction})

\begin{figure*}[htb!]
\centering
\setlength{\tabcolsep}{0pt}
\begin{tabular}{ccccc}
\includegraphics[width=0.18\linewidth]{figures/Crop_pca.pdf} &
\includegraphics[width=0.18\linewidth]{figures/ElectricDevices_pca.pdf} &
\includegraphics[width=0.18\linewidth]{figures/ECG5000_pca.pdf} &
\includegraphics[width=0.18\linewidth]{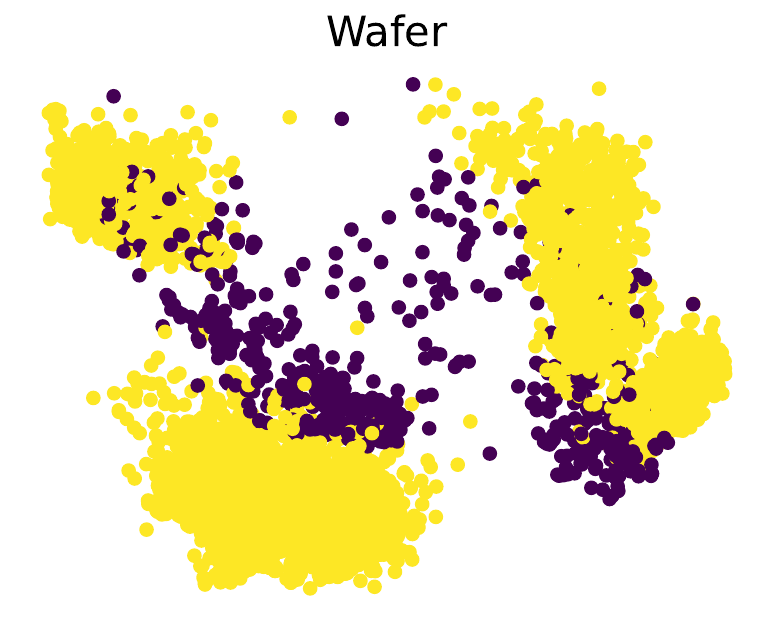} &
\includegraphics[width=0.18\linewidth]{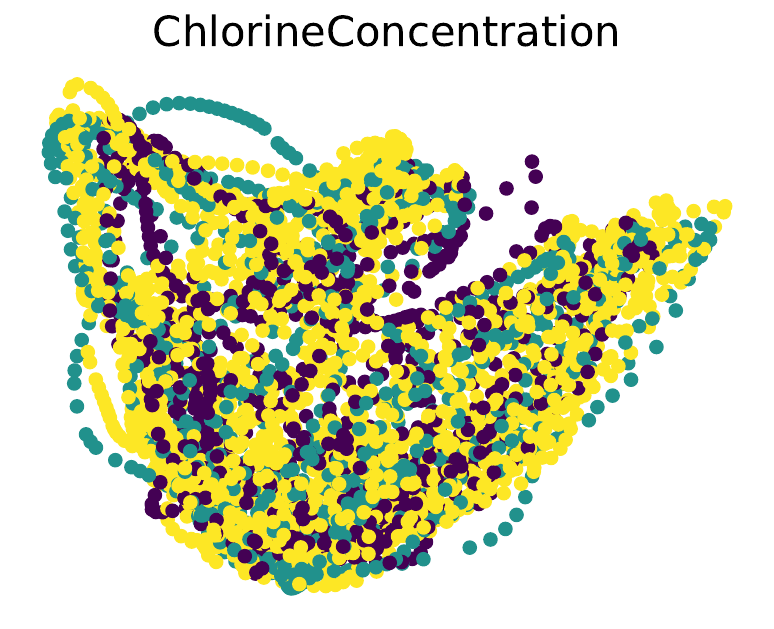} \\
\includegraphics[width=0.18\linewidth]{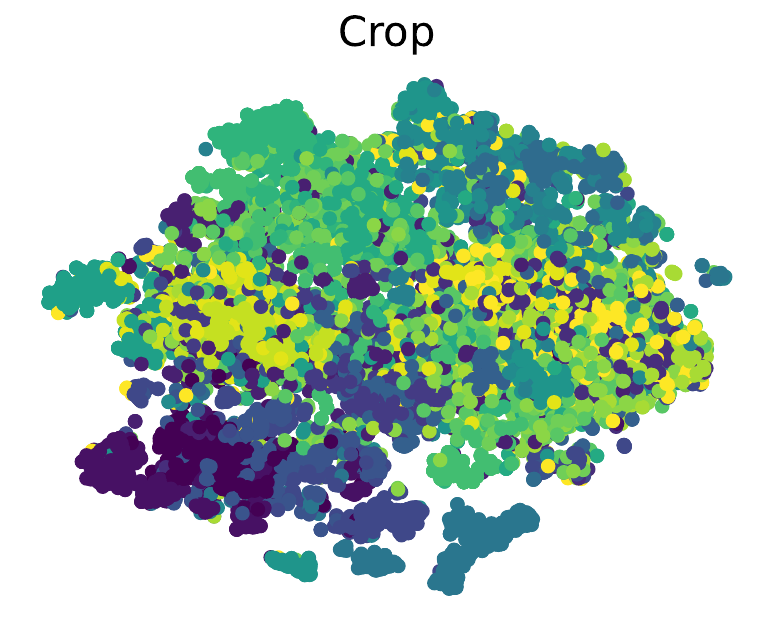} &
\includegraphics[width=0.18\linewidth]{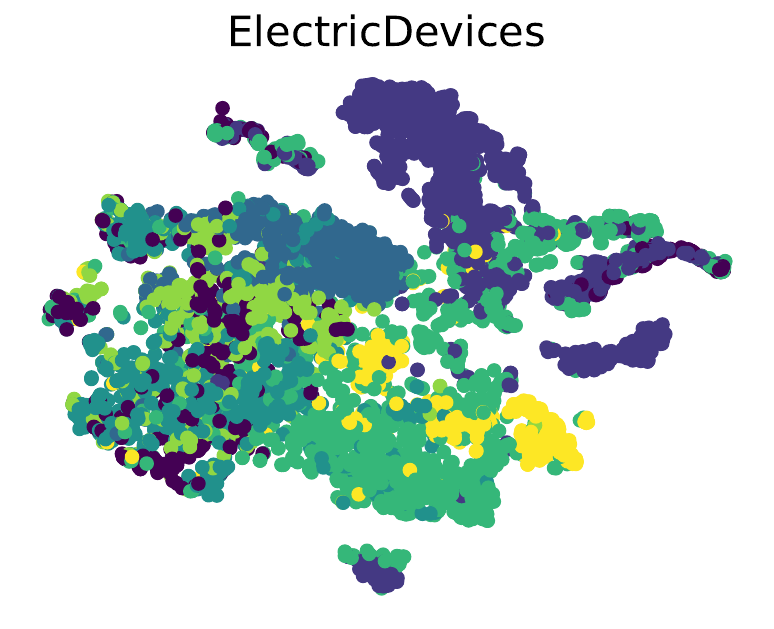} &
\includegraphics[width=0.18\linewidth]{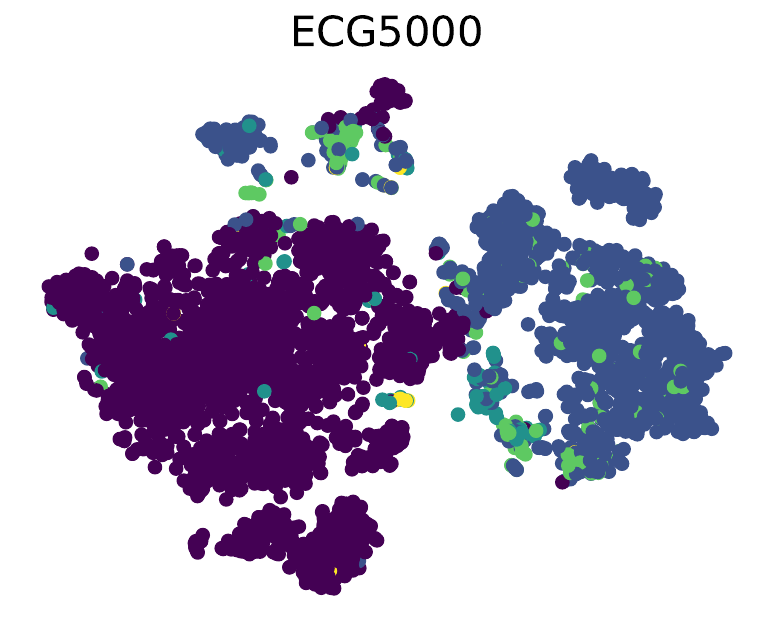} &
\includegraphics[width=0.18\linewidth]{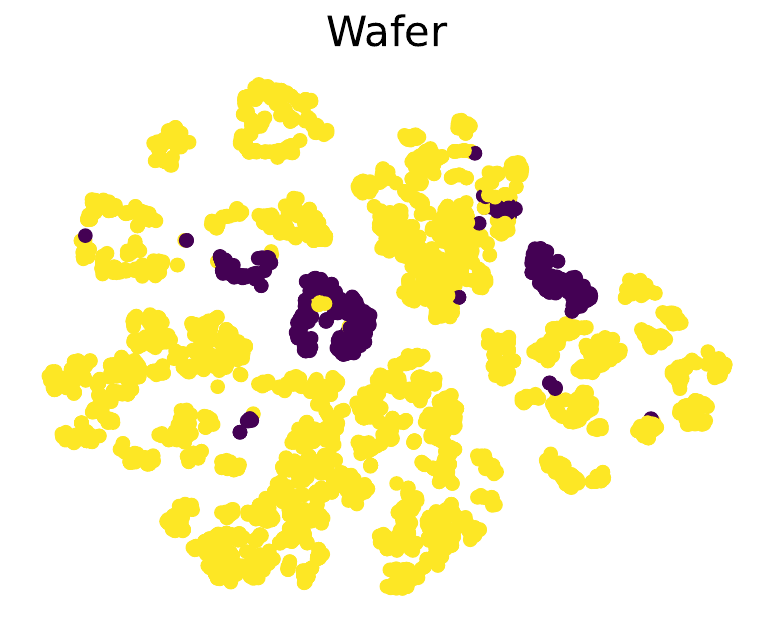} &
\includegraphics[width=0.18\linewidth]{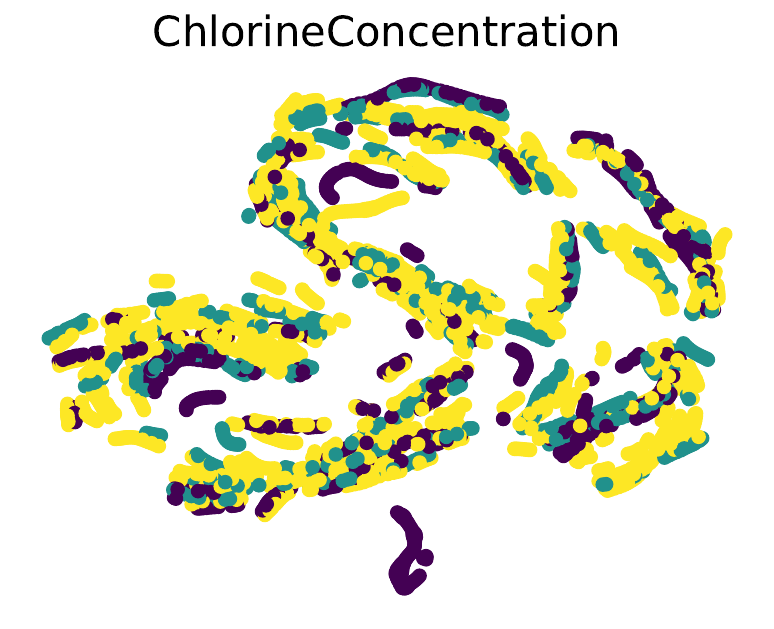} \\
\end{tabular}
\caption{PCA (top) and t-SNE (bottom) visualizations of representations learned by \texttt{MOMENT} on the 5 largest UCR datasets. Different colors represent different classes. Even without dataset-specific fine-tuning, \texttt{MOMENT} learns distinct representations for different classes}
\label{fig:clustering_ucr_appendix}
\end{figure*}

\begin{figure}[htb!]
\centering
\setlength{\tabcolsep}{0pt}
\begin{tabular}{c}
\includegraphics[width=0.8\columnwidth]{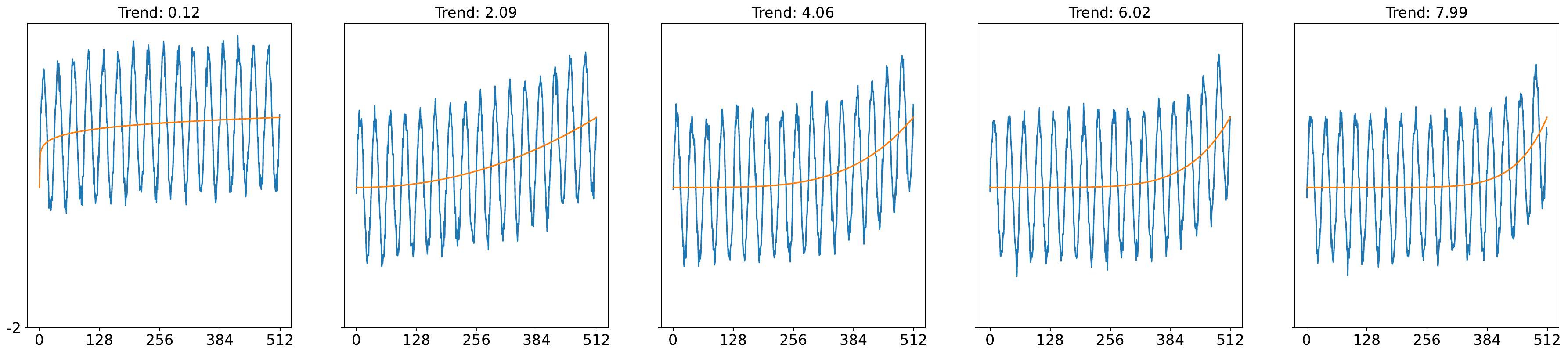} \\ 
\textit{(i)} Trend \\
\includegraphics[width=0.8\columnwidth]{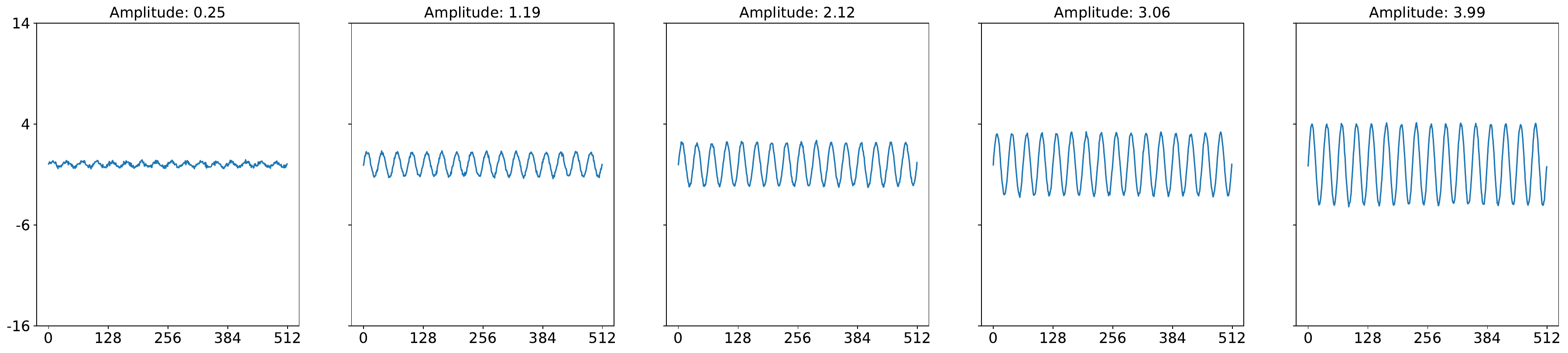} \\
\textit{(ii)} Amplitude \\
\includegraphics[width=0.8\columnwidth]{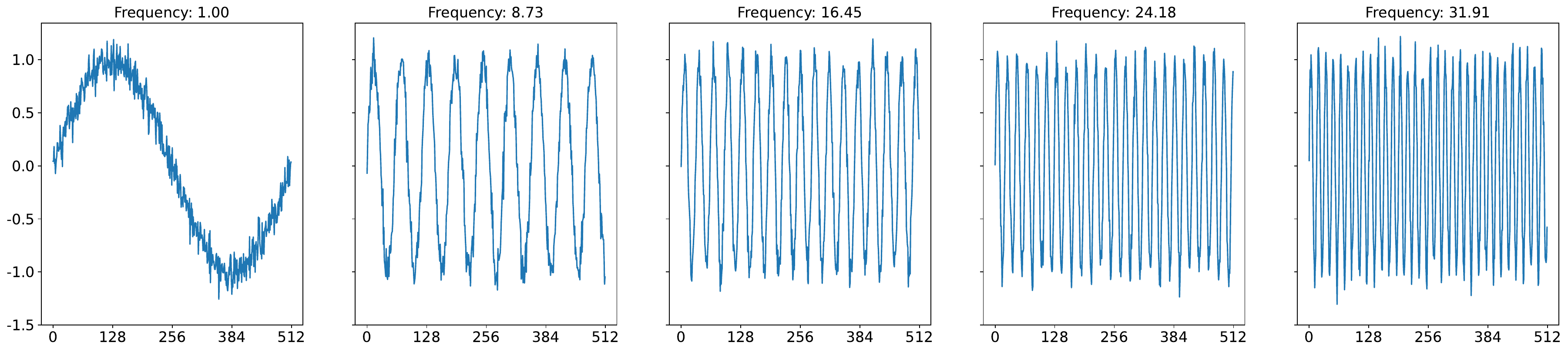} \\
\textit{(iii)} Frequency \\
\includegraphics[width=0.8\columnwidth]{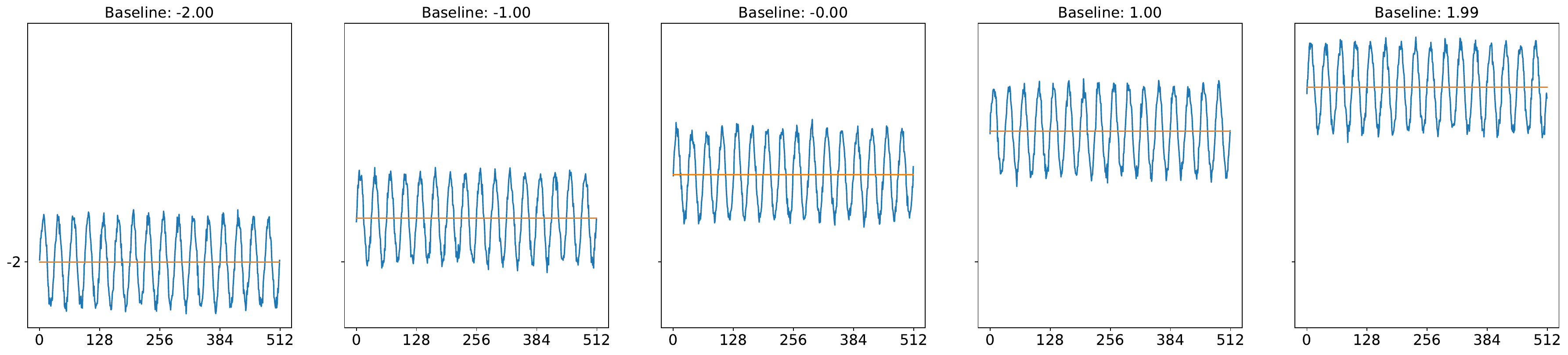} \\
\textit{(iv)} Baseline Shift \\
\includegraphics[width=0.8\columnwidth]{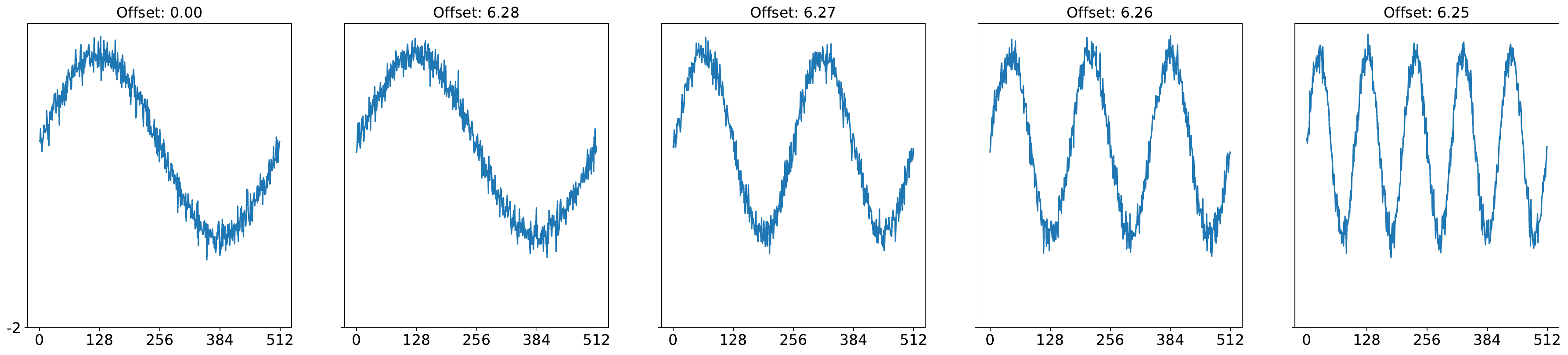} \\
\textit{(v)} Phase \\
\end{tabular}
\caption{Examples of sinusoids used in the interpretability experiments. }
\label{fig:interpretability_timeseries}
\end{figure}

\begin{figure*}[htb!]
\centering
\setlength{\tabcolsep}{0pt}
\begin{tabular}{cccc}
\includegraphics[width=0.24\linewidth]{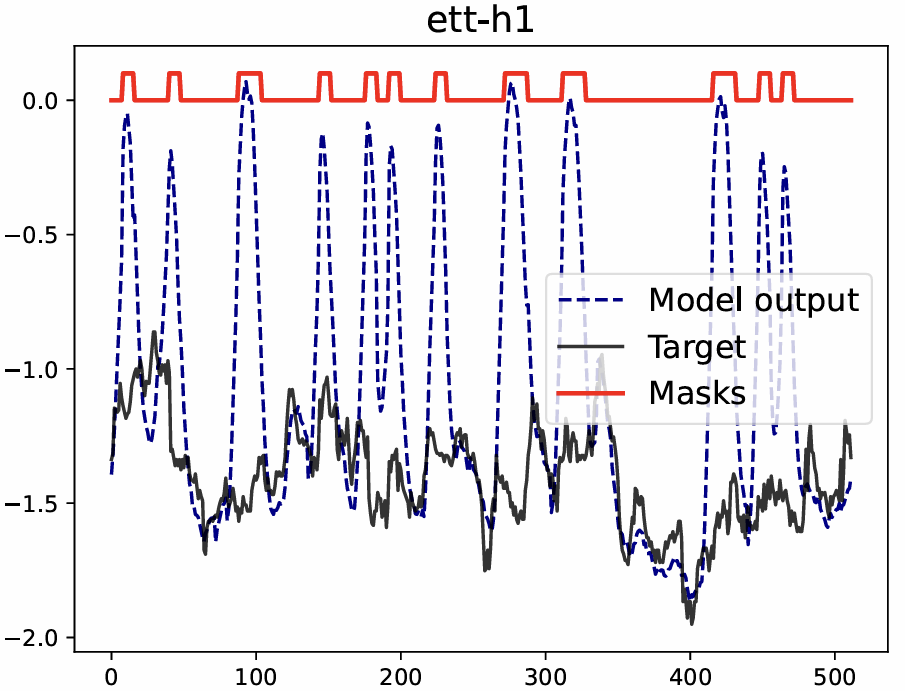} &
\includegraphics[width=0.24\linewidth]{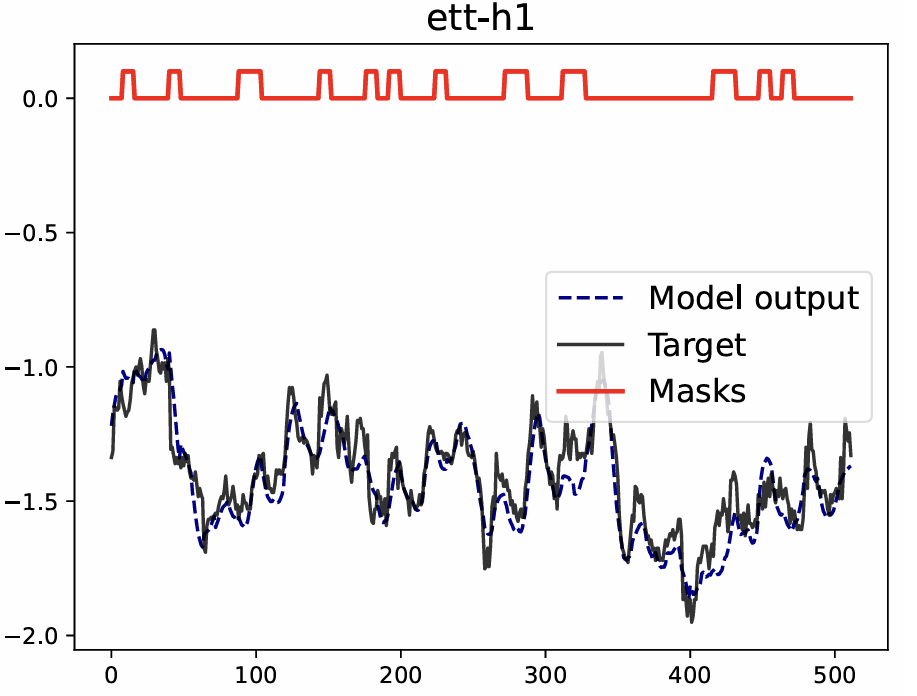} &
\includegraphics[width=0.24\linewidth]{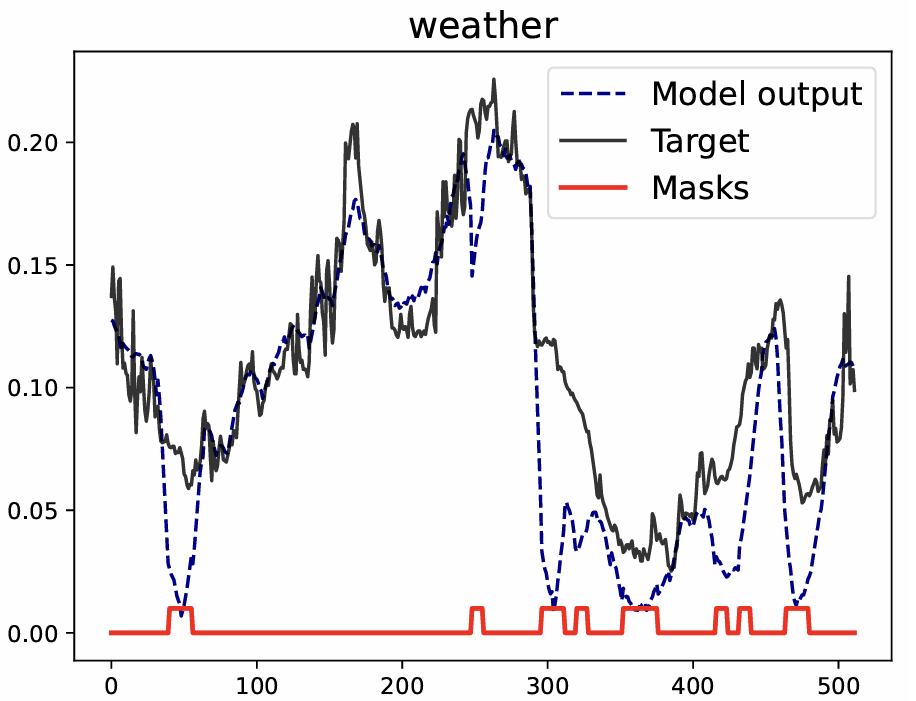} &
\includegraphics[width=0.24\linewidth]{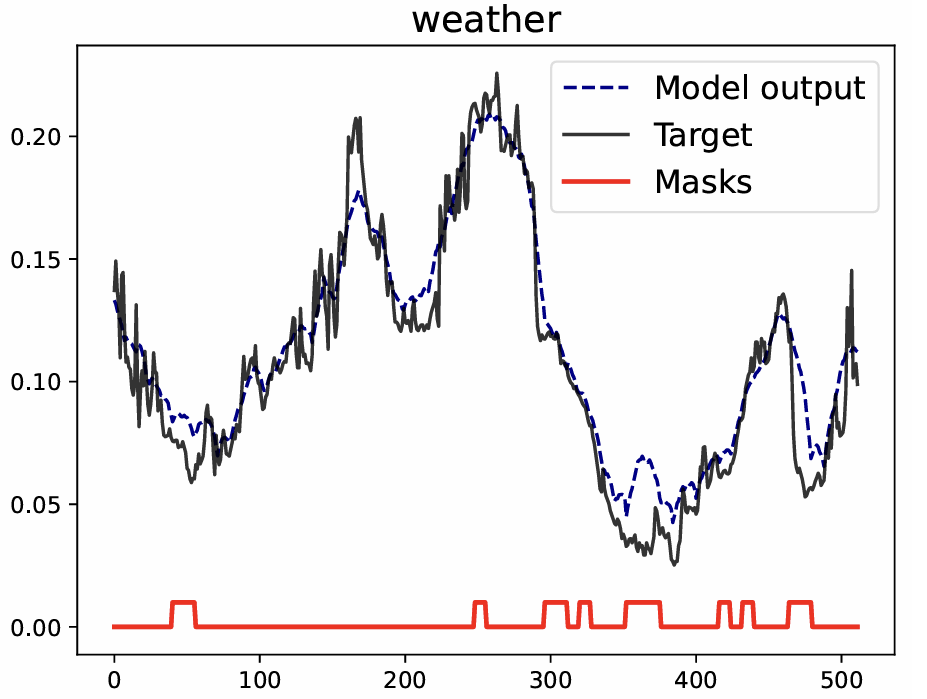} \\
\end{tabular}

\caption{Masking using a \texttt{[MASK]} tokens allows \texttt{MOMENT} to reconstruct time series in a zero-shot setting. Since zeros contain information, they bias model predictions. For two datasets ETTh1 and Weather, we mask the time series with zeros on the left and special mask tokens on the right.}
\label{fig:masking-with-zeros-appendix}
\end{figure*}

\begin{figure}
\centering
\setlength{\tabcolsep}{0pt}
\begin{tabular}{cc}
\includegraphics[width=0.45\columnwidth]{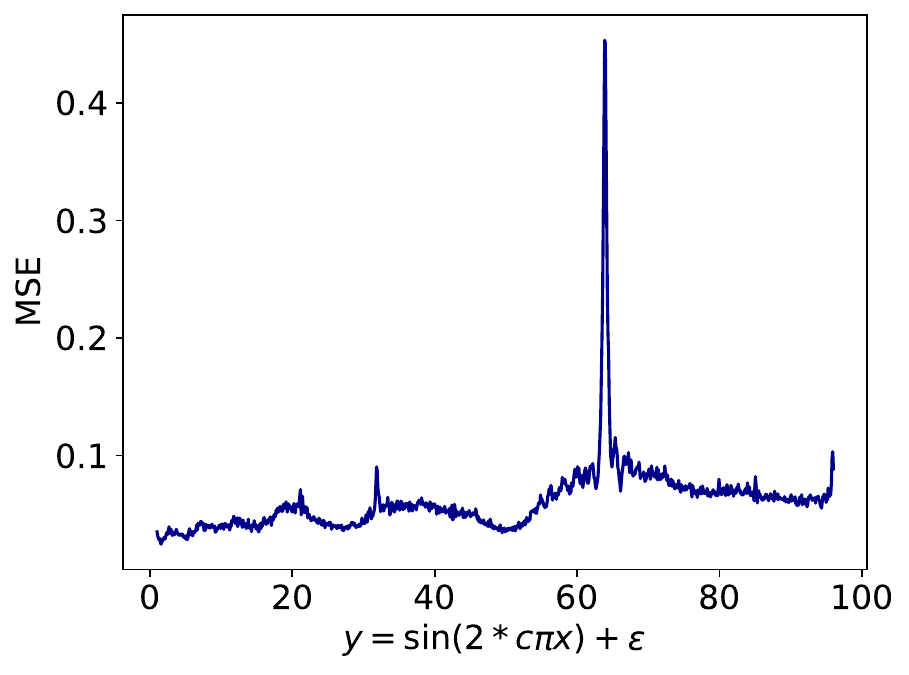}
& 
\includegraphics[width=0.45\columnwidth]{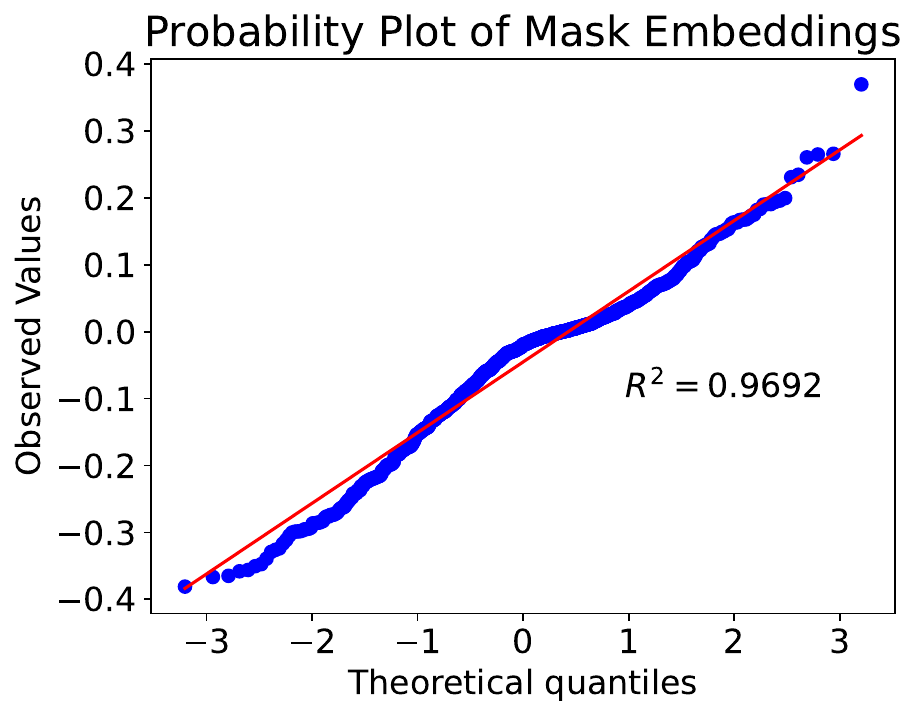} 
\end{tabular}

\caption{(\textit{Left}) \texttt{MOMENT} can reconstruct lower frequency time series better in a zero-shot setting. (\textit{Right}) The learned mask token is approximately composed of numbers drawn from a standard normal.}
\label{fig:interpretability-experiment-mask-frequency-reconstruction}
\end{figure}

\subsection{Impact of Model Size}
\
We studied the impact of scaling the size of the model and training data on zero-shot forecasting, imputation, and anomaly detection performance. As shown in Fig. \color{blue}x\color{black}, we found that increasing the size of the model generally improved zero-shot performance (lower MSE and sMAPE, higher VUS-ROC). Since varying the size of the pre-training dataset is expensive, we instead look at the zero-shot performance of model checkpoints before completing the first epoch. Our findings suggest that increasing the diversity in training data may also improve zero-shot performance.

\begin{table}[htb!]
\centering
\resizebox{0.65\linewidth}{!}{
\begin{tabular}{cccccccc}
\toprule
\multirow{2}{*}{Dataset} & \multirow{2}{*}{Pred\_horizon} & \multicolumn{2}{c}{$\mathtt{MOMENT_{small}}$} & \multicolumn{2}{c}{$\mathtt{MOMENT_{base}}$} & \multicolumn{2}{c}{$\mathtt{MOMENT_{large}}$} \\
 &  & MSE & MAE & MSE & MAE & MSE & MAE \\ \midrule
\multirow{4}{*}{Weather} & 96 & 0.167 & 0.224 & 0.156 & 0.211 & 0.151 & 0.207 \\
 & 192 & 0.210 & 0.259 & 0.198 & 0.248 & 0.195 & 0.246 \\
 & 336 & 0.256 & 0.292 & 0.247 & 0.284 & 0.245 & 0.285 \\
 & 720 & 0.315 & 0.334 & 0.315 & 0.334 & 0.316 & 0.333 \\ \midrule
\multirow{4}{*}{ETTh1} & 96 & 0.388 & 0.411 & 0.391 & 0.414 & 0.381 & 0.406 \\
 & 192 & 0.420 & 0.432 & 0.420 & 0.433 & 0.412 & 0.425 \\
 & 336 & 0.443 & 0.451 & 0.424 & 0.438 & 0.429 & 0.442 \\
 & 720 & 0.457 & 0.473 & 0.426 & 0.451 & 0.453 & 0.468 \\ \bottomrule
\end{tabular}}
\caption{Long-horizon forecasting scaling experiments for $\mathtt{MOMENT_{small}}$, $\mathtt{MOMENT_{base}}$, and $\mathtt{MOMENT_{large}}$.}
\label{tab:scaling_exp_lhf}
\end{table}

\begin{table}[!htb]
\centering
\resizebox{0.6\linewidth}{!}{
\begin{tabular}{cc|c|c|c}
\toprule
\multicolumn{2}{c}{Metric}        & $\mathtt{MOMENT_{small}}$ & $\mathtt{MOMENT_{base}}$ & $\mathtt{MOMENT_{large}}$ \\ \midrule
\multirow{3}{*}{Adj. F1} & Mean   & 0.480       & 0.572      & 0.569       \\
                         & Median & 0.450       & 0.641      & 0.607       \\
                         & Std.   & 0.378       & 0.383      & 0.372       \\ \midrule
\multirow{3}{*}{Vus ROC} & Mean   & 0.643       & 0.677      & 0.660       \\
                         & Median & 0.644       & 0.669      & 0.657       \\
                         & Std.   & 0.137       & 0.121      & 0.130       \\ \bottomrule
\end{tabular}}
\caption{Zero-shot anomaly detection scaling experiments for $\mathtt{MOMENT_{small}}$, $\mathtt{MOMENT_{base}}$, and $\mathtt{MOMENT_{large}}$.}
\label{tab:scaling_exp_zs_ad}
\end{table}

\begin{table}[!htb]
\centering
\resizebox{0.6\linewidth}{!}{
\begin{tabular}{c|ccc}
\toprule
 & $\mathtt{MOMENT_{small}}$ & $\mathtt{MOMENT_{base}}$ & $\mathtt{MOMENT_{large}}$ \\ \midrule
Mean & 0.716705 & 0.766437 & 0.763933 \\
Median & 0.720930 & 0.771078 & 0.766667 \\ \midrule
Std. & 0.155945 & 0.156763 & 0.160345 \\ \midrule
Avg. rank & 2.686813 & 1.659341 & 1.653846 \\
Median rank & 3.000000 & 2.000000 & 1.50 \\ \midrule
Wins/Losses & 28.5/153.5 & 122.0/60.0 & 122.5/59.5 \\ \bottomrule
\end{tabular}}
\caption{Zero-shot classification scaling experiments for $\mathtt{MOMENT_{small}}$, $\mathtt{MOMENT_{base}}$, and $\mathtt{MOMENT_{large}}$.}
\label{tab:scaling_exp_zs_classifcation}
\end{table}

\subsection{Training losses}
\label{sec:training_losses}

\begin{figure}[!tbh]
\centering
\setlength{\tabcolsep}{0pt}
\begin{tabular}{ccc}
\includegraphics[width=0.33\columnwidth]{figures/BETT_family_training_losses.pdf} & 
\includegraphics[width=0.33\columnwidth]{figures/random_vs_flant5_initialization.pdf} & 
\includegraphics[width=0.33\columnwidth]{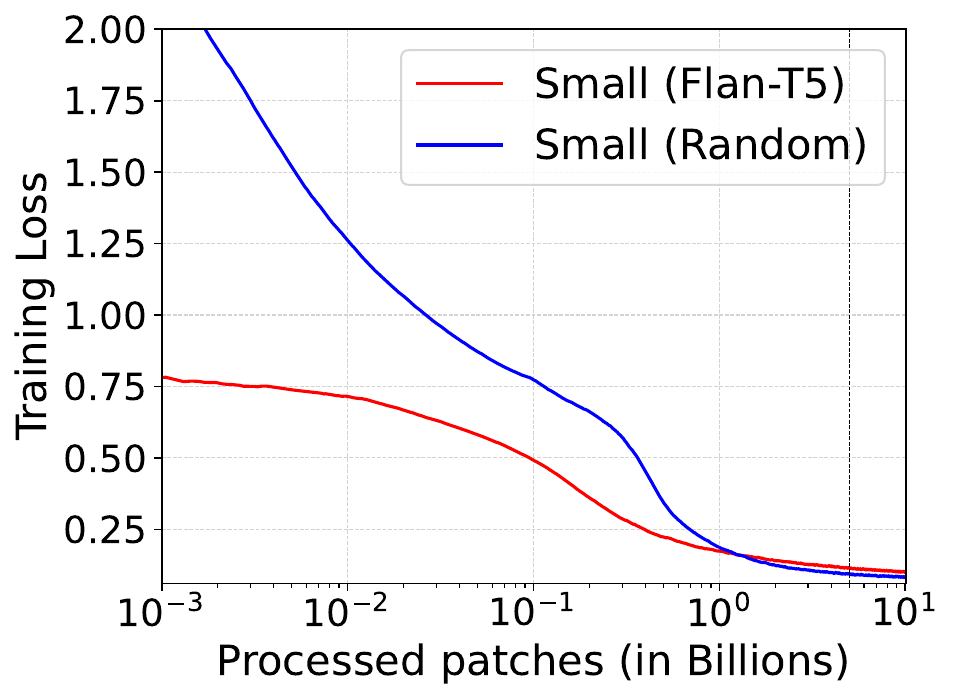} \\
\end{tabular}
\caption{\textbf{Training losses (MSE).} A dashed vertical line denotes the first epoch. All models were trained with a batch size of 131072 patches. (\textit{left}) Larger models obtain lower training loss. \textit{right} Eventually, randomly initialized \texttt{MOMENT-small} outperform the same model initialized with Flan-T5 weights. The figure on the right is in log scale.}
\label{fig:training_losses_appendix}
\end{figure}

\subsection{Efficiency Analysis}

\begin{table}[htb!]
\centering
\resizebox{0.6\linewidth}{!}{
\begin{tabular}{c|ccc}
\toprule
\multirow{2}{*}{\textbf{Model}} & \multicolumn{3}{c}{\textbf{ETTh1-96}} \\
 & \textbf{Total Param. (M)} & \textbf{Trainable Param. (M)} & \textbf{Mem. (MiB)} \\ \midrule
\texttt{MOMENT} & 347.53 & 6.29 & 2079 \\
GPT4TS & 82.28 & 1.12 & 1031 \\
TimesNet & 0.89 & 0.89 & 683 \\
Time-LLM & 3623.71 & 254.37 & 4537 \\ \toprule
\end{tabular}}

\caption{Efficiency analysis of \texttt{MOMENT} against other forecasting models on the ETTh1 with prediction horizon set to 96. \texttt{MOMENT} outperforms all the listed models and has a fraction of parameters as the most recent LLM-based forecasting method.}
\label{tab:efficiency-timellm-vs-MOMENT}
\end{table}

\section{Transparency Index}

\begin{table*}[htb!]
\centering
\begin{minipage}{0.41\linewidth}
\resizebox{\linewidth}{!}{
\begin{tabular}{l|r|c}
\toprule
\textbf{Sub-domain} & \textbf{Indicator} & \textbf{\texttt{MOMENT}} \\ \midrule
\multirow{10}{*}{Data} & Data size & \cellcolor{blue!25} 1 \\ \cline{3-3}
 & Data sources & \cellcolor{blue!25} 1 \\ \cline{3-3}
 & Data creators & \cellcolor{red!25}0 \\ \cline{3-3}
 & Data source selection & \cellcolor{blue!25}1 \\ \cline{3-3}
 & Data curation & \cellcolor{blue!25}1 \\ \cline{3-3}
 & Data augmentation & \cellcolor{blue!25}1 \\ \cline{3-3}
 & Harmful data filtration & \cellcolor{red!25}0 \\ \cline{3-3}
 & Copyrighted data & \cellcolor{blue!25}1 \\ \cline{3-3}
 & Data license & \cellcolor{blue!25}1 \\ \cline{3-3}
 & Personal information in data & \cellcolor{blue!25}1 \\ \midrule
\multirow{7}{*}{Data Labor} & Use of human labor & \cellcolor{blue!25}1 \\ \cline{3-3}
 & Employment of data laborers & \cellcolor{blue!25}1 \\ \cline{3-3}
 & Geographic distribution of data laborers & \cellcolor{blue!25}1 \\ \cline{3-3}
 & Wages & \cellcolor{blue!25}1 \\ \cline{3-3}
 & Instructions for creating data & \cellcolor{blue!25}1 \\ \cline{3-3}
 & Labor protections & \cellcolor{blue!25}1 \\ \cline{3-3}
 & Third party partners & \cellcolor{blue!25}1 \\ \midrule
\multirow{2}{*}{Data Access} & Queryable external data access & \cellcolor{blue!25}1 \\ \cline{3-3}
 & Direct external data access & \cellcolor{blue!25}1 \\ \midrule
\multirow{7}{*}{Compute} & Compute usage & \cellcolor{blue!25}1 \\ \cline{3-3}
 & Development duration & \cellcolor{blue!25}1 \\ \cline{3-3}
 & Compute hardware & \cellcolor{blue!25}1 \\ \cline{3-3}
 & Hardware owner & \cellcolor{blue!25}1 \\ \cline{3-3}
 & Energy usage & \cellcolor{blue!25}1 \\ \cline{3-3}
 & Carbon emissions & \cellcolor{blue!25}1 \\ \cline{3-3}
 & Broader environmental impact & \cellcolor{red!25}0 \\ \midrule
\multirow{4}{*}{Methods} & Model stages & \cellcolor{blue!25}1 \\ \cline{3-3}
 & Model objectives & \cellcolor{blue!25}1 \\ \cline{3-3}
 & Core frameworks & \cellcolor{blue!25}1 \\ \cline{3-3}
 & Additional dependencies & \cellcolor{blue!25}1 \\ \midrule
\multirow{2}{*}{Data Mitigations} & Mitigations for privacy & \cellcolor{red!25}0 \\ \cline{3-3}
 & Mitigations for copyright & \cellcolor{red!25}0 \\ \midrule
 & \textbf{Upstream Subtotal} & 75\% \\ \toprule
\end{tabular}}
\end{minipage}%
\begin{minipage}{0.49\linewidth}
\resizebox{\linewidth}{!}{
\begin{tabular}{l|r|c}
\toprule
\textbf{Sub-domain} & \textbf{Indicator} & \multicolumn{1}{c}{\textbf{MOMENT}} \\ \midrule
\multirow{6}{*}{Model Basics} & Input modality & \cellcolor{blue!25} 1 \\ \cline{3-3}
 & Output modality & \cellcolor{blue!25} 1 \\ \cline{3-3}
 & Model components & \cellcolor{blue!25}1 \\ \cline{3-3}
 & Model size & \cellcolor{blue!25}1 \\ \cline{3-3}
 & Model architecture & \cellcolor{blue!25}1 \\ \cline{3-3}
 & Centralized model documentation & \cellcolor{blue!25}1 \\ \midrule
\multirow{3}{*}{Model Access} & External model access protocol & \cellcolor{blue!25}1 \\ \cline{3-3}
 & Blackbox external model access & \cellcolor{blue!25}1 \\ \cline{3-3}
 & Full external model access & \cellcolor{blue!25}1 \\ \midrule
\multirow{5}{*}{Capabilities} & Capabilities description & \cellcolor{blue!25}1 \\ \cline{3-3}
 & Capabilities demonstration & \cellcolor{blue!25}1 \\ \cline{3-3}
 & Evaluation of capabilities & \cellcolor{blue!25}1 \\ \cline{3-3}
 & External reproducibility of capabilities evaluation & \cellcolor{red!25}0 \\ \cline{3-3}
 & Third party capabilities evaluation & \cellcolor{red!25}0 \\ \midrule
\multirow{3}{*}{Limitations} & Limitations description & \cellcolor{blue!25}1 \\ \cline{3-3}
 & Limitations demonstration & \cellcolor{blue!25}1 \\ \cline{3-3}
 & Third party evaluation of limitations & \cellcolor{red!25}0 \\ \midrule
\multirow{7}{*}{Risks} & Risks description & \cellcolor{blue!25} 1 \\ \cline{3-3}
 & Risks demonstration & \cellcolor{red!25}0 \\ \cline{3-3}
 & Unintentional harm evaluation & \cellcolor{red!25}0 \\ \cline{3-3}
 & External reproducibility of unintentional harm evaluation & \cellcolor{red!25}0 \\ \cline{3-3}
 & Intentional harm evaluation & \cellcolor{red!25}0 \\ \cline{3-3}
 & External reproducibility of intentional harm evaluation & \cellcolor{red!25}0 \\ \cline{3-3}
 & Third party risks evaluation & \cellcolor{red!25}0 \\ \midrule
\multirow{2}{*}{Model Mitigations} & Mitigations description & \cellcolor{red!25}0 \\ \cline{3-3}
 & Mitigations demonstration & \cellcolor{red!25}0 \\ \midrule
\multirow{3}{*}{Mitigations} & Mitigations evaluation & \cellcolor{red!25}0 \\ \cline{3-3}
 & External reproducibility of mitigations evaluation & \cellcolor{red!25}0 \\ \cline{3-3}
 & Third party mitigations evaluation & \cellcolor{red!25}0 \\ \midrule
\multirow{2}{*}{Trustworthiness} & Trustworthiness evaluation & \cellcolor{red!25}0 \\ \cline{3-3}
 & External reproducibility of trustworthiness evaluation & \cellcolor{red!25}0 \\ \midrule
\multirow{2}{*}{Inference} & Inference duration evaluation & \cellcolor{blue!25}1 \\ \cline{3-3}
 & Inference compute evaluation & \cellcolor{blue!25}1 \\ \midrule
 & \textbf{Model Subtotal} & 51.5\% \\ \toprule
\end{tabular}
}
\end{minipage}
\caption{Expected (\textit{left}) upstream and (\textit{right}) model transparency scores. \texttt{MOMENT} has one of the highest upstream transparency. Our model transparency scores are lower due to (third-party) harm, mitigations, trustworthiness evaluation, which are not well understood for time series modeling.}
\label{tab:transparency-index-model}
\end{table*}

\section{Results Sources}

\begin{table}[htb!]
\centering
\resizebox{\linewidth}{!}{
\begin{tabular}{c|cccc}
\toprule
\textbf{Task} & \textbf{Method} & \textbf{Type} & \textbf{Reimplementation/ Rerun} & \textbf{Source} \\ \midrule
\multirow{8}{*}{\begin{tabular}[c]{@{}c@{}}Long-horizon\\ Forecasting\end{tabular}} & Time-LLM & \multirow{2}{*}{LLM-based} & $\checkmark$ & \href{https://github.com/KimMeen/Time-LLM}{\texttt{Time-LLM}} \\
 & GPT4TS &  & $\times$ & \href{https://arxiv.org/pdf/2302.11939.pdf}{\texttt{One Fits All}} \\ \cline{2-5}
 & \begin{tabular}[c]{@{}c@{}}PatchTST, Fedformer, Autoformer,\\ Stationary, ETSformer, LightTS,\\ Informer, Reformer\end{tabular} & \multirow{2}{*}{Transformer-based} & $\times$ & \href{https://arxiv.org/pdf/2302.11939.pdf}{\texttt{One Fits All}} \\
 & Pyraformer, LogTrans &  & $\times$ & \href{https://arxiv.org/pdf/2210.02186.pdf}{\texttt{TimesNet}} \\ \cline{2-5}
 & TimesNet, DLinear & \multirow{2}{*}{Deep learning} & $\times$ & \href{https://arxiv.org/pdf/2302.11939.pdf}{\texttt{One Fits All}} \\
 & N-BEATS &  & $\checkmark$ & \href{https://github.com/philipperemy/n-beats/tree/master}{\texttt{N-BEATS}} \\ \midrule
\multirow{5}{*}{\begin{tabular}[c]{@{}c@{}}Short-horizon\\ Forecasting\end{tabular}} & GPT4TS & LLM-based & $\checkmark$ & \href{https://arxiv.org/pdf/2302.11939.pdf}{\texttt{One Fits All}} \\ \cline{2-5}
 & TimesNet & \multirow{2}{*}{Deep learning} & $\checkmark$ & \href{https://github.com/thuml/Time-Series-Library}{\texttt{TimesNet}} \\ 
 & N-BEATS &  & $\checkmark$ & \href{https://github.com/philipperemy/n-beats/tree/master}{\texttt{N-BEATS}} \\ \cline{2-5}
 & \begin{tabular}[c]{@{}c@{}}AutoARIMA, AutoTheta, AutoETS,\\ Seasonal Naive, Naive, Random Walk\end{tabular} & Statistical learning & $\checkmark$ & \href{https://github.com/Nixtla/statsforecast}{\texttt{Nixtla Statsforecast Repository}} \\ \midrule
\multirow{5}{*}{Classification} & GPT4TS & LLM-based & $\checkmark$ & \href{https://arxiv.org/pdf/2302.11939.pdf}{\texttt{One Fits All}} \\ \cline{2-5}
 & TimesNet & Deep learning & $\checkmark$ & \href{https://github.com/thuml/Time-Series-Library}{\texttt{TimesNet}} \\ \cline{2-5}
 & TS2Vec, T-Loss, TNC, TS-TCC, TST & Unsupervised Representation learning & $\times$ & \href{https://github.com/yuezhihan/ts2vec}{\texttt{TS2Vec}} \\ \cline{2-5}
 & \begin{tabular}[c]{@{}c@{}}CNN, Encoder, FCN, MCNN, MLP,\\ ResNet, t-LeNet, TWIESN\end{tabular} & Deep learning & $\times$ & \href{https://github.com/hfawaz/dl-4-tsc}{\texttt{DL4TSC Repository}} \\ \cline{2-5}
 & DTW & Statistical learning & $\times$ & \href{https://github.com/yuezhihan/ts2vec}{\texttt{TS2Vec}} \\ \midrule
\multirow{5}{*}{\begin{tabular}[c]{@{}c@{}}Anomaly\\ Detection\end{tabular}} & GPT4TS & LLM-based & $\checkmark$ & \href{https://arxiv.org/pdf/2302.11939.pdf}{\texttt{One Fits All}} \\ \cline{2-5}
 & TimesNet & Deep learning & $\checkmark$ & \href{https://github.com/thuml/Time-Series-Library}{\texttt{TimesNet}} \\ \cline{2-5}
 & Anomaly Transformer & Transformer-based & $\checkmark$ & \href{https://github.com/thuml/Anomaly-Transformer}{\texttt{Anomaly Transformer}} \\ \cline{2-5}
 & DGHL & Deep learning & $\checkmark$ & \href{https://github.com/mononitogoswami/tsad-model-selection}{\texttt{Time Series Model Selection}} \\ \cline{2-5}
 & $k$-NN & Statistical learning & $\checkmark$ & \href{https://github.com/mononitogoswami/tsad-model-selection}{\texttt{Time Series Model Selection}} \\ \midrule
\multirow{8}{*}{Imputation} & GPT4TS & LLM-based & $\checkmark$ & \href{https://arxiv.org/pdf/2302.11939.pdf}{\texttt{One Fits All}} \\ \cline{2-5}
 & TimesNet & Deep learning & $\checkmark$ & \href{https://github.com/thuml/Time-Series-Library}{\texttt{TimesNet}} \\ \cline{2-5}
 & \begin{tabular}[c]{@{}c@{}}PatchTST, ETSformer, LightTS,\\ Fedformer,Stationary, Autoformer,\\ Informer, Reformer\end{tabular} & Transformer-based & $\times$ & \href{https://arxiv.org/pdf/2302.11939.pdf}{\texttt{One Fits All}} \\ \cline{2-5}
 & DLinear & Deep learning & $\times$ & \href{https://arxiv.org/pdf/2302.11939.pdf}{\texttt{One Fits All}} \\ \cline{2-5}
 & Naive & \multirow{2}{*}{Statistical learning} & $\checkmark$ & \href{https://pandas.pydata.org/docs/reference/api/pandas.DataFrame.ffill.html}{\texttt{Pandas FFill}}, \href{https://pandas.pydata.org/docs/reference/api/pandas.DataFrame.bfill.html}{\texttt{Pandas BFill}} \\
 & Linear, Nearest, Cubic &  & $\checkmark$ & \href{https://docs.scipy.org/doc/scipy/reference/generated/scipy.interpolate.interp1d.html}{\texttt{Scipy Interp1D}} \\ \midrule
\end{tabular}}

\caption{Source for the results for each baseline for all downstream task.}
\label{tab:results-sources}
\end{table}

\section{Radar Plot}
We generate a radar plot (Fig.~\ref{fig:model_comparison}) to visually compare \texttt{MOMENT} with GPT4TS and TimesNet. The values obtained by each method for a given task are min-max normalized with respect to the other methods for each of the 5 downstream tasks. For imputation, long- and short-horizon forecasting, we report $1 -$ the normalized MSE or sMAPE for the methods on the weather and (subset of) M4 datasets, respectively. For classification and anomaly detection, we report the average accuracy and VUS-ROC of the methods across all the datasets. 

\end{document}